\documentclass{article}
\usepackage[preprint]{neurips_2026}
\usepackage[utf8]{inputenc}
\usepackage[T1]{fontenc}
\usepackage[dvipsnames]{xcolor}
\usepackage[colorlinks]{hyperref}
\hypersetup{
 colorlinks=True,
 linkcolor=nhdpurple,
 citecolor=gray,
 urlcolor=nhdpurple}
\usepackage{url}
\usepackage{graphicx}
\usepackage{wrapfig}
\usepackage{amsmath}
\usepackage{amssymb}
\usepackage{mathtools}
\usepackage{amsthm}
\usepackage{amsfonts}
\usepackage{nicefrac}
\usepackage{enumitem}
\usepackage{colortbl}
\usepackage{makecell}
\usepackage[most]{tcolorbox}
\usepackage{algorithm}
\usepackage[noend]{algorithmic}
\usepackage[normalem]{ulem}
\usepackage[format=plain,
            labelfont=it,
            labelsep=period]{caption}
\usepackage{subcaption}
\usepackage{booktabs}
\usepackage{bbding}
\usepackage{fancyvrb}
\usepackage{titletoc}
\usepackage{listings}
\usepackage{overpic}
\definecolor{codegreen}{rgb}{0,0.5,0}
\definecolor{codered}{rgb}{0.7,0.1,0.1}
\definecolor{codegray}{rgb}{0.5,0.5,0.5}
\definecolor{codepurple}{rgb}{0.58,0,0.82}
\definecolor{backcolour}{rgb}{1,1,1}
\lstdefinestyle{python}{
    language=Python,
    backgroundcolor=\color{backcolour},   
    commentstyle=\color{codered}\textit,
    keywordstyle=\bfseries\color{codegreen},
    numberstyle=\tiny\color{codegray},
    stringstyle=\color{codepurple},
    basicstyle=\ttfamily\scriptsize,
    breakatwhitespace=false,         
    breaklines=true,                 
    captionpos=b,                    
    keepspaces=true,                 
    numbers=left,                    
    numbersep=4pt,                  
    showspaces=false,                
    showstringspaces=false,
    showtabs=false,                  
    tabsize=1,
    fancyvrb=true
}
\definecolor{nhblue}{HTML}{28679E}
\definecolor{nhlblue}{HTML}{F5F7F9}
\definecolor{nhgblue}{HTML}{6C8194}
\definecolor{nhgray}{HTML}{ABACAB}
\definecolor{nhdgray}{HTML}{505050}
\definecolor{nhbrown}{HTML}{855637}
\definecolor{nhburgundy}{HTML}{90285C}
\definecolor{nhgreen}{HTML}{619028}
\definecolor{nhpurple}{HTML}{9150BE}
\definecolor{nhdpurple}{HTML}{75409C}
\definecolor{nhred}{HTML}{D62727}
\definecolor{nhteal}{HTML}{66C2A4}
\definecolor{nhorange}{HTML}{BF8130}
\definecolor{gred}{rgb}{0.859,0.267,0.216}
\definecolor{ggreen}{rgb}{0.059,0.616,0.345}
\definecolor{gblue}{rgb}{0.259,0.522,0.957}
\definecolor{gyellow}{rgb}{0.957,0.706,0}
\definecolor{gpurple}{rgb}{0.565,0.173,0.894}
\definecolor{cella}{rgb}{1.0, 0.92, 0.92}

\newcommand{\cmark}{\textcolor{nhdpurple}{\Checkmark}}
\newcommand{\xmark}{\textcolor{nhgray}{\XSolidBrush}}
\newcommand{\pmark}{\textcolor{nhorange}{$\mathbf{\sim}$}}

\newcommand{\panellabel}[1]{\vphantom{Cpy}#1}

\renewenvironment{abstract}
  {\vspace{0.5ex}}
  {\vskip 1ex}

\title{Hallucination in World Models is\\Predictable and Preventable}

\author{Nicklas Hansen\textcolor{gray}{$^{1}$}\hspace*{7pt}
  \textbf{Xiaolong Wang}\textcolor{gray}{$^{1}$}\vspace{7pt}\\\vspace{3pt}\textcolor{gray}{$^{1}$}UC San Diego
}

\begin{document}

\maketitle

\begin{abstract}
\vspace{-0.3in}
\begin{tcolorbox}[
    enhanced,
    colback=white,
    colframe=gray!25!white,
    boxsep=9pt,
    arc=2mm,
    boxrule=1pt,
    width=\textwidth-2pc,
    center,
]
\vspace{-1pt}
\textbf{Abstract.} Modern generative world models render increasingly realistic action-controllable futures, yet they frequently \emph{hallucinate}: rollouts remain visually fluent while drifting from the ground-truth dynamics. We hypothesize that hallucination concentrates in low-coverage regions of the state-action space, where lightweight data-centric signals can both detect it and guide mitigation. To test this, we introduce MMBench2, a $427$-hour, $210$-task dataset for visual world modeling with ground-truth actions, rewards, and live simulators, and train a $350$M-parameter world model on it. We identify three distinct hallucination modes: perceptual, action-marginalized, and scene-diverging -- each anchored to a different stage of the pipeline, and develop three signals that accurately predict where the model will fail. To close coverage gaps at training time, we develop a coverage-aware sampling technique; to close them online, our hallucination predictors serve as curiosity rewards for targeted data collection, yielding a data-efficient finetuning recipe that adapts the pretrained world model to entirely unseen environments with as few as $50$ real environment trajectories. Overall, our findings reveal that hallucination in world models is inherently a data coverage issue, and that the same signals used to detect it can also be used for mitigation.
\vspace{3pt}
\begin{center}
    \textbf{Webpage: \url{https://nicklashansen.com/mmbench2}}
\end{center}
\vspace{-4pt}
\end{tcolorbox}
\end{abstract}

\vspace{0.1in}
\section{Introduction}
\label{sec:introduction}
Modern generative world models render strikingly realistic, action-controllable futures across diverse environments \citep{alonso2024diffusion, valevski2024gamengen, hafner2025training, deepmind2025genie3}. However, the rollouts they produce frequently \emph{hallucinate}: they remain visually fluent and superficially plausible while drifting away from the ground-truth dynamics \citep{ Janner2019WhenTT}. The term is borrowed from the language modeling literature \citep{ji2023hallucination, huang2025hallucination}, where it typically denotes generation of factually incorrect text; analogous failures have also been studied in image \citep{li2023pope} and video generation \citep{huang2024vbench}. In a world model, the failure is arguably more consequential: hallucinated trajectories are fed directly into downstream planners and policies \citep{schrittwieser2020mastering, hafner2023dreamerv3, hansen2024tdmpc2}, so silent hallucination during rollout translates into silently incorrect decisions during control.

\begin{figure}[h]
    \centering
    \begin{minipage}{0.3\textwidth}%
        \centering    
        \begin{overpic}[width=0.475\textwidth]{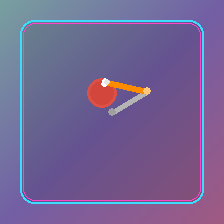}
            \put(-0.1,86.85){\colorbox{white}{\textcolor{black}{enc}}}
        \end{overpic}
        \begin{overpic}[width=0.475\textwidth]{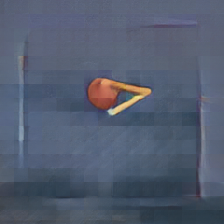}
            \put(-0.1,86.85){\colorbox{white}{\textcolor{black}{dec}}}
        \end{overpic}
        \\\textcolor{nhdgray}{\small Reconstruction of OOD scene poor but structurally faithful {\footnotesize\Checkmark}}\\[1em]
        \begin{overpic}[width=0.475\textwidth]{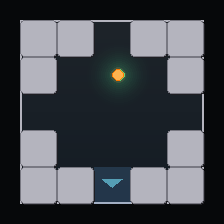}
            \put(-0.1,86.85){\colorbox{white}{\textcolor{black}{enc}}}
        \end{overpic}
        \begin{overpic}[width=0.475\textwidth]{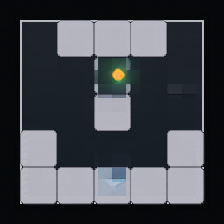}
            \put(-0.1,86.85){\colorbox{white}{\textcolor{black}{dec}}}
        \end{overpic}
        \\\textcolor{nhdpurple}{\small \textbf{Unseen layout is reconstructed as a similar, \emph{seen} layout}} {\footnotesize\textcolor{nhdpurple}{\XSolidBrush}}\\[0.5em]
        {\emph{(i)} \textbf{Perceptual}}
    \end{minipage}\hfill
    \begin{minipage}{0.3\textwidth}%
        \centering
        \includegraphics[width=0.475\textwidth]{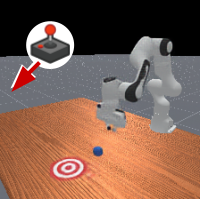}
        \includegraphics[width=0.475\textwidth]{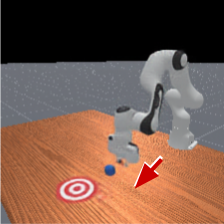}
        \\\textcolor{nhdgray}{\small Predicted next frame reflects the action taken by the user {\footnotesize\Checkmark}}\\[1em]
        \includegraphics[width=0.475\textwidth]{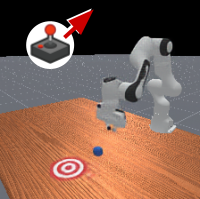}
        \includegraphics[width=0.475\textwidth]{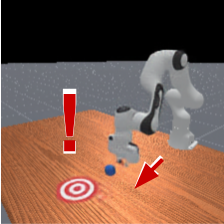}\\\textcolor{nhdpurple}{\small \textbf{Visually plausible but \emph{ignores} the action actually taken}} {\footnotesize\textcolor{nhdpurple}{\XSolidBrush}}\\[0.5em]
        {\emph{(ii)} \textbf{Action marginalization}}
    \end{minipage}\hfill
    \begin{minipage}{0.3\textwidth}%
        \centering
        \includegraphics[width=0.475\textwidth]{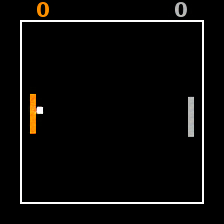}
        \includegraphics[width=0.475\textwidth]{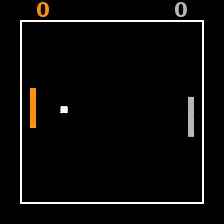}
        \\\textcolor{nhdgray}{\small Plausible dynamics despite slight error accumulation {\footnotesize\Checkmark}}\\[1em]
        \begin{overpic}[width=0.475\textwidth]{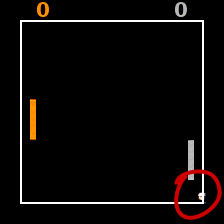}
            \put(56,86.85){\colorbox{white}{\textcolor{black}{\small$t{=}34$}}}
        \end{overpic}
        \begin{overpic}[width=0.475\textwidth]{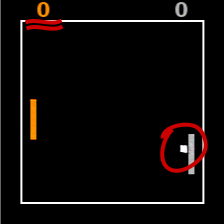}
            \put(56,86.85){\colorbox{white}{\textcolor{black}{\small$t{=}36$}}}
        \end{overpic}
        \\\textcolor{nhdpurple}{\small \textbf{\emph{Implausible} dynamics (ball teleports back into play)}} {\footnotesize\textcolor{nhdpurple}{\XSolidBrush}}\\[0.5em]
        {\emph{(iii)} \textbf{Scene divergence}}
    \end{minipage}
    \caption{\textbf{Hallucination.} We categorize \textbf{\textcolor{nhdpurple}{hallucinations}} as three distinct failure modes: \emph{perceptual}, \emph{action marginalization}, and \emph{scene divergence}, and develop methods for detection and mitigation.}
    \label{fig:hallucination}
    \vspace{-0.2in}
\end{figure}

Despite the increasing fidelity of these models, \emph{where} an autoregressive rollout will hallucinate, and \emph{why}, is poorly understood. A natural reading is that hallucination is an architectural problem to be solved by ever-larger backbones and more training compute \citep{hoffmann2022training}. However, we argue that \textbf{hallucination in world models is primarily a data coverage problem:} it concentrates in low-coverage regions of the state-action space \citep{levine2020offline, gadre2023datacomp} and is therefore both \emph{predictable} from signals available at runtime \citep{lakshminarayanan2017simple, gal2016dropout, Chua2018DeepRL} and \emph{preventable} by adjusting which data the model is trained on rather than architectural changes. Our experiments reveal that a single underlying cause -- coverage gaps -- explains failures at every stage of the model pipeline: tokenizer, action-conditioning, and multi-step rollout, and that it manifests as three distinct failure modes shown in Figure~\ref{fig:hallucination}.

In this work, we set out to characterize different types of hallucinations, predict when they will occur using signals internal to the model, and use that information to close the underlying coverage gap. We explore two concrete ways in which the gap can be closed: \emph{(i)} using coverage-aware sampling techniques during large-scale pretraining, and \emph{(ii)} by targeted online data collection using our derived predictors of hallucination. Understanding when and why hallucinations happen requires three resources that no single benchmark currently provides: full control over the training corpus, behaviorally diverse data spanning many tasks and domains, as well as live environments to probe coverage gaps via online interaction. To address this gap we develop \textbf{MMBench2}, a massively multitask dataset for visual world modeling that extends MMBench~\citep{Hansen2025Newt} with $65.6$k mixed-quality trajectories equivalent to $427$ hours of $224{\times}224$ video at $15$ fps, complete with ground-truth action and reward labels and live simulators across $210$ tasks spanning $10$ domains. Of the $210$ tasks, $200$ form the pretraining corpus and $10$ are held out as entirely unseen transfer tasks, allowing us to probe coverage both within and beyond the training distribution.

Using MMBench2, we train a $350$M parameter Dreamer 4 \citep{hafner2025training} world model and dissect its hallucination behavior. We identify three distinct hallucination modes, each anchored to a different stage of the pipeline: \emph{perceptual} hallucination in the encoder-decoder pair, \emph{action marginalization} in the dynamics model, and \emph{scene divergence} during multi-step rollouts. We develop three predictors of hallucination: tokenizer round-trip residual, flow instability, and inter-seed denoising variance, and use them to detect hallucination at runtime, as well as curiosity reward for online data collection \citep{sekar2020plan} in seen and unseen tasks. Empirically, our hallucination-driven data collection enables adaptation to entirely unseen environments with just $50$ real trajectories, approaching the effectiveness of human data collection. Together, these results show that hallucination in modern world models is, to a large extent, a coverage problem, and that the same signals that detect it can also be used to mitigate it. \textbf{We summarize our contributions as follows:}
\begin{itemize}[itemsep=2pt, topsep=2pt, parsep=0pt, leftmargin=1.75em]
    \vspace{-0.05in}
    \item \textbf{MMBench2:} a large $427$-hour dataset for visual world modeling with ground-truth actions and rewards, live environments, and behaviorally diverse data including human play.
    \item A \textbf{stage-by-stage characterization of hallucination} in generative world models tied to tokenizer, action marginalization, and rollout failures.
    \item \textbf{Three hallucination predictors} that detect hallucination without labels or additional training.
    \item A \textbf{coverage-aware training recipe} that reduces hallucination across all three predictors and improves rollout fidelity at no additional cost.
    \item A framework for \textbf{targeted data collection} that adapts a pretrained $350$M world model to unseen environments with as few as $50$ trajectories.
\end{itemize}
To support further research on generative world modeling, \textbf{\emph{we release our full dataset, code for training and evaluation, model checkpoints, and a browser interface for open-ended interaction with the world model.}} See \textbf{\url{https://nicklashansen.com/mmbench2}} for videos and resources.

\begin{figure}
    \centering
    \includegraphics[width=0.111\linewidth]{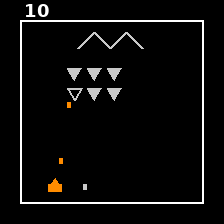}%
    \includegraphics[width=0.111\linewidth]{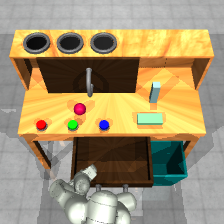}%
    \includegraphics[width=0.111\linewidth]{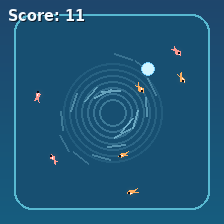}%
    \includegraphics[width=0.111\linewidth]{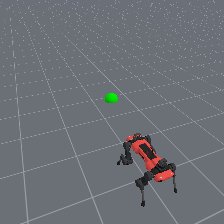}%
    \includegraphics[width=0.111\linewidth]{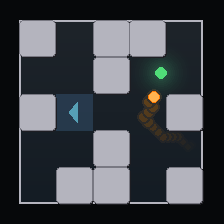}%
    \includegraphics[width=0.111\linewidth]{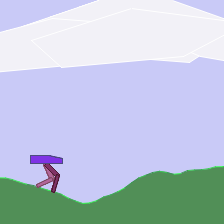}%
    \includegraphics[width=0.111\linewidth]{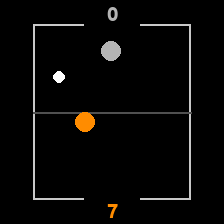}%
    \includegraphics[width=0.111\linewidth]{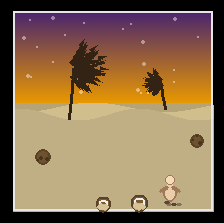}%
    \includegraphics[width=0.111\linewidth]{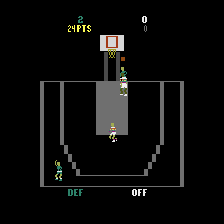}
    \\[-0.1em]
    \includegraphics[width=0.111\linewidth]{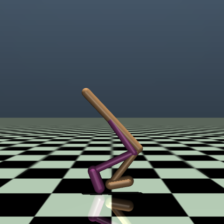}%
    \includegraphics[width=0.111\linewidth]{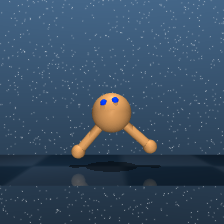}%
    \includegraphics[width=0.111\linewidth]{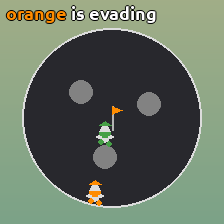}%
    \includegraphics[width=0.111\linewidth]{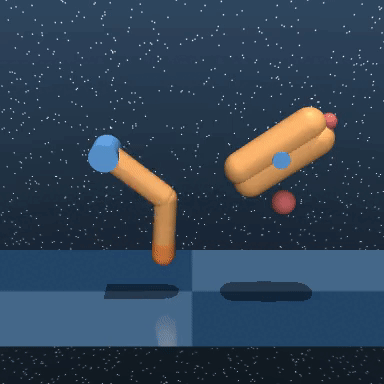}%
    \includegraphics[width=0.111\linewidth]{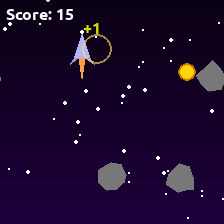}%
    \includegraphics[width=0.111\linewidth]{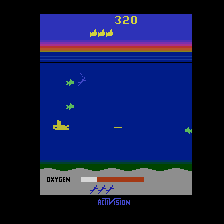}%
    \includegraphics[width=0.111\linewidth]{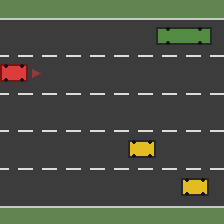}%
    \includegraphics[width=0.111\linewidth]{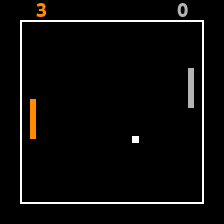}%
    \includegraphics[width=0.111\linewidth]{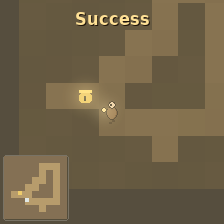}
    \\[-0.1em]
    \includegraphics[width=0.111\linewidth]{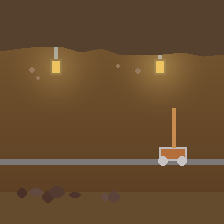}%
    \includegraphics[width=0.111\linewidth]{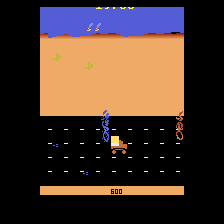}%
    \includegraphics[width=0.111\linewidth]{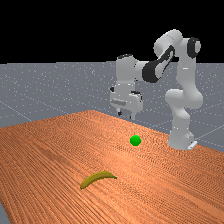}%
    \includegraphics[width=0.111\linewidth]{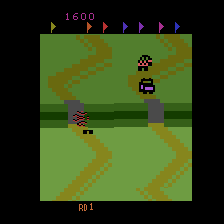}%
    \includegraphics[width=0.111\linewidth]{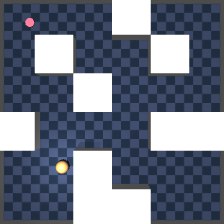}%
    \includegraphics[width=0.111\linewidth]{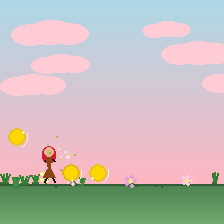}%
    \includegraphics[width=0.111\linewidth]{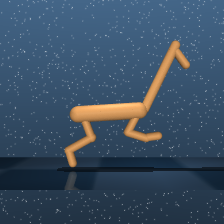}%
    \includegraphics[width=0.111\linewidth]{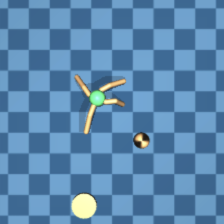}%
    \includegraphics[width=0.111\linewidth]{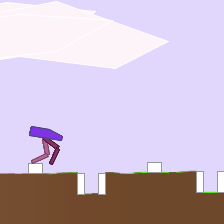}%
    \\[-0.1em]
    \includegraphics[width=0.111\linewidth]{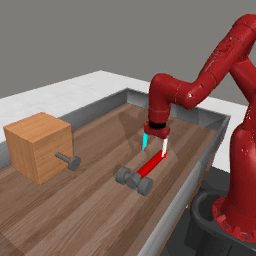}%
    \includegraphics[width=0.111\linewidth]{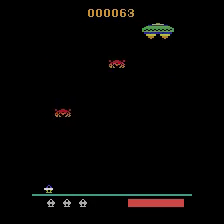}%
    \includegraphics[width=0.111\linewidth]{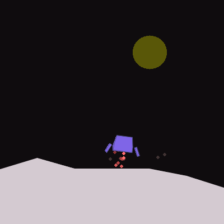}%
    \includegraphics[width=0.111\linewidth]{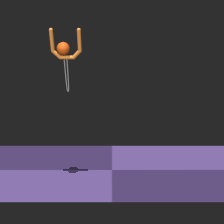}%
    \includegraphics[width=0.111\linewidth]{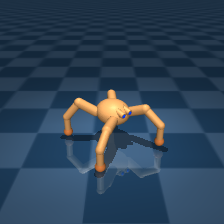}%
    \includegraphics[width=0.111\linewidth]{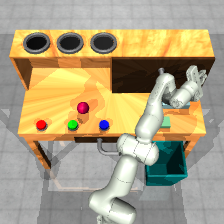}%
    \includegraphics[width=0.111\linewidth]{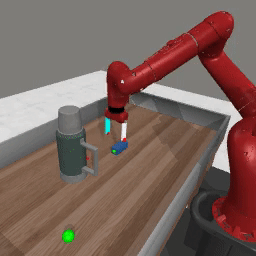}%
    \includegraphics[width=0.111\linewidth]{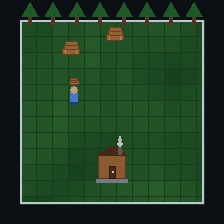}%
    \includegraphics[width=0.111\linewidth]{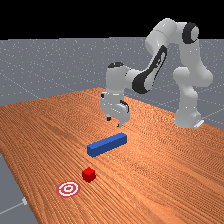}%
    \caption{\textbf{MMBench2 tasks.} A sample of $36$ of the $210$ tasks in MMBench2, illustrating the visual and morphological diversity of the corpus. All observations are $224{\times}224$ RGB frames.}
    \label{fig:task-overview}
    \vspace{-0.075in}
\end{figure}

\section{MMBench2: A Dataset for Visual World Modeling}
\label{sec:dataset}

To fully understand hallucination in large generative world models, it is necessary to have control over the training process as well as access to its training data, but (to our knowledge) no such resource exists today at a satisfactory scale. To address this gap, we propose \textbf{MMBench2}: a massively multitask dataset for visual world modeling. Our dataset consists of $\mathbf{65{,}600}$ trajectories ($427$ hours of $224{\times}224$ video at $15$ fps for a total of $23$M frames) with diverse behaviors (including human play data) across $\mathbf{210}$ different continuous control tasks, complete with ground-truth action and reward labels as well as live environments and language instructions for every task included in MMBench2. As the name implies, MMBench2 directly builds on top of MMBench which provides a collection of 200 live environments unified under a common interface. Figure~\ref{fig:task-overview} shows the breadth of our tasks.

\textbf{Task overview.} MMBench2 spans $\mathbf{10}$ task domains: DMControl~\citep{deepmindcontrolsuite2018}, DMControl Extended~\citep{Hansen2025Newt}, Meta-World~\citep{yu2019meta}, ManiSkill3~\citep{taomaniskill3}, MuJoCo~\citep{todorov2012mujoco}, MiniArcade~\citep{Hansen2025Newt}, Box2D~\citep{brockman2016gym}, RoboDesk~\citep{kannan2021robodesk}, OGBench~\citep{ogbench_park2025}, and Continuous Atari~\citep{farebrother2024cale}. Together they cover locomotion, dexterous and tabletop manipulation, goal-conditioned navigation, and arcade-style environments, with continuous action spaces of $1$ to $16$ dimensions; we zero-pad every action vector to the maximum dimension ($d_a{=}16$) and supply a per-dimension validity mask so models can ignore inactive dimensions per task. Of the $\mathbf{210}$ tasks, $200$ form the pretraining set and $10$ (proposed in this work) are held out as \emph{unseen} transfer tasks.

\begin{figure}
    \centering
    \includegraphics[width=0.925\linewidth]{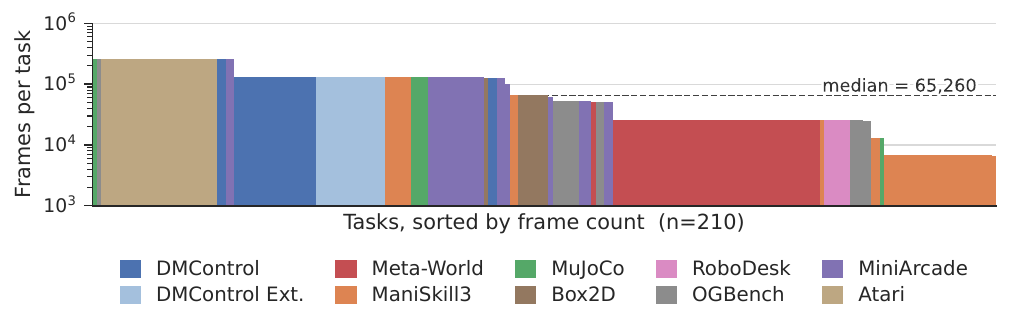}
    \vspace{-0.125in}
    \caption{\textbf{Dataset composition.} Per-task frame counts across the $210$ tasks in our training corpus (20M frames), sorted in descending order and colored by domain. The distribution is heavy-tailed: the top $20$ tasks account for $26\%$ of all frames (dominated by Atari) while the bottom $20$ contribute only $0.7\%$ (short-horizon manipulation tasks). The dashed line marks the per-task median ($65$k).}
    \label{fig:dataset}
    \vspace{-0.15in}
\end{figure}

The 200-task pretraining corpus contains 260 episodes per task. However, episode lengths vary by task, ranging from $25$ for certain ManiSkill3 manipulation tasks up to $1{,}000$ for Atari games. As a result, the corpus is highly non-uniform across tasks as shown in Figure~\ref{fig:dataset}. Additionally, the visual diversity \emph{within} a task also differs wildly across the corpus with, \emph{e.g.}, Inverted Pendulum (MuJoCo) varies little whereas tasks like Dungeon Explorer (proposed in this work), Bipedal Walker (Box2D), and Krull (Atari) vary greatly between frames and episodes. Please refer to Appendix~\ref{sec:appendix-environments} for an overview of tasks and Appendix~\ref{sec:data-collection} for a detailed description of the data collection process.

\section{Training a Large Visual World Model}
\label{sec:methods}

We seek to answer the following question: \emph{why} do world models hallucinate, and can we predict \emph{when} they are likely to hallucinate? To answer this question, we choose to develop and train a 350M parameter visual world model that largely follows the overall architecture and training recipe introduced in Dreamer~4~\citep{hafner2025training}. As such, our world model is an action-conditioned generative model that follows an increasingly common two-stage training recipe: \emph{(1)} first, a video \emph{tokenizer} is trained via masked auto-encoding of visual observations, and \emph{(2)} a dynamics model implemented as a block-causal Transformer \citep{vaswani2017attention} is subsequently trained via flow-matching over spatial latent tokens produced by the frozen tokenizer with additional conditioning on action tokens. We first pretrain our world model on the MMBench2 training corpus, and then address our specific research questions by conducting a series of targeted finetuning experiments. This section aims to provide an overview of our architecture and training recipe.

\textbf{Tokenizer.} Our video tokenizer is instantiated as a symmetric encoder-decoder Transformer architecture. The encoder ingests $224{\times}224$ RGB frames ``patchified'' at stride $14$ ($256$ patch tokens per frame), prepends $64$ learnable latent queries, and projects the latent stream to a $64$-dimensional bottleneck bounded by a $\tanh$ activation, producing a per-frame code $z \in [-1, 1]^{64\times 64}$. The decoder then reconstructs images conditioned only on latent codes. Training uses a masked reconstruction objective \citep{he2022masked} where a fraction of input patches drawn per-frame from $\mathcal{U}(0, 0.9)$ are replaced by a learned mask token, and the loss (pixel MSE + LPIPS \citep{zhang2018unreasonable}) is computed only on masked positions. We normalize losses by their running RMS to reduce sensitivity to hyperparameters. Encoder and decoder are 50M learnable parameters each.

\textbf{Dynamics model.} Our dynamics model is a 250M parameter block-causal Transformer over packed tokenizer latents, trained on top of the \emph{frozen} tokenizer introduced above. Per timestep, the input sequence consists of an action token (a $2$-layer MLP over the $16$-dimensional padded action), a shortcut-conditioning token (encoding noise level $\sigma$ and step size $d$), $32$ packed spatial latent tokens, $4$ register tokens, and optional agent tokens used as the readout for the reward and behavior-cloning heads. The model is trained with the \emph{shortcut flow-matching} objective of~\citet{frans2025one}, which interleaves an empirical one-step regression term with a self-consistency bootstrap that distills two coarser-step predictions into one finer-step target. At inference this enables next-frame sampling in as few as $4$ Euler substeps.

\textbf{Training recipe.} The training recipe of Dreamer 4 was developed for a setting where only a small percentage of the training data has action and reward labels. However, MMBench2 provides ground-truth actions and rewards for every trajectory in the dataset which we choose to take advantage of. We first pretrain the tokenizer on the full training corpus, and then proceed to also pretrain the dynamics model, conditioning on actions. After this initial pretraining phase, we experiment with additional mid-training and finetuning recipes introduced in the following sections. As part of our experiments, we initialize additional reward prediction and behavior cloning (BC) policy heads after pretraining. The reward predictor is trained via $L{=}8$ multi-step discrete regression (using $\operatorname{symlog}$ two-hot encoding) with gradients backpropagated through the dynamics model. The BC policy is a deterministic Gaussian policy trained to predict ground-truth actions via an MSE loss -- this is a departure from Dreamer 4 which only considered discrete actions.

\section{Hallucination in World Models}
\label{sec:hallucination}
A generative world model imagines a future by chaining three distinct operations: \emph{(1)} an encoder maps each observation into a latent code, \emph{(2)} an action-conditioned dynamics head predicts the next latent, and \emph{(3)} a decoder renders a reconstruction back to pixel space. Each of these stages is a learned function trained on a finite slice of the state-action space, and each can therefore fail \emph{independently} when asked to extrapolate beyond what it has seen. We use the term \emph{hallucination} to refer to any such failure: an output that is fluent and visually plausible, yet decoupled from reality. Crucially, because the three stages compose sequentially, a hallucination introduced early (\emph{e.g.}\ a corrupted encoding) is propagated and amplified by the stages that follow, so naming \emph{which} stage produces a given failure is a prerequisite for diagnosing and fixing it. This section explores how hallucinations can be characterized, detected, and finally mitigated.

\subsection{Characterizing Hallucination}
\label{sec:characterizing-hallucination}
We identify three distinct ways in which a generative world model may hallucinate, each tied to a different stage of the imagination pipeline. We illustrate each type of hallucination in Figure~\ref{fig:hallucination} and define them as follows:
\begin{enumerate}[label=\emph{(\roman*)}, itemsep=2pt, topsep=2pt, parsep=0pt, leftmargin=1.75em]
    \vspace{-0.05in}
    \item \textbf{Perceptual hallucination.} The tokenizer's reconstruction of an observation already differs from the observation itself, before any dynamics rollout has been executed. Concretely, the encoder--decoder pair projects out-of-distribution scene structure onto the closest in-distribution exemplar in its learned latent ``vocabulary''. For example, an unseen maze layout might be reconstructed with the agent and goal in the correct positions but with the walls of an entirely \emph{different} layout seen during training. The dynamics head then rolls out against this corrupted scene as if it were ground-truth. This failure mode is a property of the frozen encoder--decoder pair alone and persists even at horizon $H{=}0$.
    \item \textbf{Action-marginalized (ignored) hallucination.} Conditional on a context, the predicted next latent is largely insensitive to the input action. The rollout is visually plausible but collapses onto an action-marginalized future, so the model behaves more like a video generator than a controllable world model. Operationally, we expose this mode by intervening on the action stream at evaluation time, \emph{e.g.} by randomly shuffling actions within a batch and measuring the resulting change in flow MSE; a model that hallucinates in this way is one whose flow MSE barely moves under the intervention.
    \item \textbf{Scene-diverging hallucination.} It is well understood that autoregressive rollouts accumulate compounding error as the prediction horizon increases. However, \emph{scene-diverging} hallucination is a very specific failure mode where physically implausible events (such as a ball teleporting back into play when scoring in Pong) are predicted. This type of hallucination is most frequent in states with poor data coverage.
\end{enumerate}
These three types of failure modes probe disjoint pieces of the model: the tokenizer, the action-conditioning of the dynamics model, and the multi-step accumulation of dynamics error.

\subsection{Detecting Hallucination}
\label{sec:detecting-hallucination}
Next, we develop \emph{three} distinct predictors designed to track model hallucination. Empirically \textbf{we find that these three metrics, although mechanistically distinct, are all strong predictors of hallucination events}. We view this convergence as a feature and not a redundancy, since three mechanistically distinct predictors agreeing with respect to hallucination events is in itself evidence that the underlying signal is real. None of them require any labels nor additional training which makes them especially suitable for runtime detection.

\textcolor{nhdpurple}{\textbf{Proposed predictor 1: tokenizer round-trip residual.}} $u_r = \|\hat z - \mathrm{Encode}(\mathrm{Decode}(\hat z))\|$, the latent-space residual of a single decode–encode round-trip of the dynamics-predicted next latent $\hat z$, targets perceptual hallucination by measuring the very symptom that defines it: a predicted latent whose decoded frame falls off the tokenizer's manifold, \emph{e.g.}\ a corrupted scene layout or a fabricated object, does not survive re-encoding and produces a large $u_r$.

\textcolor{nhdpurple}{\textbf{Proposed predictor 2: flow instability.}} $u_f$, the \emph{flow instability} of the dynamics head at a given $(\text{context}, \text{action})$ pair, measures how much the denoiser's clean-target prediction $\hat{x}_1$ moves between successive Euler integration substeps, averaged over the second half of substeps. A sharp, well-conditioned dynamics head converges quickly to a stable $\hat{x}_1$ (low $u_f$); a head whose conditioning provides little signal keeps oscillating across substeps (high $u_f$).

\textcolor{nhdpurple}{\textbf{Proposed predictor 3: inter-seed variance.}}
$u_s$, the \emph{inter-seed variance} of the next-latent prediction across $N$ independent denoising trajectories at fixed past and action, targets scene-diverging hallucination by measuring epistemic uncertainty across noise seeds: regions where seeds disagree are precisely those where multi-step rollouts will fan out.

\begin{figure}[t]
    \centering
    \includegraphics[width=0.1585\linewidth]{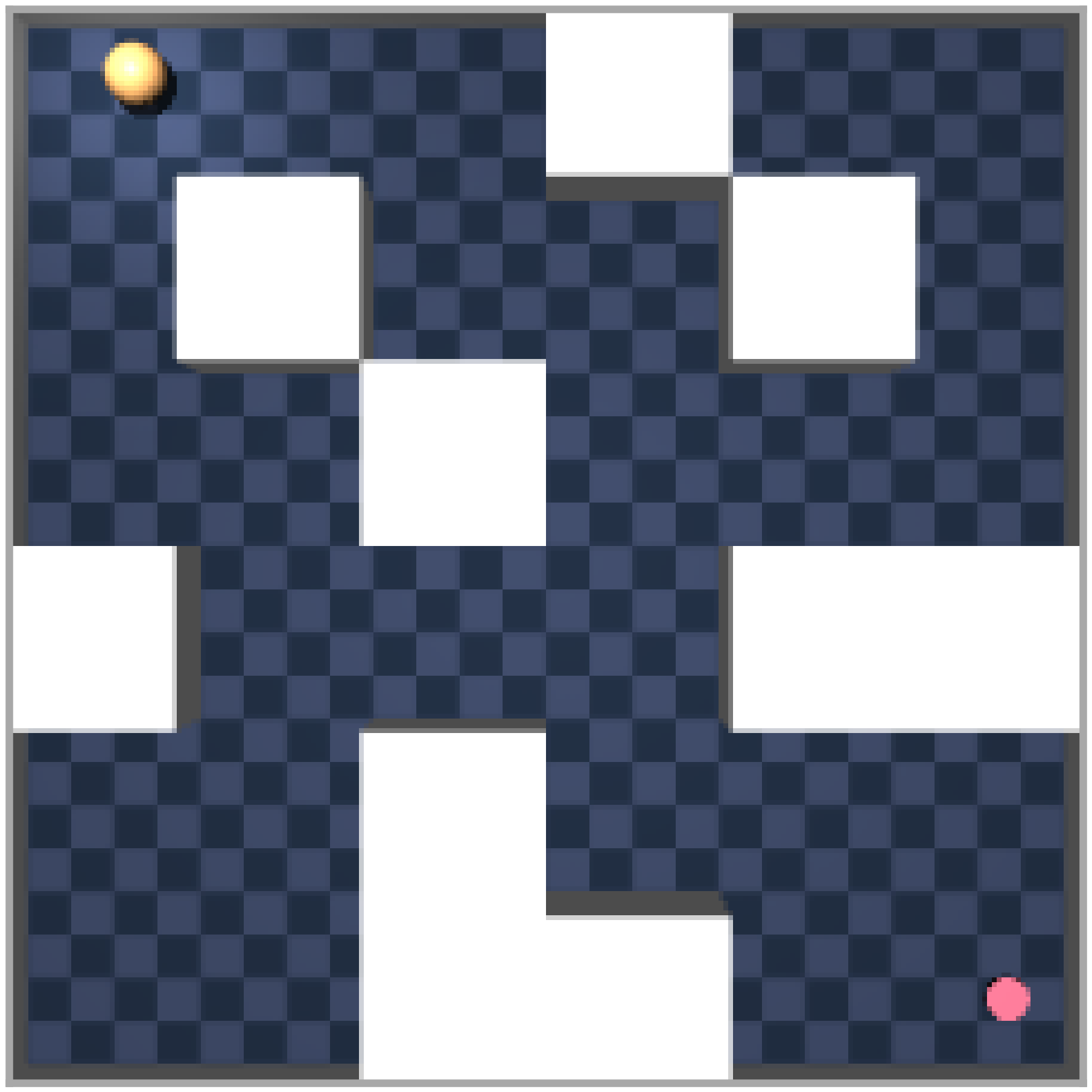}
    \includegraphics[width=0.1585\linewidth]{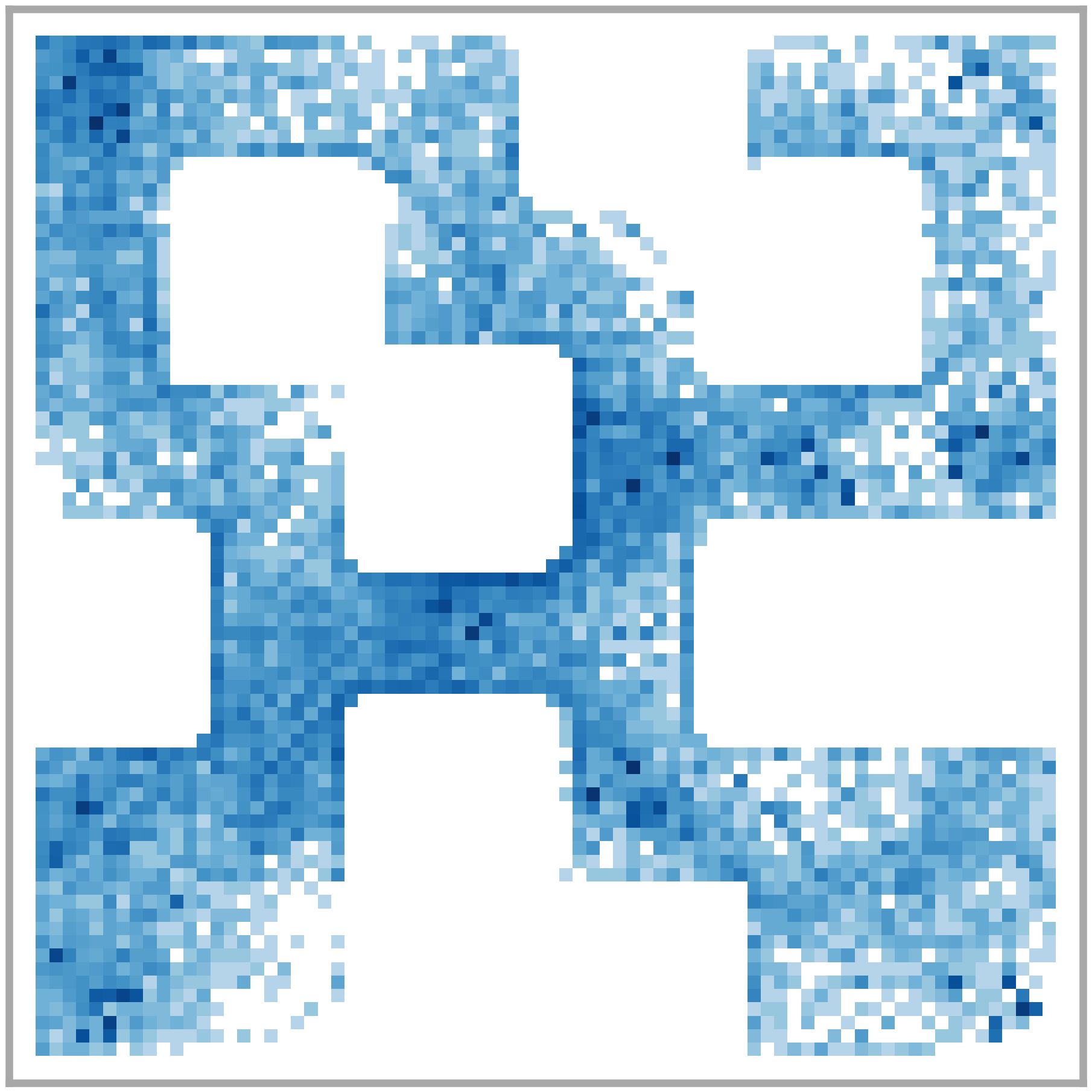}
    \includegraphics[width=0.1585\linewidth]{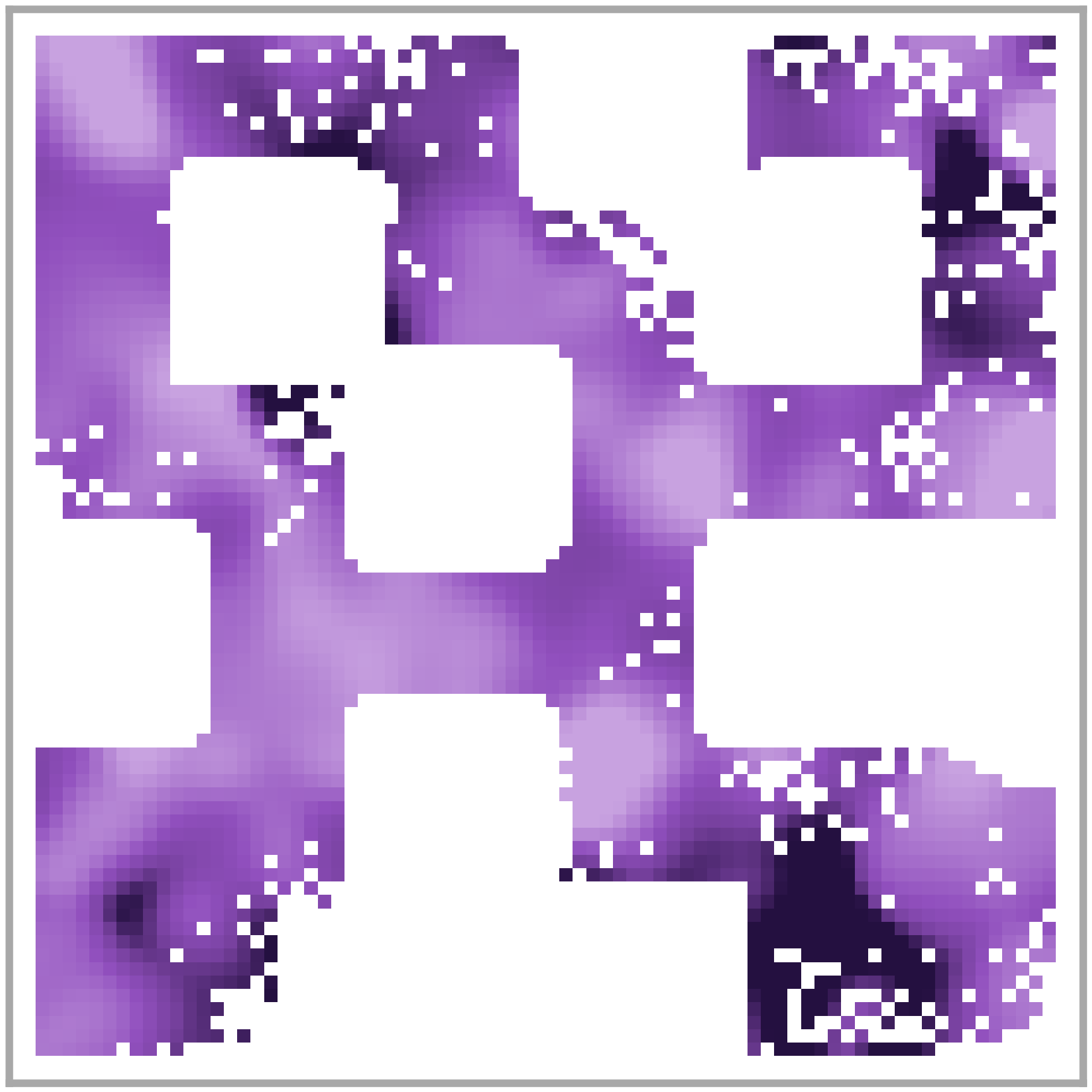}\quad
    \includegraphics[width=0.1585\linewidth]{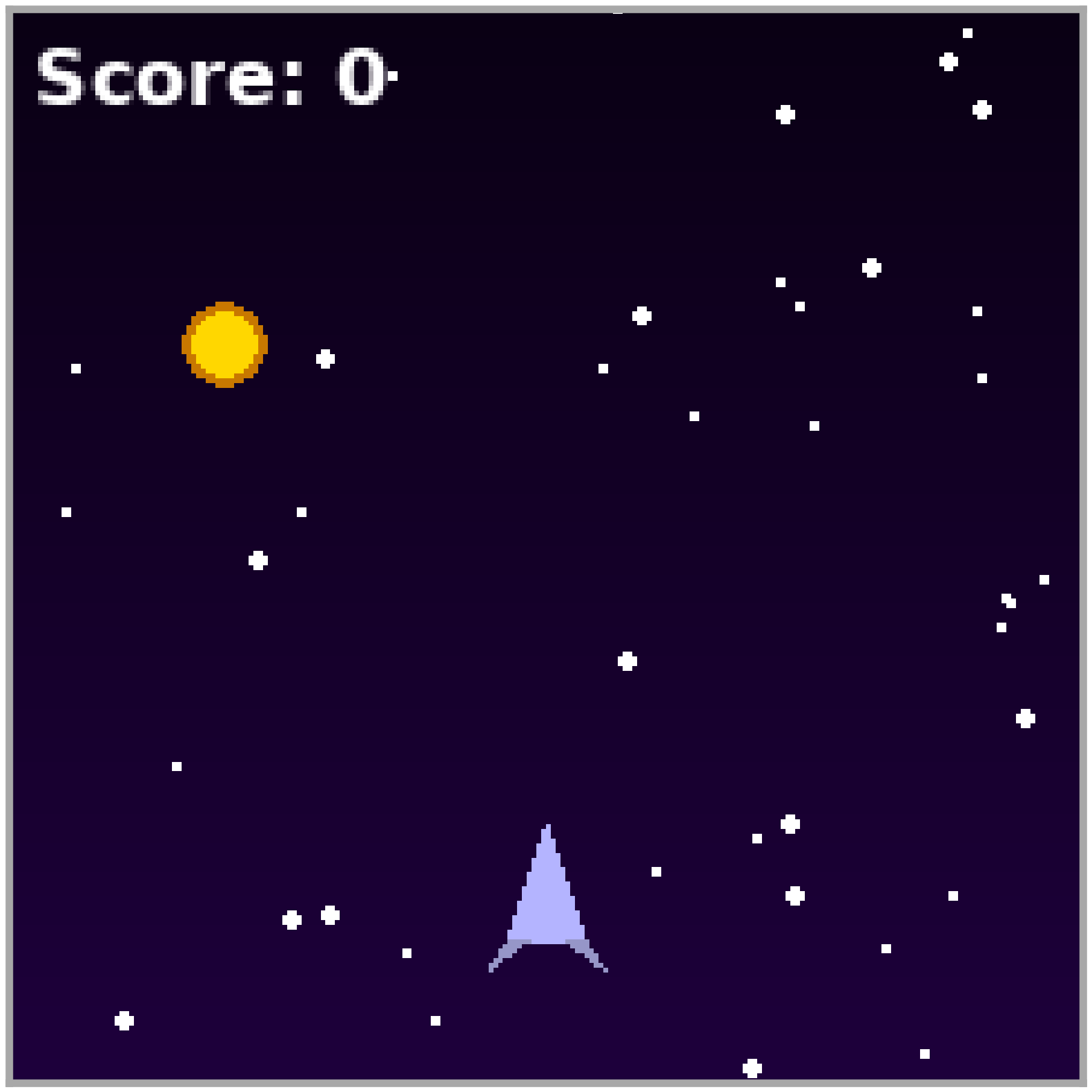}
    \includegraphics[width=0.1585\linewidth]{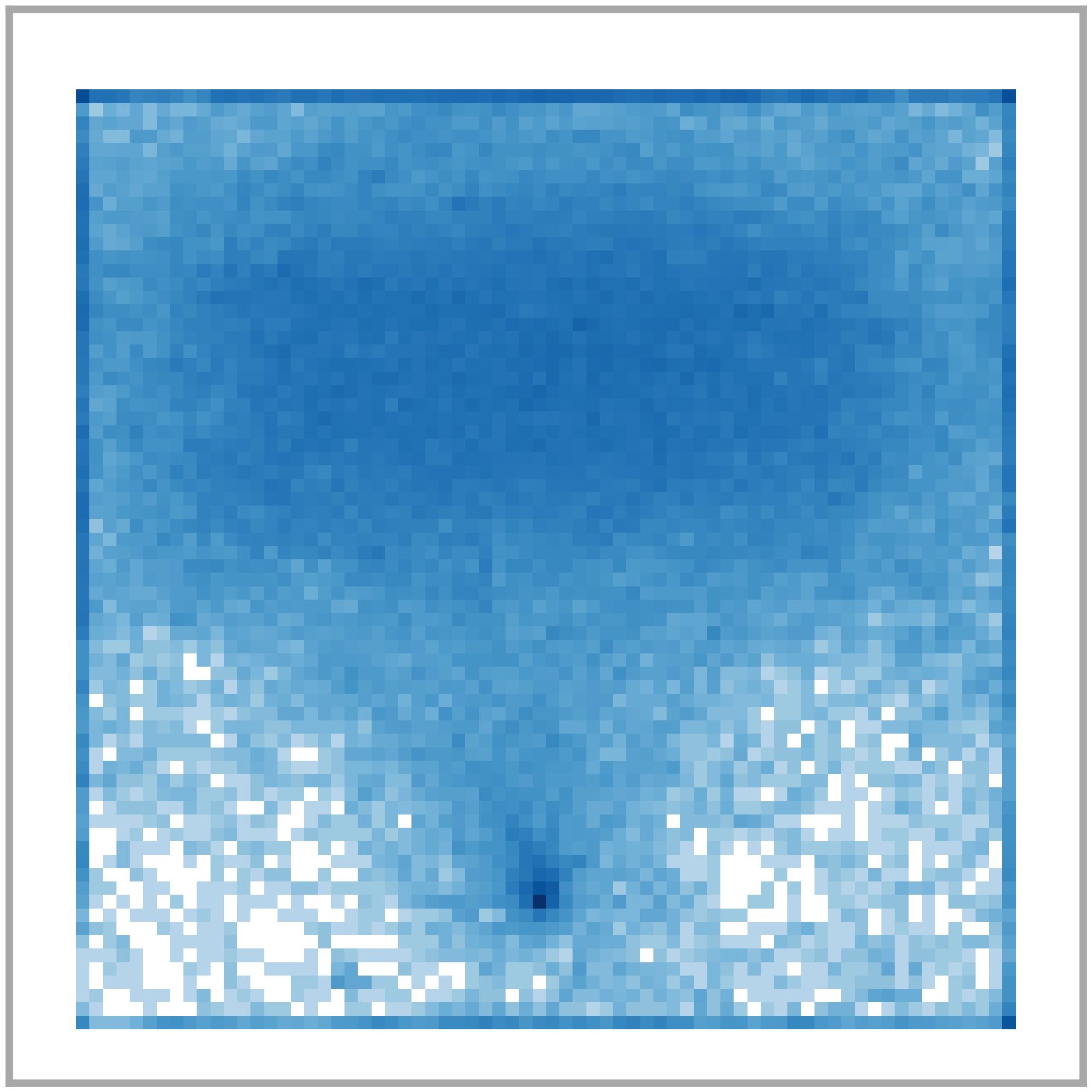}
    \includegraphics[width=0.1585\linewidth]{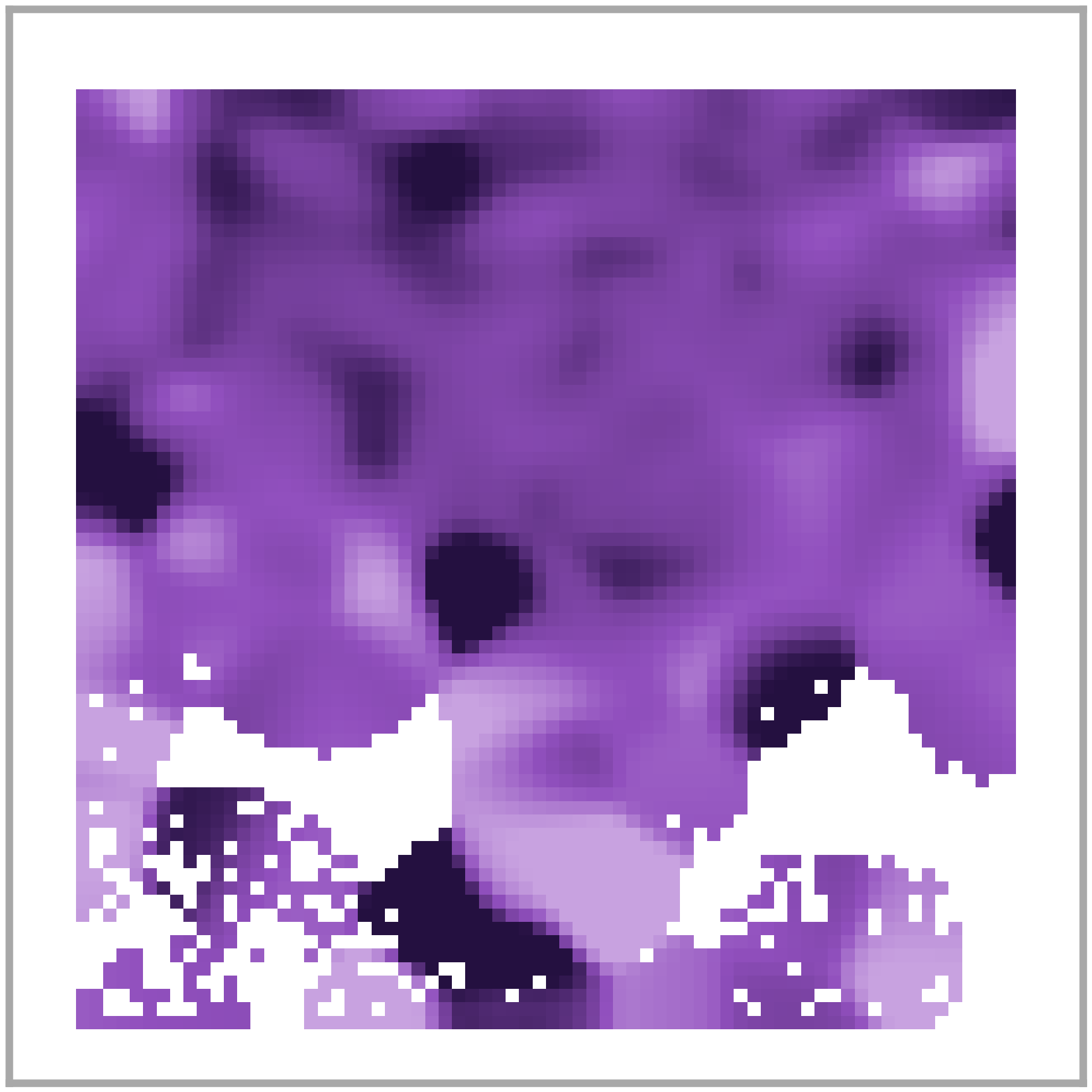}\\[0.1in]
    \includegraphics[width=0.1585\linewidth]{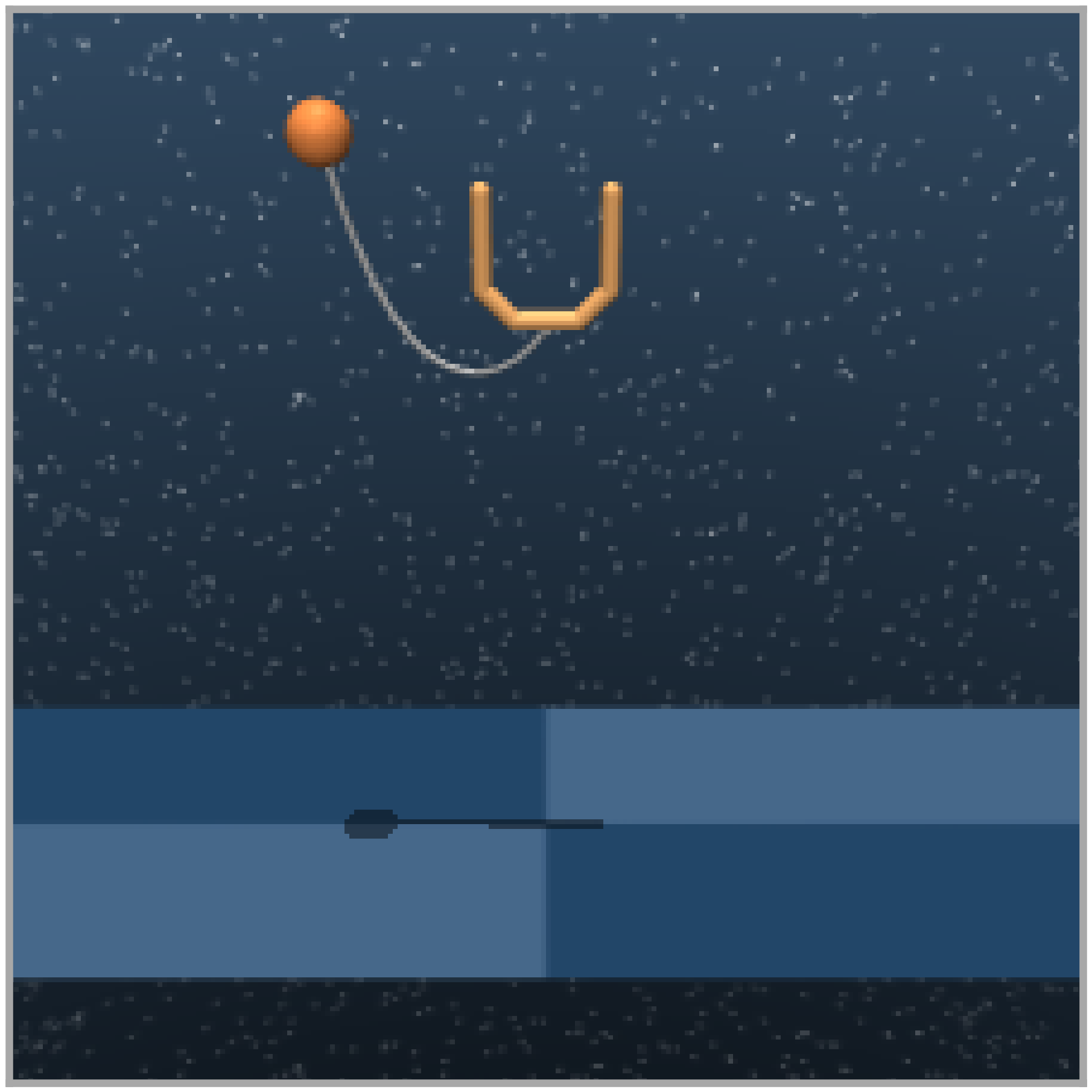}
    \includegraphics[width=0.1585\linewidth]{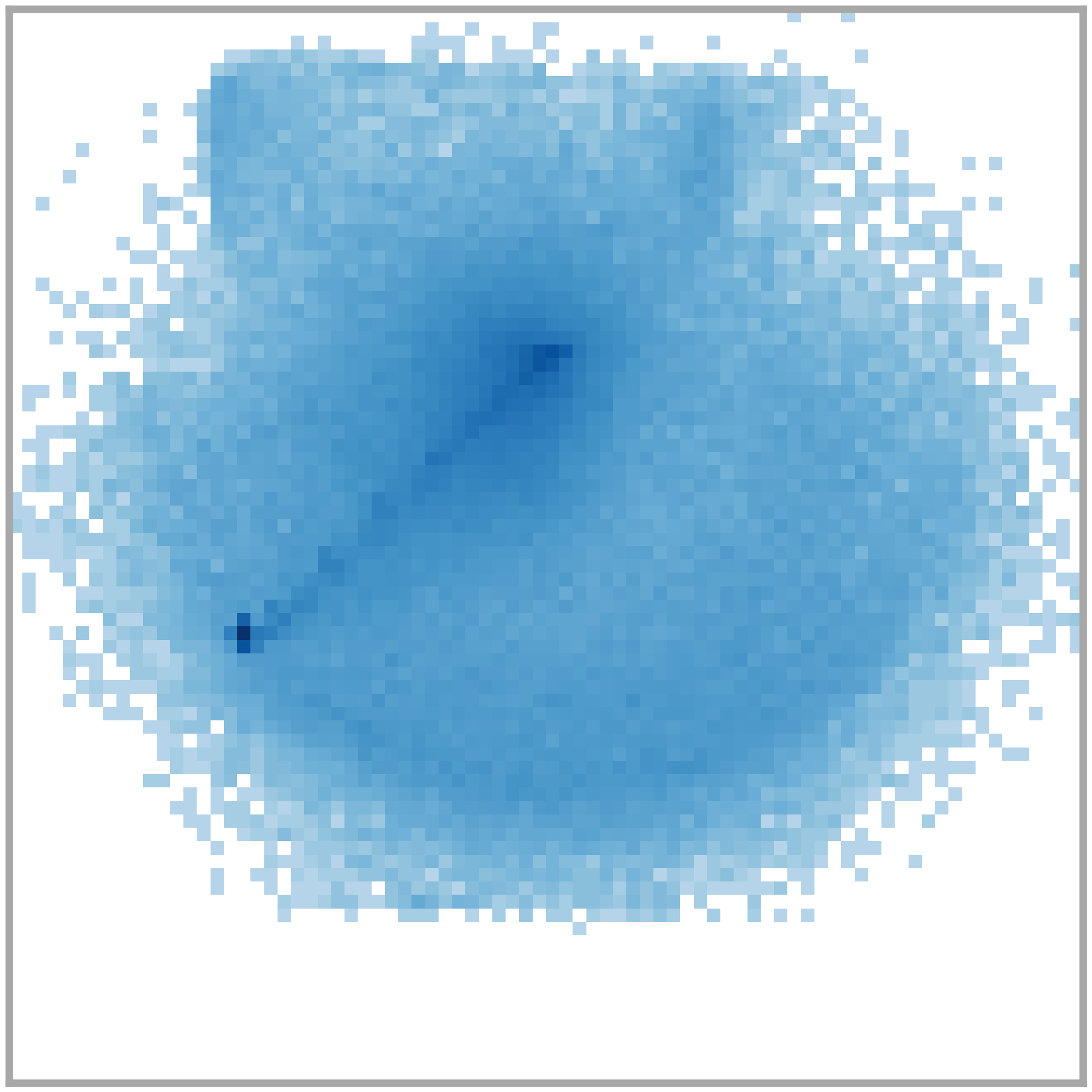}
    \includegraphics[width=0.1585\linewidth]{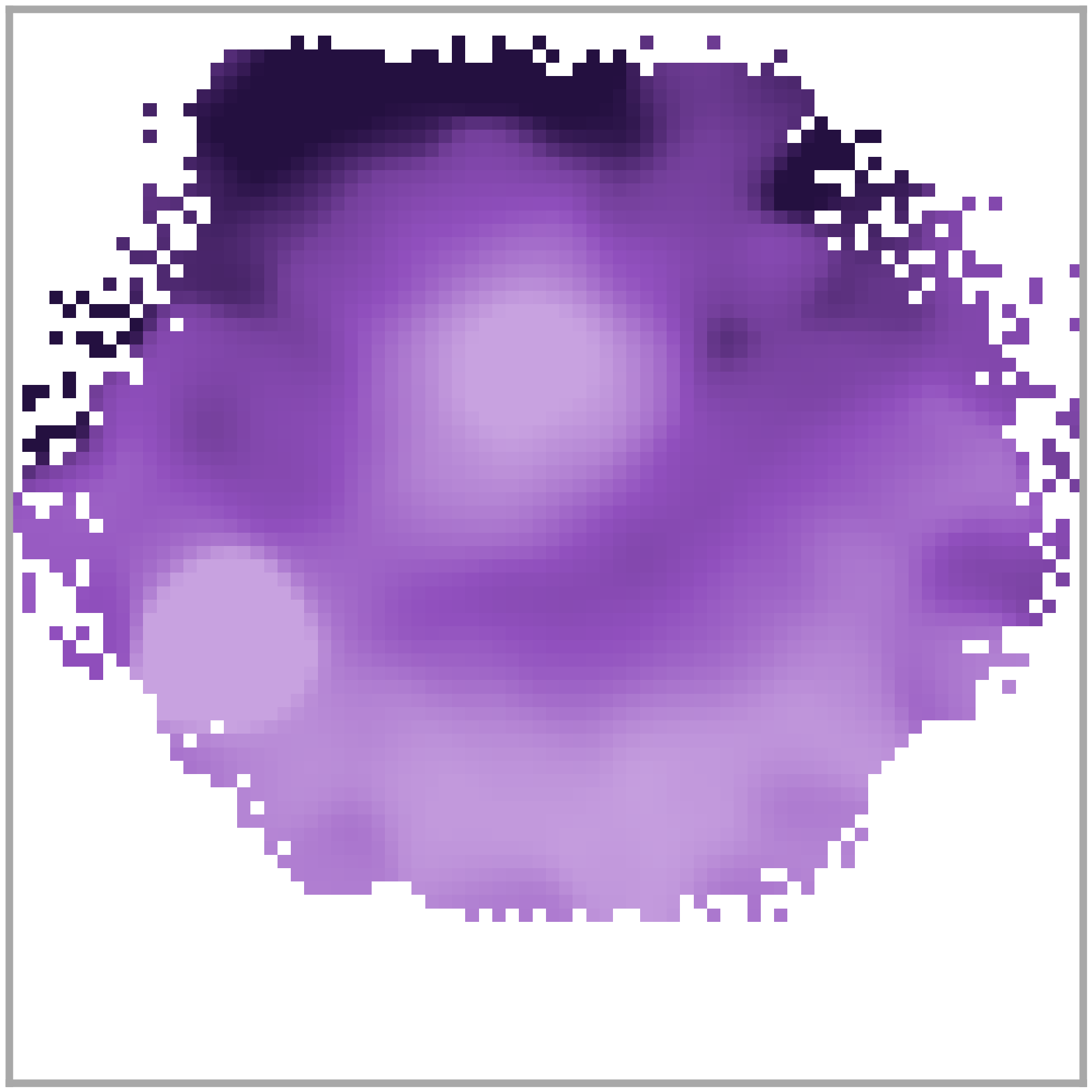}\quad
    \includegraphics[width=0.1585\linewidth]{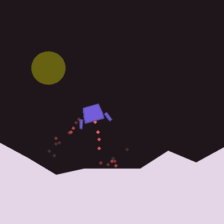}
    \includegraphics[width=0.1585\linewidth]{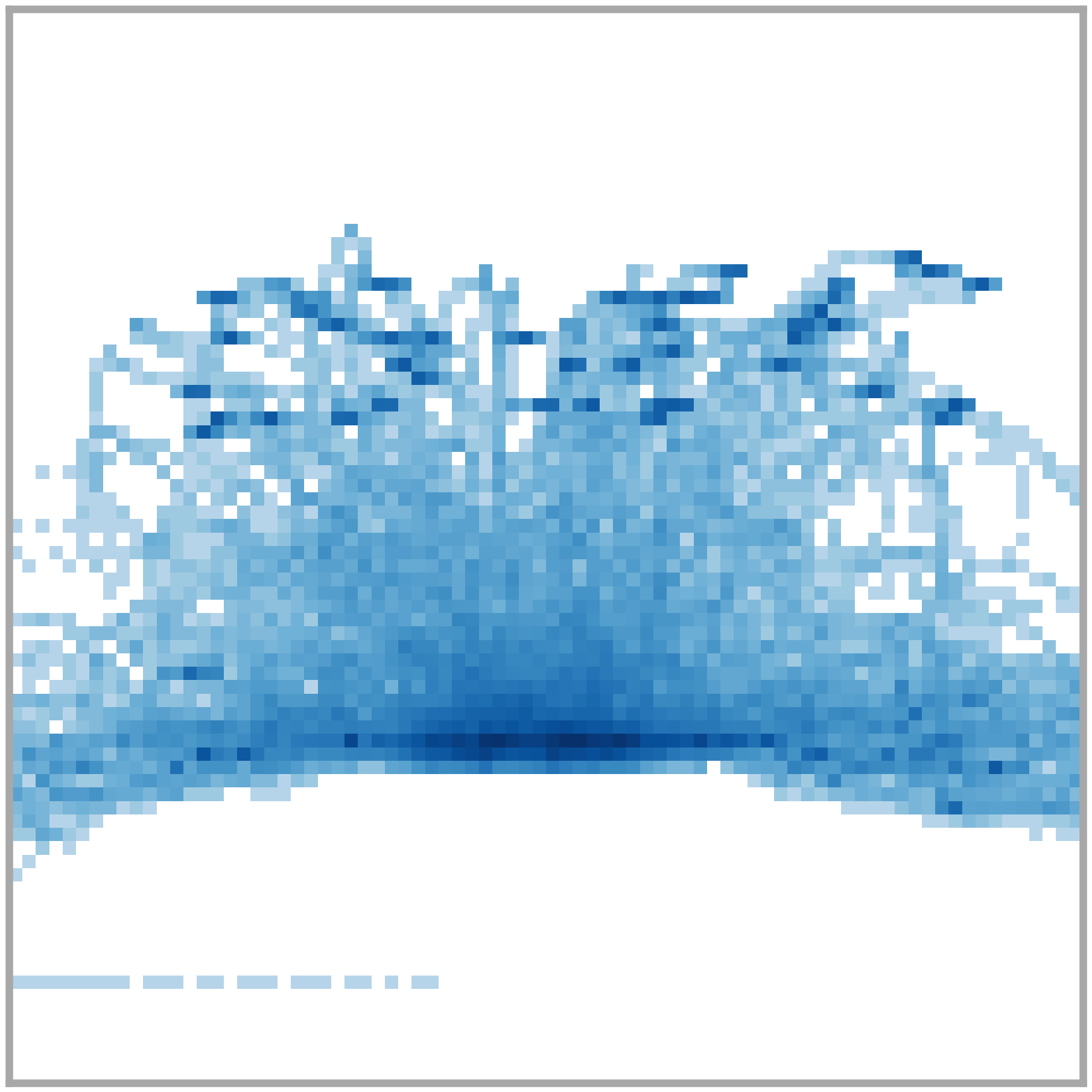}
    \includegraphics[width=0.1585\linewidth]{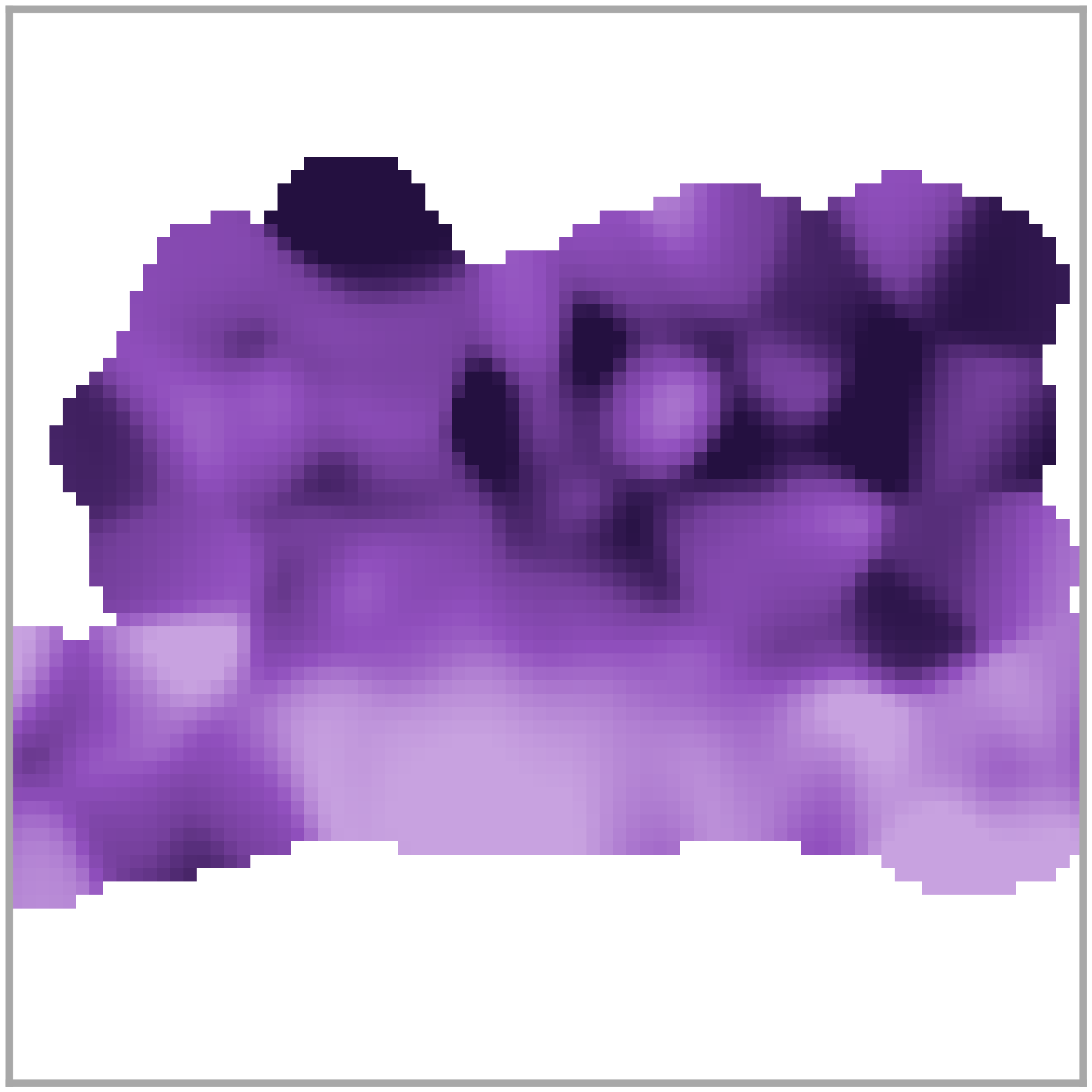}
    \caption{\textbf{Data coverage and hallucinations.} From left to right: sample frame, state density of key agent/object position, and tokenizer round-trip residual $u_r$ of the world model. We use \textbf{\textcolor{nhblue}{dark blue}} and \textbf{\textcolor{nhdpurple}{dark purple}} to denote \textbf{\textcolor{nhblue}{high state density}} and \textbf{\textcolor{nhdpurple}{high}} $\textcolor{nhdpurple}{\mathbf{u_r}}$, respectively. Hallucinations concentrate in low-coverage regions, especially near the periphery of the visited state distribution.}
    \label{fig:coverage}
    \vspace{-0.175in}
\end{figure}

\textbf{Controlling for scene motion.} In practice, we find that a naive use of the three signals above is confounded by scene activity: high-motion transitions inflate tokenizer residuals, flow instability, and seed-to-seed denoising dispersion alike. To remove this confound we instead report dynamism-normalized variants $u^{\text{norm}} \doteq u/m$ where $m$ is the per-step RMS change of the latent representation, computed as a per-task average over the dataset, or as a running estimate when used online for data collection. Normalizing by scene motion means that each predictor tracks model uncertainty \emph{relative to how much is happening in the scene}. We treat $u_{r}^{\text{norm}}, u_{f}^{\text{norm}}, u_{s}^{\text{norm}}$ as our primary hallucination predictors in all subsequent analysis.

\subsection{Mitigating Hallucination}
\label{sec:mitigating-hallucination}

The taxonomy and predictors above suggest a single data-centric lens on
hallucination: each of the three failure modes are, mechanistically, a
consequence of the model having seen too little of some region of the state-action space. A perceptual hallucination is a coverage gap in the
tokenizer's reconstruction distribution, an action-marginalized hallucination
is a coverage gap in action-conditional transitions, and a scene-diverging
hallucination is a coverage gap along the trajectory the model is asked to
imagine. Figure~\ref{fig:coverage} shows the relationship between data coverage and hallucination on four of our tasks. Two interventions follow naturally:

\textbf{Coverage-aware training.} We propose to resample the existing dataset
to upweight under-represented regions of the state-action space, then ask
whether closing those gaps at training time reduces all three failure
modes simultaneously. Because the lens above identifies coverage as the
underlying lever for every type of hallucination, a single reweighting recipe is expected to move all three signals in the right direction at once, rather than requiring a separate intervention per type. In practice, we rebalance training data by adjusting sampling to be uniform across \emph{tasks} rather than \emph{frames}. We also experiment with loss re-weighting but find interventions in sampling to be superior.

\textbf{Targeted data collection.} When the existing dataset does not
have enough coverage of a region for re-weighting alone to help, the hallucination predictors can themselves act as an objective for targeted, curiosity-driven data collection. Specifically, during interaction with a live environment, candidate trajectories are rolled out in the world model, scored by predicted hallucination, and the highest-ranked trajectory is executed in the environment, producing data that, by construction, covers transitions that previously caused the model to hallucinate. In practice, we collect data in a closed-loop manner with a prediction horizon of $H=32$ and replanning every $K=16$ steps.

\section{Experiments}
\label{sec:experiments}

We train a 350M parameter action-conditioned world model on the training corpus of MMBench2. Our training data consists of approximately 20M frames at $224\times224$ resolution across 200 distinct continuous control tasks, and we reserve the remaining $3$M frames for testing. Although low-dimensional state information is readily available in MMBench2, we choose to strictly consider RGB observations in this work. We describe our experimental setup below.

\textbf{Evaluation.} We evaluate our world model and its variants across four successive training phases: \emph{(1)} action-conditioned pretraining \emph{without} reward labels, \emph{(2)} coverage-aware "mid-training" that extends the previous phase with our proposed reweighted sampling, \emph{(3)} action-conditioned world modeling \emph{with} reward labels, and \emph{(4)} finetuning on seen and unseen tasks via targeted data collection. Our key evaluation metrics include:
\begin{enumerate}[label=\emph{(\roman*)}, itemsep=2pt, topsep=2pt, parsep=0pt, leftmargin=1.75em]
    \vspace{-0.05in}
    \item \textbf{Reconstruction PSNR} $\uparrow$ evaluates encoder/decoder quality without considering dynamics.
    \item \textbf{Rollout PSNR gain (dB)} $\uparrow$ evaluates quality of generated rollouts relative to a baseline that repeats the last frame across the entire rollout horizon. Although naive, this baseline can be surprisingly strong depending on the task. We consider a scene \emph{divergent} when $\Delta$PSNR $\le 0$.
    \item \textbf{Action shuffle ratio} $\uparrow$ evaluates action sensitivity in the dynamics model by measuring one-step teacher-forced flow MSE relative to batch-shuffled actions. We consider actions to be \emph{ignored} (marginalized) when this ratio $\le 1.1$.
    \item \textbf{Downstream task performance (normalized score)} $\uparrow$ evaluates how useful the world model is for downstream tasks via closed-loop MPC using the Cross-Entropy Method (CEM) with horizon $H=32$ and replanning every $16$ steps. We report normalized score $s \in [0, 1]$ rather than raw reward since reward scales differ significantly (by several orders of magnitude) across tasks.

\end{enumerate}

\textbf{Implementation details.} The encoder and decoder each have 50M learnable parameters, and the dynamics model (including prediction heads) has 250M parameters. We use $8\times$ NVIDIA H100 GPUs for training. The tokenizer is pretrained for $300$k steps ($14$ GPU days), and the dynamics model is pretrained for $180$k steps ($24$ GPU days). Our final checkpoints (post-finetuning) are trained for $380$k and $210$k steps, respectively, for a total of $58$ GPU days using a context length of $T=24$. The reward head and BC policy are conditioned on CLIP-ViT/B embeddings of per-task language instructions. See Appendix~\ref{sec:appendix-implementation-details} for more implementation details.

We seek to answer the following questions:
\begin{enumerate}[label=\emph{(Q\arabic*)}, itemsep=2pt, topsep=2pt, parsep=0pt, leftmargin=2.5em]
    \vspace{-0.05in}
    \item \textbf{Hallucination detection.} Do our proposed metrics accurately predict model hallucination?
    \item \textbf{Pretraining.} Does coverage-aware training meaningfully reduce hallucination?
    \item \textbf{Targeted data collection.} Can model hallucination be mitigated by collecting additional data? Given a fixed budget, which type of data improves the world model the most? 
    \item \textbf{Generalization.} How can a pretrained world model be adapted to unseen tasks? Does our world model generalize zero-shot or is finetuning necessary? When does it fail to transfer?
\end{enumerate}

\begin{figure}
    \centering
    \includegraphics[width=0.975\linewidth]{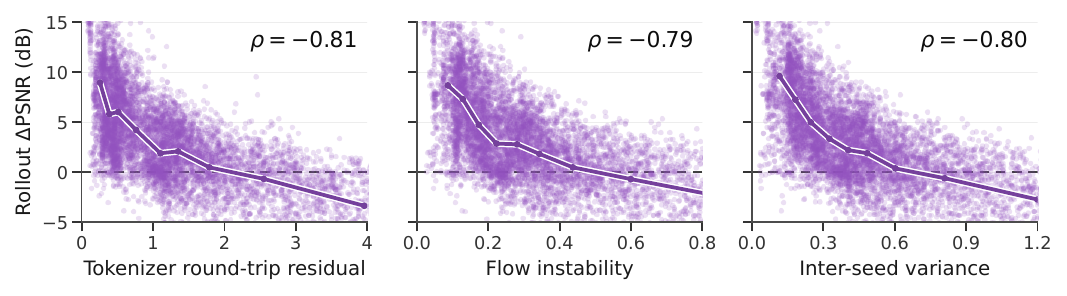}
    \vspace{-0.1in}
    \caption{\textbf{Our three hallucination predictors track the same realized rollout error.} Each point corresponds to one of $9$k held-out 24-frame sequences from any of 200 training tasks, computed using our pretrained base model; the purple curve shows the median for each of 8 bins.}
    \label{fig:hallucination-calibration}
    \vspace{-0.175in}
\end{figure}

\subsection{Main results}
\label{sec:experiments-results}
\vspace{-0.05in}

\textbf{Detecting hallucination events.} To determine whether our proposed hallucination metrics are predictive of hallucination and by extension model performance, we compute Spearman correlations between open-loop rollout $\Delta$PSNR and each of our predictors; results are shown in Figure~\ref{fig:hallucination-calibration}. We find a strong negative correlation ($\rho\approx0.80$) across all three predictors, which indicates that they all track the same underlying error. We additionally measure their per-task aggregated AUROC against two binary hallucination labels, \emph{action ignored} (action shuffle ratio $\le 1.1$) and \emph{scene divergent} (rollout $\Delta$PSNR vs. frame-repeating baseline $\le 0$); results are shown in Appendix~\ref{sec:appendix-additional-results}. We find that our predictors consistently outperform their unnormalized counterparts (\emph{i.e.}, $u_f^{\text{norm}}$ is a better predictor than $u_f$) as well as baselines that rely on latent scene motion $m$ or per-task frame count.

\begin{wraptable}[16]{r}{0.6\textwidth}
\centering
\vspace{-0.025in}
\caption{\textbf{Coverage-aware training, by stage.} Mean change with coverage-aware training vs. the base model on held-out trajectories across 200 tasks. \emph{Tok ft} finetunes the tokenizer for $30$k steps using coverage-aware training and then finetunes the dynamics model for $30$k steps using default sampling, \emph{Dyn ft} reverses this, and \emph{Both} finetunes both for $30$k steps each using coverage-aware training. Best is in bold.}
\label{tab:coverage-aware-pretraining}
\vspace{-0.05in}
\begin{tabular}{lrr>{\columncolor{nhpurple!20}}r}
\toprule
Metric & Tok ft & Dyn ft & \textbf{Both} \\
\midrule
Recon PSNR (dB) $\uparrow$       & $\mathbf{+0.46}$ & $-0.01$          & $+0.44$ \\
Action-shuffle ratio $\uparrow$      & $+0.02$          & $\mathbf{+0.27}$ & $+0.29$ \\
Rollout $\Delta$PSNR (dB) $\uparrow$    & $+0.42$          & $+0.68$          & $\mathbf{+0.88}$ \\
$u_r^{\text{norm}}$ $\downarrow$     & $-0.07$          & $-0.16$          & $\mathbf{-0.20}$ \\
$u_f^{\text{norm}}$ $\downarrow$     & $-0.03$          & $-0.06$          & $\mathbf{-0.07}$ \\
$u_s^{\text{norm}}$ $\downarrow$     & $-0.06$          & $-0.13$          & $\mathbf{-0.14}$ \\
\bottomrule
\end{tabular}
\end{wraptable}

\textbf{Coverage-aware training.} We evaluate the efficacy of our proposed coverage-aware training by extending the pretraining of our base model by an additional $30$k steps for the tokenizer and dynamics model each, while varying the sampling method: uniform sampling across \emph{frames} (default), and uniform sampling across \emph{tasks} (ours). Results are shown in Table~\ref{tab:coverage-aware-pretraining}. We find that tokenizer and dynamics model both benefit from coverage-aware training, and that adopting it for both achieves best overall performance.

\begin{figure}[t]
    \centering
    \begin{minipage}{0.49\textwidth}%
        \centering
        \begin{minipage}[b]{0.315\textwidth}%
            \centering
            \panellabel{Expert $\pi$}\\[0.1em]
            \includegraphics[width=\linewidth]{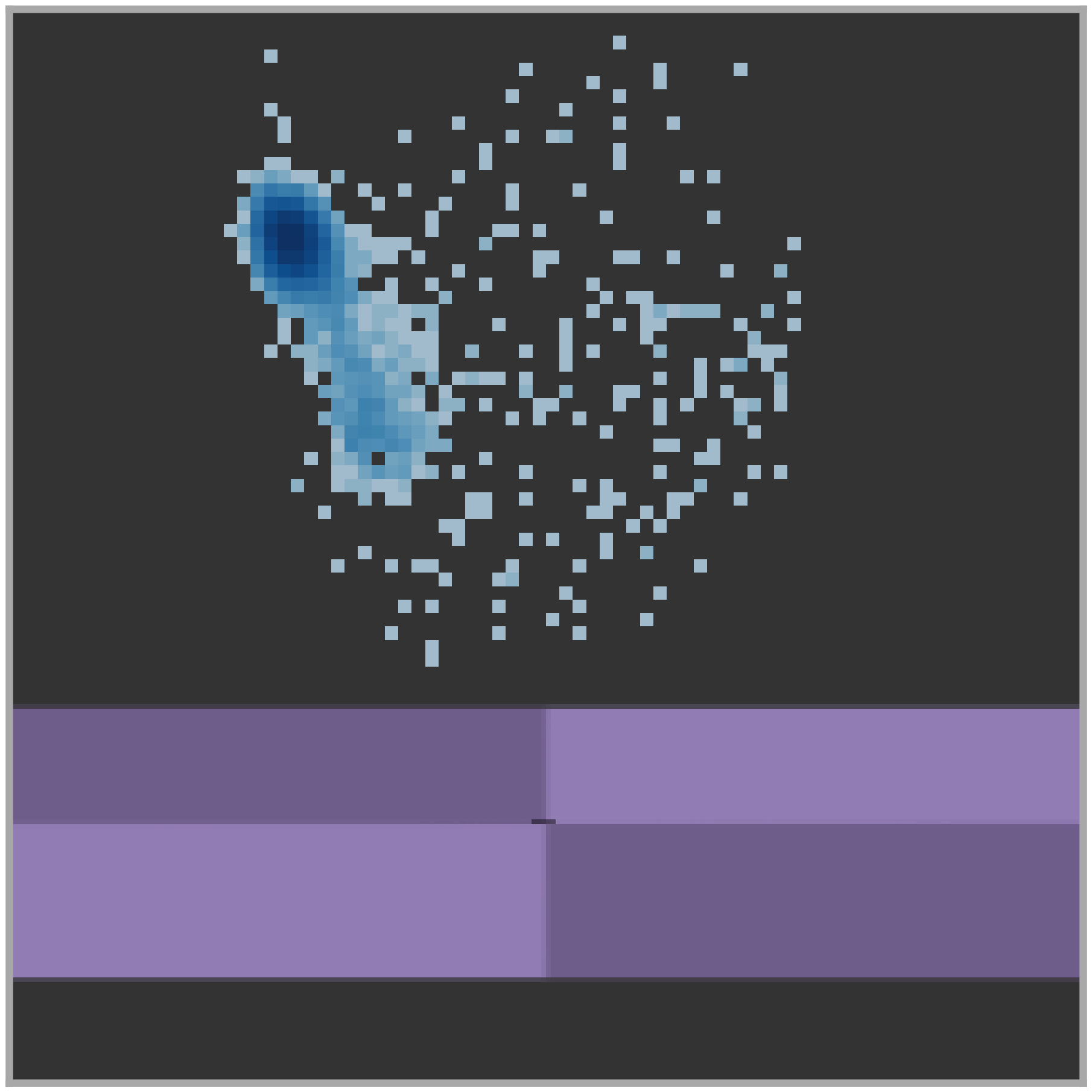}
        \end{minipage}
        \begin{minipage}[b]{0.315\textwidth}%
            \centering
            \panellabel{Curiosity}\\[0.1em]
            \includegraphics[width=\linewidth]{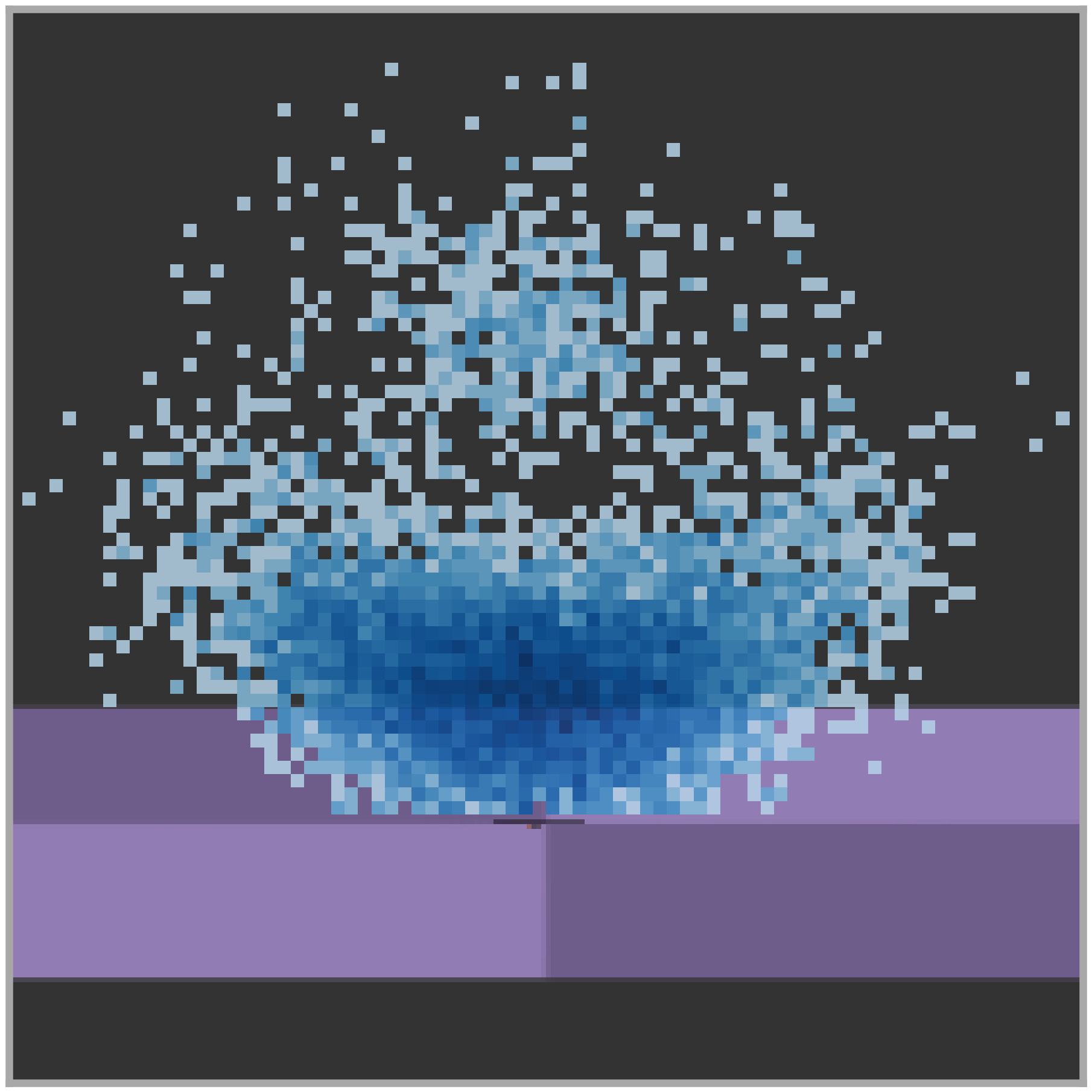}
        \end{minipage}
        \begin{minipage}[b]{0.315\textwidth}%
            \centering
            \panellabel{Human}\\[0.1em]
            \includegraphics[width=\linewidth]{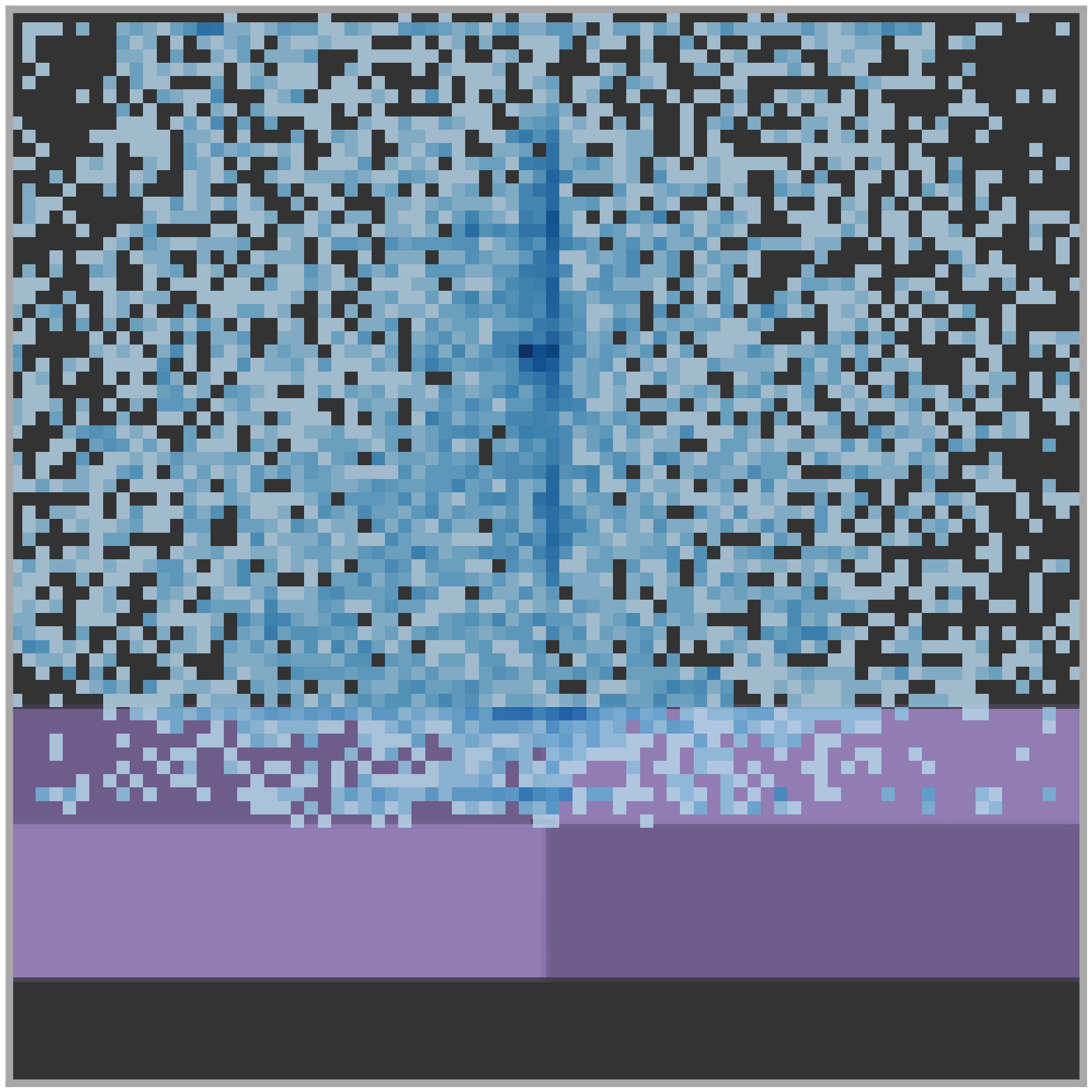}
        \end{minipage}\\[0.2em]
        Cup Catch (DMControl)
    \end{minipage}
    \begin{minipage}{0.49\textwidth}%
        \centering
        \begin{minipage}[b]{0.315\textwidth}%
            \centering
            \panellabel{Expert $\pi$}\\[0.1em]
            \includegraphics[width=\linewidth]{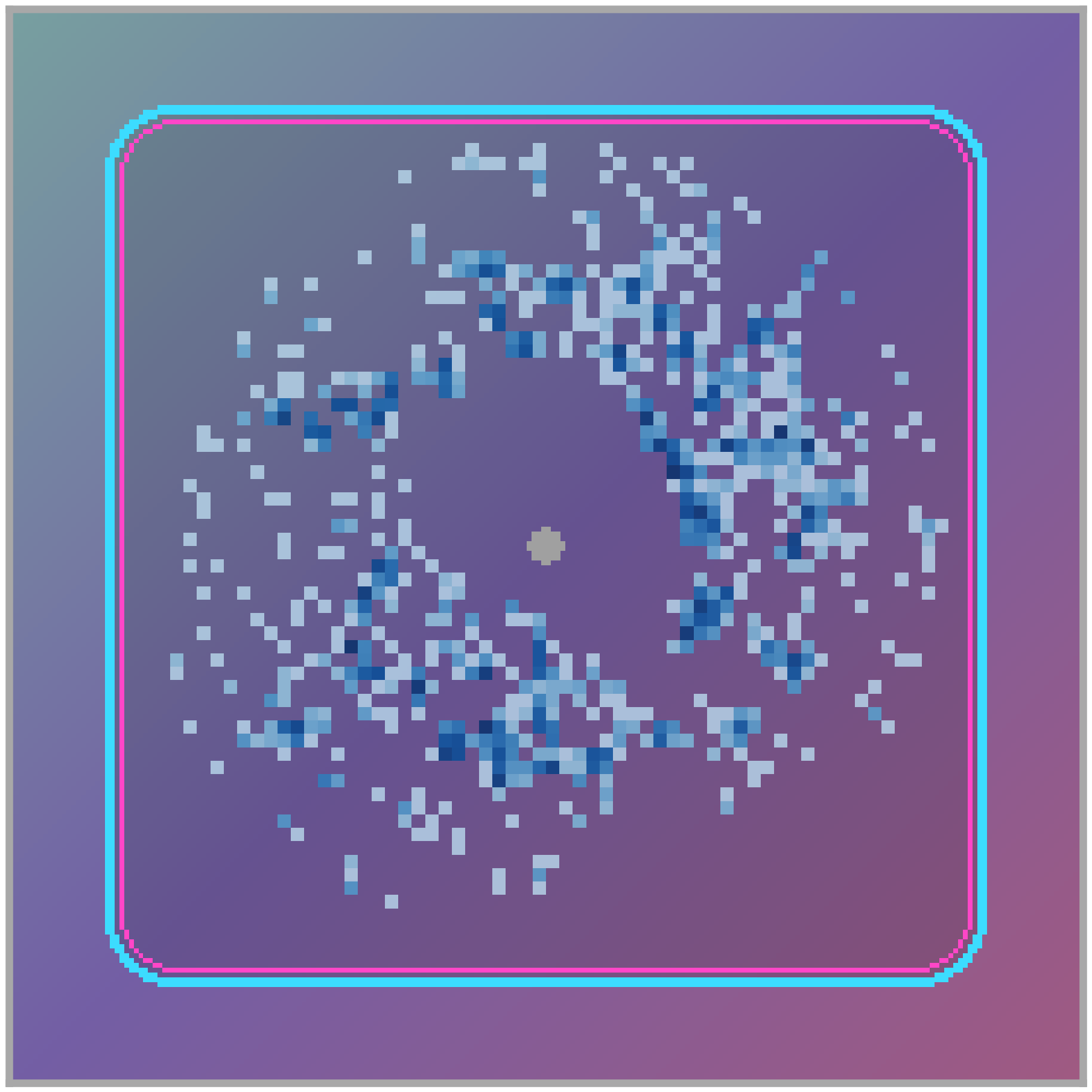}
        \end{minipage}
        \begin{minipage}[b]{0.315\textwidth}%
            \centering
            \panellabel{Curiosity}\\[0.1em]
            \includegraphics[width=\linewidth]{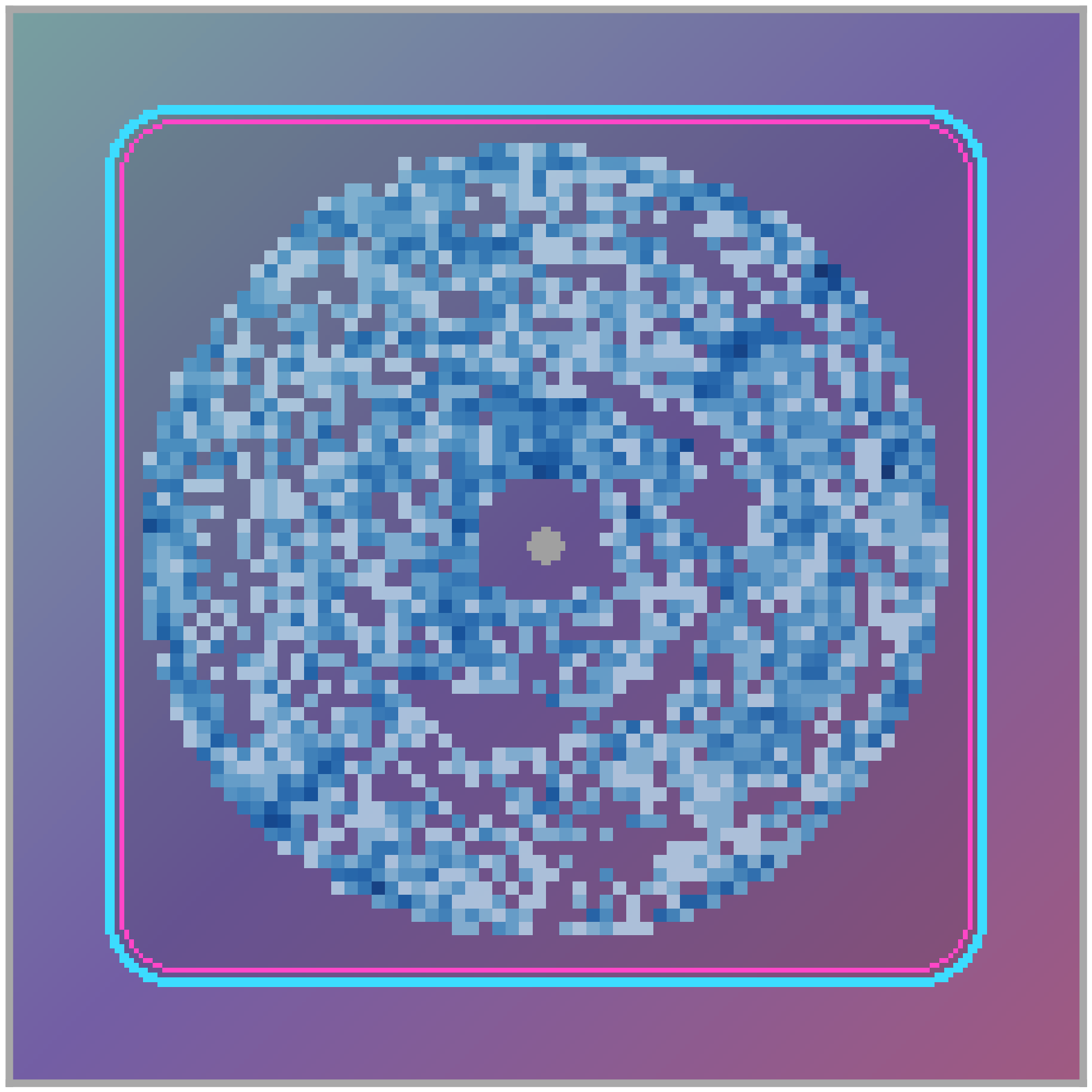}
        \end{minipage}
        \begin{minipage}[b]{0.315\textwidth}%
            \centering
            \panellabel{Human}\\[0.1em]
            \includegraphics[width=\linewidth]{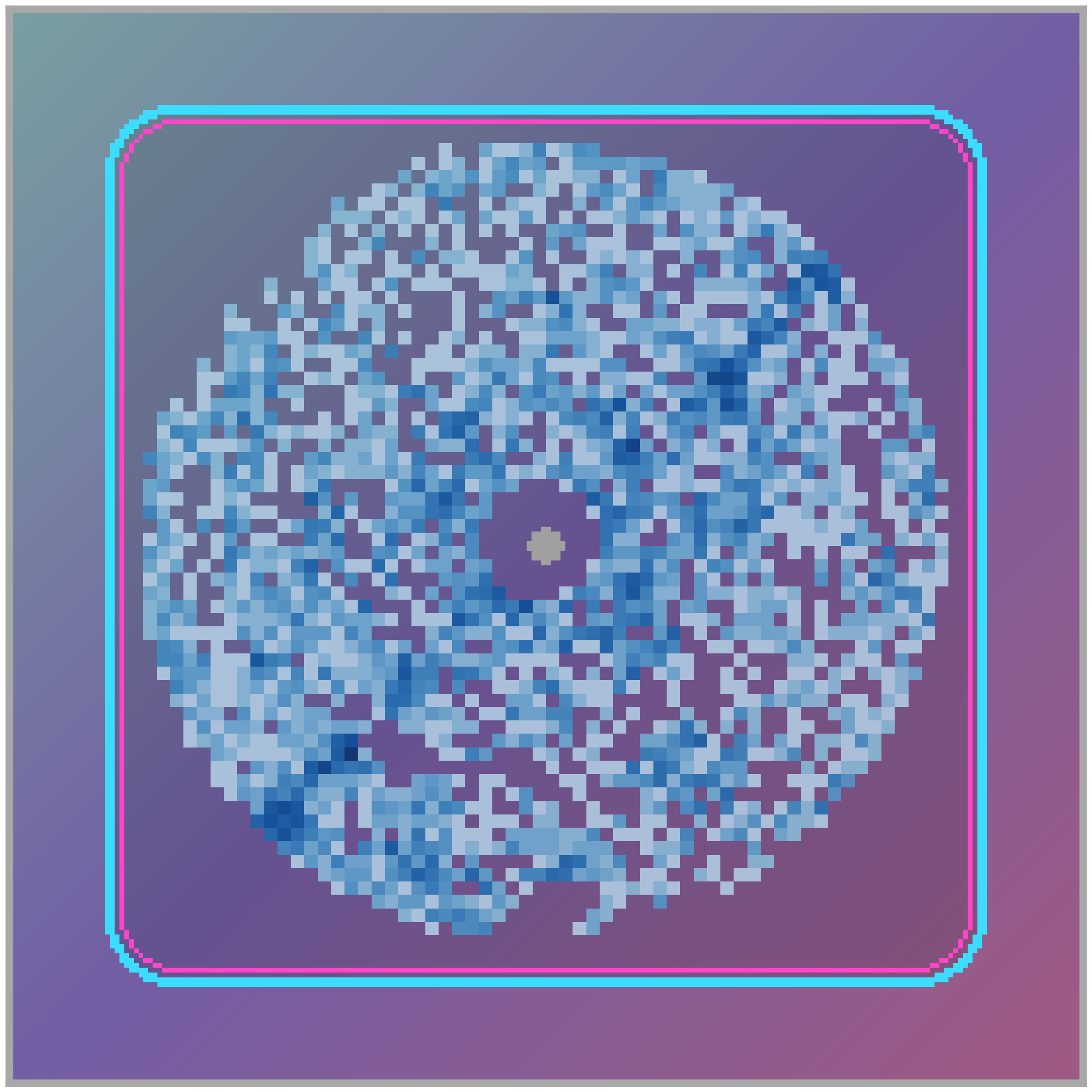}
        \end{minipage}\\[0.2em]
        Reacher Easy (MiniArcade)
    \end{minipage}
    \vspace{-0.025in}
    \caption{\textbf{Data coverage by collection method.} We show state densities for three online data collection policies (expert, curiosity, and human) across two tasks in the unseen task set.}
    \label{fig:data-collection-policies}
    \vspace{-0.175in}
\end{figure}

\begin{table}[t]
\centering
\small
\caption{\textbf{Targeted data collection for finetuning on 10 unseen tasks.} We finetune our world model on a set of 10 seen + 10 unseen tasks, varying data source and finetuning strategy. Each data source contains 50 trajectories per task. Offline metrics were computed using a test set of expert trajectories and human play data in equal amount. Task performance is measured via closed-loop MPC using CEM; we report mean performance normalized to $[0, 1]$ across 3 episodes per task.}
\label{tab:e8-unseen}
\vspace{0.075in}
\begin{tabular}{lcc@{\hskip 8pt}ccccc}
\toprule
Method & Tok FT & Dyn FT & \makecell{Recon\\PSNR $\uparrow$} & \makecell{Rollout\\$\Delta$PSNR $\uparrow$} & \makecell{Action\\shuf. $\uparrow$} & \makecell{$u_r^{\mathrm{norm}} \downarrow$} & \makecell{Task perf.\\(MPC) $\uparrow$} \\
\midrule
Random policy                                    &   &   & ---     & ---      & ---    & ---     & $0.118$ \\
Base                                             & \xmark & \xmark & $17.37$ & $-12.44$ & $1.12$ & $3.860$ & --- \\
Coverage-aware                                   & \xmark & \xmark & $17.21$ & $-12.52$ & $1.29$ & $3.769$ & $0.276$ \\
\midrule
No-op actions                                    & \xmark & \cmark & $17.21$ & $-11.66$ & $1.41$ & $4.175$ & --- \\
No-op actions                                    & \cmark & \cmark & $34.74$ & $+0.66$  & $1.55$ & $1.486$ & $0.163$ \\
Random policy                                    & \xmark & \cmark & $17.21$ & $-11.29$ & $1.73$ & $4.167$ & --- \\
Random policy                                    & \cmark & \cmark & $35.81$ & $+2.66$  & $2.00$ & $1.201$ & $0.228$ \\
Expert policy                                    & \cmark & \cmark & $35.86$ & $+2.84$  & $2.04$ & $1.131$ & $0.362$ \\
Human play                                       & \cmark & \cmark & $37.11$ & $+3.89$  & $\mathbf{2.42}$ & $1.002$ & $0.362$ \\
\rowcolor{nhpurple!20} Curiosity ($u_r^{\mathrm{norm}}$)       & \cmark & \cmark & $36.05$ & $+3.00$  & $2.00$ & $1.144$ & $0.325$ \\
\midrule
\textbf{All} (combined)                          & \cmark & \cmark & $\mathbf{37.91}$ & $\mathbf{+4.02}$ & $2.34$ & $\mathbf{0.975}$ & $\mathbf{0.390}$ \\
\bottomrule
\end{tabular}
\vspace{-0.15in}
\end{table}

\textbf{Mitigating hallucination via targeted data collection.} Since we observe a strong relationship between data coverage and hallucination, a question that naturally follows is \emph{can hallucination be mitigated by filling the gaps in coverage via targeted data collection?} To answer this question, we construct a task set that consists of 10 tasks seen during pretraining and 10 completely unseen tasks, and then collect 50 episodes for each of these tasks varying only the behavior policy used. Specifically, we evaluate the following approaches: no-op (all-zero) actions, random actions, expert policies, human play (via keyboard), and curiosity-based exploration \citep{sekar2020plan} using our proposed hallucination predictor $u_r^{\mathrm{norm}}$ as objective. State densities for different collection policies are shown in Figure~\ref{fig:data-collection-policies}. We finetune tokenizer and dynamics model for $50$k steps and $30$k steps, respectively, and report offline performance metrics and downstream task performance via closed-loop MPC (planning with CEM). Results for the 10 unseen tasks are shown in Table~\ref{tab:e8-unseen}, and additional results can be found in Appendix~\ref{sec:appendix-additional-results}. Pretraining transfers, to some extent, zero-shot ($0.276$, $2.3{\times}$ the $0.118$ random policy baseline), and collecting just $50$ trajectories per task using $u_r^{\mathrm{norm}}$-based curiosity lifts performance to $0.325$, within $\sim 90\%$ of the expert/human oracles ($0.362$) despite using no privileged behavior.

\subsection{Discussion}
\label{sec:experiments-discussion}

Empirically, we find that hallucination can be mitigated, to a great extent, by targeted data collection, but that not all data sources are equally informative. We believe that there are two concrete reasons for this: \emph{(i)} world modeling requires broad state-action coverage which, \emph{e.g.}, a random policy does not guarantee, and \emph{(ii)} downstream tasks are goal-directed and have a much narrower state-action distribution than the general space, so \emph{e.g.} our closed-loop MPC evaluations mainly measure model accuracy around that narrower subspace. That is, arguably, by design: there are certain behaviors we are more interested in modeling than others, but it raises a broader question about how we can best evaluate world models moving forward.

\textbf{Do off-the-shelf tokenizers improve perception?} One concrete way in which perceptual hallucination can be mitigated on unseen tasks is by leveraging off-the-shelf tokenizers trained on datasets several orders of magnitude larger than the one considered in this work. To investigate the viability of this approach, we compare our trained tokenizers (before and after finetuning on the unseen task set) against four off-the-shelf tokenizers; results are shown in Table~\ref{tab:tokenizers}. We find that off-the-shelf tokenizers such as Wan 2.1 VAE (the strongest) underperform compared to our in-domain tokenizer when evaluated on tasks from the $200$-task training corpus, but perform significantly better on the unseen task set unless we finetune our model in which case our in-domain tokenizer wins again. These results suggest that off-the-shelf tokenizers indeed are promising for world modeling but that there are still tangible benefits to in-domain finetuning when possible.

\textbf{Related work.} We include a detailed discussion of related work in Appendix~\ref{sec:related-work}, and position MMBench2 vs. existing datasets for world modeling in Appendix~\ref{sec:appendix-existing-datasets}.

\textbf{In conclusion,} we argue that hallucination in generative world models is, first and foremost, a data-coverage problem, and identify three failure modes: perceptual, action marginalization, and scene divergence. Our predictors track them with $\rho \approx 0.80$ against rollout $\Delta$PSNR and yield two complementary recipes: coverage-aware sampling reduces all three failure modes simultaneously, and using the predictors as curiosity rewards adapts our pretrained model to entirely unseen tasks.

\begin{table}[t]
\centering
\small
\caption{\textbf{Comparison to off-the-shelf tokenizers.} Reconstruction PSNR and LPIPS for our tokenizer as well as four off-the-shelf tokenizers, evaluated on the 10 ``seen'' and ``unseen'' task sets used in our finetuning experiments. Our tokenizers outperform the best off-the-shelf tokenizer, Wan 2.1 VAE, by a large margin on tasks in the training set but generalizes poorly to new tasks without additional finetuning. Best result in bold; arrows indicate whether higher ($\uparrow$) or lower ($\downarrow$) is better.}
\label{tab:tokenizers}
\vspace{0.05in}
\begin{tabular}{lrcrrrrr}
\toprule
 & & & \multicolumn{3}{c}{PSNR (dB) $\uparrow$} & \multicolumn{2}{c}{LPIPS $\downarrow$} \\
\cmidrule(lr){4-6} \cmidrule(lr){7-8}
Tokenizer & Params & Latent/frame & Seen & Unseen & $\Delta_{S-U}$ & Seen & Unseen \\
\midrule
\multicolumn{8}{l}{\emph{Ours}} \\
Base              & $102$\,M & $4096$ & $38.29$ & $17.34$ & $+20.95$ & $0.011$ & $0.389$ \\
Coverage-aware    & $102$\,M & $4096$ & $38.93$ & $17.12$ & $+21.81$ & $0.008$ & $0.348$ \\
\rowcolor{nhpurple!20} \textbf{Post-FT} & $102$\,M & $4096$ & $\mathbf{39.66}$ & $\mathbf{38.04}$ & $\mathbf{+1.62}$ & $\mathbf{0.007}$ & $\mathbf{0.010}$ \\
\midrule
\multicolumn{8}{l}{\emph{Off-the-shelf}} \\
SD-VAE-MSE                    & $84$\,M  & $3136$ & $33.32$ & $32.39$ & $+0.93$ & $0.031$ & $0.030$ \\
Cosmos-CV8x8x8 (1.0)          & $106$\,M   & $2048$ & $32.80$ & $32.72$ & $+0.08$ & $0.050$ & $0.042$ \\
Wan~2.1 VAE                   & $127$\,M & $4096$ & $36.45$ & $36.62$ & $-0.17$ & $0.010$ & $0.010$ \\
DC-AE-f32c32                  & $323$\,M & $2048$ & $31.49$ & $32.15$ & $-0.66$ & $0.035$ & $0.031$ \\
\bottomrule
\end{tabular}
\vspace{-0.1in}
\end{table}

\textbf{Limitations.} Our study operates at the $350$M parameter scale across $210$ simulated control tasks. Whether our findings translate to billion-parameter models or to real robot data with sensor noise and partial observability remains an open empirical question. We also acknowledge the significant computational cost of training large world models, as well as the need for more diverse datasets.

\clearpage
\newpage

\bibliographystyle{plainnat}
{\small
\bibliography{main}
}

\newpage

\appendix

\section{Related Work}
\label{sec:related-work}

\textbf{World models for control.} World models that learn environment dynamics from data have a long history in model-based RL, ranging from abstract latent dynamics models \citep{ha2018worldmodels, Hafner2020DreamTC, Hansen2022tdmpc, hansen2024tdmpc2} to high-capacity generative models that render full pixel observations \citep{micheli2023iris, alonso2024diffusion, valevski2024gamengen}. Recent work scales these models to heterogeneous video corpora \citep{bruce2024genie, agarwal2025cosmos, wan2025} and to real-time, playable neural environments \citep{decart2024oasis, lucid2024v1, deepmind2025genie3}. We build on Dreamer~4 \citep{hafner2025training}, an integrated tokenizer-plus-dynamics architecture with strong action conditioning, but the signals and interventions we propose are largely model-agnostic and apply, in principle, to any modern generative world model. Despite striking visual fidelity, these models continue to hallucinate under distribution shift --- a failure mode that we set out to characterize, predict, and mitigate.

\textbf{Hallucination and uncertainty in generative models.} Hallucination has been studied extensively in language models \citep{ji2023hallucination, huang2025hallucination} and, increasingly, in image \citep{li2023pope} and video generation \citep{huang2024vbench}, but the question of \emph{where} an autoregressive world model will fail along a rollout has received comparatively little attention. A separate but related literature studies uncertainty estimation in deep networks through deep ensembles \citep{lakshminarayanan2017simple}, MC dropout \citep{gal2016dropout}, and post-hoc out-of-distribution detectors \citep{lee2018mahalanobis, liu2020energy}, with applications in offline model-based RL via uncertainty-penalized policies \citep{yu2020mopo, kidambi2020morel}. Closest in spirit, prior work uses ensemble disagreement as an exploration signal in single-task RL \citep{pathak2019self, sekar2020plan}. In contrast, our proposed predictors are derived directly from the existing world model and target the three distinct stages at which a generative world model can hallucinate (encoder, dynamics, decoder), without auxiliary networks or labels.

\textbf{Coverage-aware training and data collection.} A growing body of work argues that data scale and composition are first-order levers for generative model performance \citep{hoffmann2022training, gadre2023datacomp}. In offline RL specifically, data coverage is known to bound policy improvement \citep{levine2020offline, kumar2020conservative}, motivating curated datasets such as ExoRL \citep{yarats2022don}, RL Unplugged \citep{gulcehre2020rl}, and V-D4RL \citep{lu2023vd4rl}. Curiosity-driven exploration is the natural online counterpart: agents are incentivized to visit states with high prediction error \citep{pathak2017curiosity}, high feature-network novelty \citep{burda2019rnd}, or high model disagreement \citep{pathak2019self}, scaled to imagined trajectories in Plan2Explore \citep{sekar2020plan}. Whereas prior curiosity work uses these signals to drive \emph{single-task policy exploration}, we adapt them to \emph{data-collection for generative world modeling} in two complementary ways: \emph{(i)} uniform-task resampling closes a substantial fraction of the hallucination gap at no additional data cost, and \emph{(ii)} using our proposed predictors as curiosity rewards yields a data-efficient finetuning recipe that generalizes a 350M-parameter world model to entirely unseen environments with as few as 50 trajectories from the target task.

\section{Comparison to Existing Datasets}
\label{sec:appendix-existing-datasets}

Table~\ref{tab:overview-datasets} compares MMBench2 to a representative set of prior datasets used for offline reinforcement learning, robot imitation learning, and large-scale generative video and world modeling. The offline RL datasets we compare against include RL Unplugged~\citep{gulcehre2020rl}, V-D4RL~\citep{lu2023vd4rl}, ExoRL~\citep{yarats2022don}, the TD-MPC2 multi-task dataset~\citep{hansen2024tdmpc2}, and the Atari DQN Replay dataset~\citep{agarwal2020optimistic}. The robot imitation learning datasets we compare against are RoboNet~\citep{dasari2019robonet}, BridgeData~V2~\citep{walke2023bridgedata}, DROID~\citep{khazatsky2024droid}, and Open X-Embodiment~\citep{open_x_embodiment_rt_x_2023}. For large-scale video pretraining corpora with pseudo-labeled actions, we compare against VPT~\citep{baker2022video} and NitroGen~\citep{magne2026nitrogen}. Finally, we include MMBench~\citep{Hansen2025Newt}, the multi-task benchmark on which MMBench2 builds. Across these datasets, MMBench2 is the only corpus that simultaneously offers ground-truth action \emph{and} reward labels, live simulators for every task, mixed-quality behavior, and broad coverage across both task domains and embodiments.

\begin{table}[h!]
    \centering
    \caption{\textbf{MMBench2 vs. existing datasets.} Our dataset consists of mixed-quality data spanning a larger number of tasks and domains, complete with ground-truth action and reward labels as well as live environments. We summarize key characteristics of MMBench2 and existing datasets below.\\ {\footnotesize\textsuperscript{a}Per-trajectory binary success flag only. \textsuperscript{b}Actions are pseudo-labels. \textsuperscript{c}Estimated from reported 40k hours.}}
    \vspace{0.075in}
    \label{tab:overview-datasets}
    \resizebox{\textwidth}{!}{%
    \begin{tabular}[b]{lcccccccccc}
    \toprule
    Dataset & Tasks & Domains & Trajs & Frames & Resolution & Action & Reward & Live env. & Behavior \\
    \midrule
    RL Unplugged            & 66          & 4          & ---          & 12B          & Varies    & \cmark              & \cmark              & \cmark & Replay      \\
    V-D4RL                     & 3           & 1          & ---          & 6M           & $64{\times}64$               & \cmark              & \cmark              & \cmark & Mixed       \\
    ExoRL                  & 11          & 1          & ---          & 45M          & State                        & \cmark              & \cmark & \cmark & Exploratory \\
    TD-MPC2               & 80          & 2          & ---          & 545M         & State                        & \cmark              & \cmark              & \cmark & Replay      \\
    Atari DQN Replay & 60          & 1          & ---          & 15B          & $84{\times}84$               & \cmark              & \cmark              & \cmark & Replay      \\
    RoboNet              & ---         & 1          & 162k         & 15M          & $128{\times}128$             & \cmark              & \xmark              & \xmark & Scripted    \\
    BridgeData V2        & 13          & 1          & 60k          & 2.3M         & $640{\times}480$             & \cmark              & \xmark              & \xmark & Demos       \\
    DROID               & 86          & 1          & 76k          & 19M          & $1280{\times}720$            & \cmark              & \pmark\textsuperscript{a} & \xmark & Demos       \\
    Open X-Embod.                  & 527         & 22         & 1M+          & ---          & Varies                       & \cmark              & \xmark              & \xmark & Demos       \\
    VPT                       & ---         & 1          & ---          & 5B           & $128{\times}128$             & \pmark\textsuperscript{b} & \xmark              & \cmark & Human       \\
    NitroGen                  & 1{,}000+    & ---        & 39k          & 4B\textsuperscript{c} & $256{\times}256$ & \pmark\textsuperscript{b} & \xmark              & \cmark & Human       \\
    MMBench                       & 200         & 10         & 4k           & 1.8M  & $224{\times}224$    & \cmark              & \cmark              & \cmark & Demos       \\
    \midrule
    \rowcolor{nhpurple!20} \textbf{MMBench2}                       & \textbf{210} & \textbf{10} & \textbf{65.6k} & \textbf{23M} & $\mathbf{224{\times}224}$    & \cmark              & \cmark              & \cmark & \textbf{Mixed}       \\

    \bottomrule
    \end{tabular}
    }
\end{table}

\section{Task Domains}
\label{sec:appendix-environments}

We consider \textbf{210} tasks across \textbf{10} domains. Our task set is comprised of diverse continuous control tasks spanning robot manipulation, locomotion, navigation, arcade games, and classic control problems, each varying in task complexity, time horizon, observation and action space dimensionality, and reward formulation. Table~\ref{tab:appendix-task-domains} provides an overview of our task domains.

\begin{table}[h!]
\centering
\parbox{\textwidth}{
\caption{\textbf{Overview of task domains.} Our dataset covers a wide range of task types, state and action dimensionalities, time horizons, and reward formulations. Table courtesy of \citet{Hansen2025Newt}.}
\label{tab:appendix-task-domains}
\vspace{0.05in}
\centering
\begin{tabular}{lccccccc}
\toprule
\textbf{Task domain} & \textbf{Tasks} & \textbf{Observation} & \multicolumn{2}{c}{\textbf{Action dim}} & \multicolumn{2}{c}{\textbf{Ep. length}} & \textbf{Reward} \\
& & & min & max & min & max & \\ \midrule
\addlinespace[0.6ex]\raisebox{-0.25\height}{\includegraphics[width=0.04\linewidth]{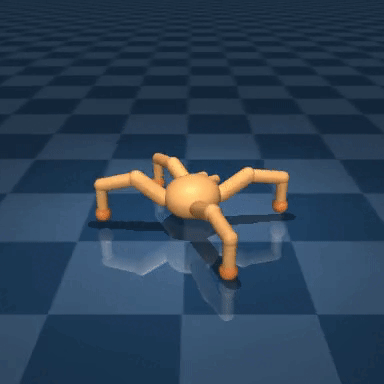}}~~DMControl & $23$ & $224\times224$ & $1$ & $12$ & $500$ & $500$ & dense/sparse \\
\addlinespace[0.6ex]\raisebox{-0.25\height}{\includegraphics[width=0.04\linewidth]{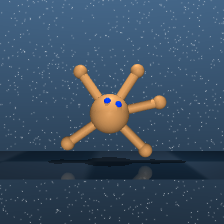}}~~DMControl Ext. & $16$ & $224\times224$ & $1$ & $7$ & $500$ & $500$ & dense/sparse \\
\addlinespace[0.6ex]\raisebox{-0.25\height}{\includegraphics[width=0.04\linewidth]{visualizations/domains/metaworld/mw-stick-pull.png}}~~Meta-World & $49$ & $224\times224$ & $4$ & $4$ & $100$ & $100$ & dense \\
\addlinespace[0.6ex]\raisebox{-0.25\height}{\includegraphics[width=0.04\linewidth]{visualizations/domains/maniskill/ms-poke-cube.png}}~~ManiSkill3 & $37$ & $224\times224$ & $1$ & $12$ & $25$ & $500$ & dense/sparse \\
\addlinespace[0.6ex]\raisebox{-0.25\height}{\includegraphics[width=0.04\linewidth]{visualizations/domains/mujoco/mujoco-walker.png}}~~MuJoCo & $6$ & $224\times224$ & $1$ & $8$ & $50$ & $1000$ & dense/sparse \\
\addlinespace[0.6ex]\raisebox{-0.25\height}{\includegraphics[width=0.04\linewidth]{visualizations/domains/pygame/pygame-coconut-dodge.png}}~~MiniArcade & $24$ & $224\times224$ & $1$ & $2$ & $200$ & $500$ & dense/sparse \\
\addlinespace[0.6ex]\raisebox{-0.25\height}{\includegraphics[width=0.04\linewidth]{visualizations/domains/box2d/bipedal-walker-obstacles.png}}~~Box2D & $8$ & $224\times224$ & $2$ & $4$ & $500$ & $500$ & dense \\
\addlinespace[0.6ex]\raisebox{-0.25\height}{\includegraphics[width=0.04\linewidth]{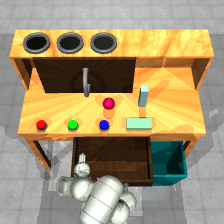}}~~RoboDesk & $6$ & $224\times224$ & $5$ & $5$ & $100$ & $100$ & dense \\
\addlinespace[0.6ex]\raisebox{-0.25\height}{\includegraphics[width=0.04\linewidth]{visualizations/domains/ogbench/og-antball.png}}~~OGBench & $14$ & $224\times224$ & $2$ & $8$ & $100$ & $1000$ & dense \\
\addlinespace[0.6ex]\raisebox{-0.25\height}{\includegraphics[width=0.04\linewidth]{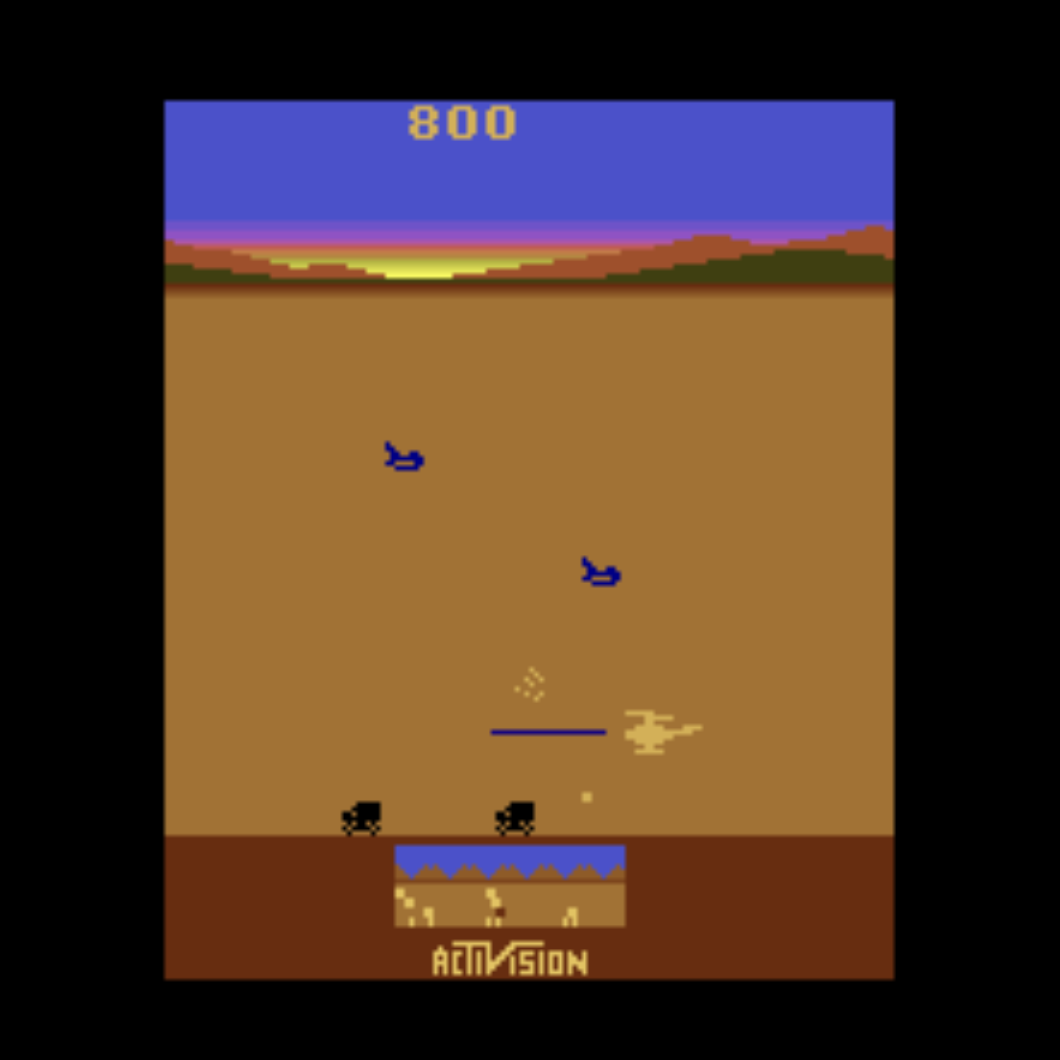}}~~Atari & $27$ & $224\times224$ & $3$ & $3$ & $1000$ & $1000$ & sparse \\\bottomrule
\end{tabular}%
}
\end{table}

\begin{figure}[h]
    \centering
    \begin{minipage}{\textwidth}%
        \centering
        DMControl\\[0.2em]
        \includegraphics[width=0.16\linewidth]{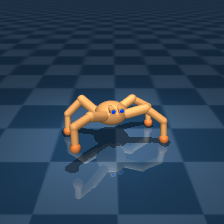}
        \includegraphics[width=0.16\linewidth]{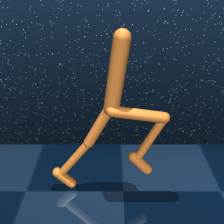}
        \includegraphics[width=0.16\linewidth]{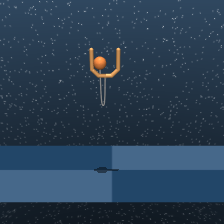}
        \includegraphics[width=0.16\linewidth]{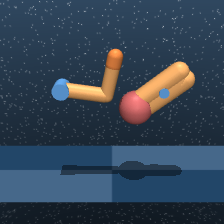}
        \includegraphics[width=0.16\linewidth]{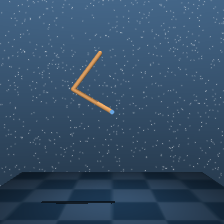}
        \includegraphics[width=0.16\linewidth]{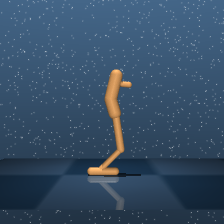}
    \end{minipage}\\[1em]
    \begin{minipage}{\textwidth}%
        \centering
        DMControl Extended\\[0.2em]
        \includegraphics[width=0.16\linewidth]{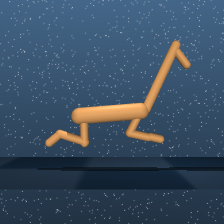}
        \includegraphics[width=0.16\linewidth]{visualizations/tasks/cup-catch-var1-1.png}
        \includegraphics[width=0.16\linewidth]{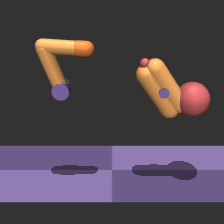}
        \includegraphics[width=0.16\linewidth]{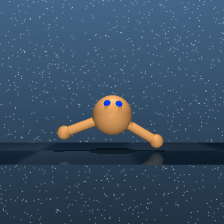}
        \includegraphics[width=0.16\linewidth]{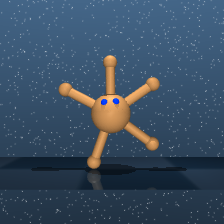}
        \includegraphics[width=0.16\linewidth]{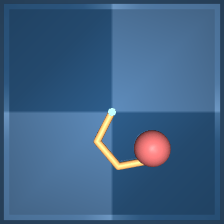}
    \end{minipage}\\[1em]
    \begin{minipage}{\textwidth}%
        \centering
        Meta-World\\[0.2em]
        \includegraphics[width=0.16\linewidth]{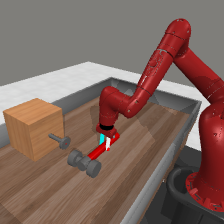}
        \includegraphics[width=0.16\linewidth]{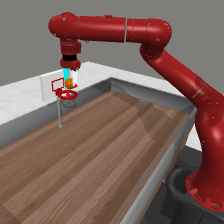}
        \includegraphics[width=0.16\linewidth]{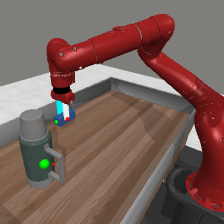}
        \includegraphics[width=0.16\linewidth]{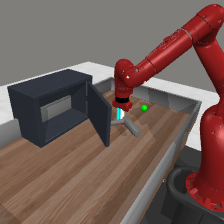}
        \includegraphics[width=0.16\linewidth]{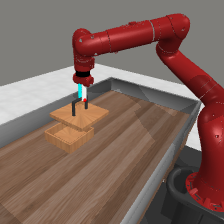}
        \includegraphics[width=0.16\linewidth]{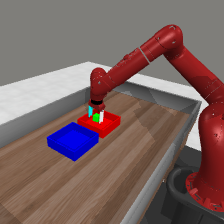}
    \end{minipage}\\[1em]
    \begin{minipage}{\textwidth}%
        \centering
        ManiSkill3\\[0.2em]
        \includegraphics[width=0.16\linewidth]{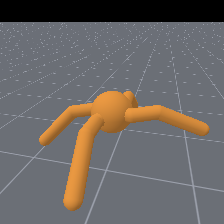}
        \includegraphics[width=0.16\linewidth]{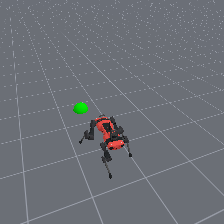}
        \includegraphics[width=0.16\linewidth]{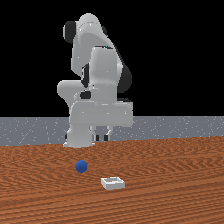}
        \includegraphics[width=0.16\linewidth]{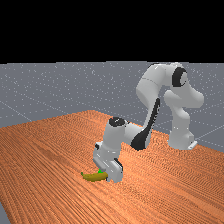}
        \includegraphics[width=0.16\linewidth]{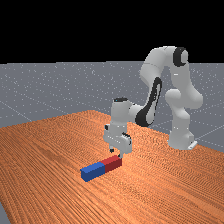}
        \includegraphics[width=0.16\linewidth]{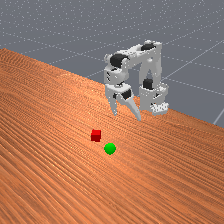}
    \end{minipage}\\[1em]
    \begin{minipage}{\textwidth}%
        \centering
        MuJoCo\\[0.2em]
        \includegraphics[width=0.16\linewidth]{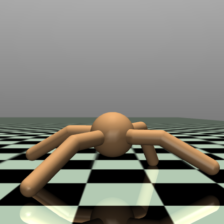}
        \includegraphics[width=0.16\linewidth]{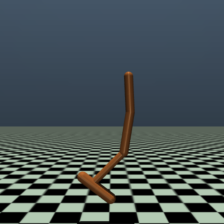}
        \includegraphics[width=0.16\linewidth]{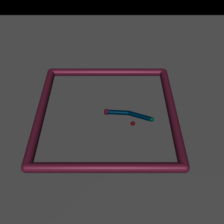}
        \includegraphics[width=0.16\linewidth]{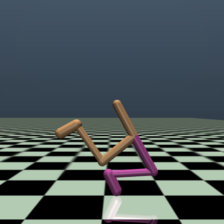}
        \includegraphics[width=0.16\linewidth]{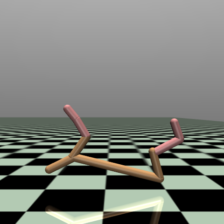}
        \includegraphics[width=0.16\linewidth]{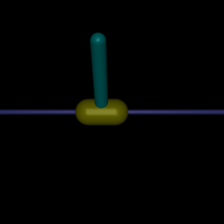}
    \end{minipage}\\[1em]
    \begin{minipage}{\textwidth}%
        \centering
        Box2D\\[0.2em]
        \includegraphics[width=0.16\linewidth]{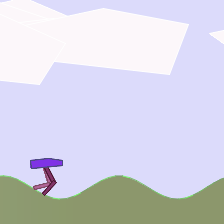}
        \includegraphics[width=0.16\linewidth]{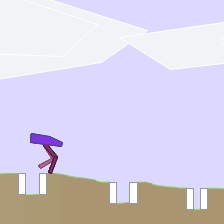}
        \includegraphics[width=0.16\linewidth]{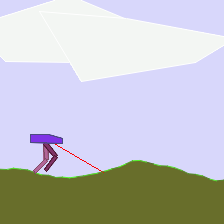}
        \includegraphics[width=0.16\linewidth]{visualizations/tasks/lunarlander-takeoff-5.png}
        \includegraphics[width=0.16\linewidth]{visualizations/tasks/ms-lift-peg-2.png}
        \includegraphics[width=0.16\linewidth]{visualizations/tasks/ms-pick-cube-so-0.png}
    \end{minipage}\\[1em]
    \caption{\textbf{Visualization of task domains (1 of 2).} We show sample tasks from each of the 10 task domains that we consider.}
    \label{fig:appendix-task-grid}
\end{figure}

\begin{figure}[h]
    \centering
    \begin{minipage}{\textwidth}%
        \centering
        RoboDesk\\[0.2em]
        \includegraphics[width=0.16\linewidth]{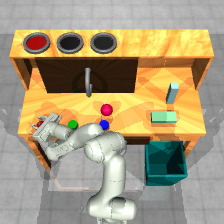}
        \includegraphics[width=0.16\linewidth]{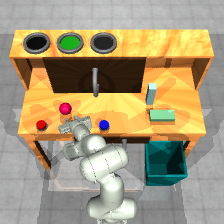}
        \includegraphics[width=0.16\linewidth]{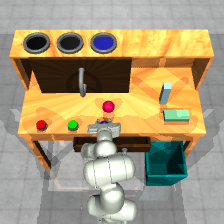}
        \includegraphics[width=0.16\linewidth]{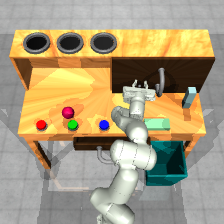}
        \includegraphics[width=0.16\linewidth]{visualizations/tasks/rd-open-drawer-1.png}
        \includegraphics[width=0.16\linewidth]{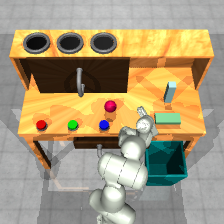}
    \end{minipage}\\[1em]
    \begin{minipage}{\textwidth}%
        \centering
        OGBench\\[0.2em]
        \includegraphics[width=0.16\linewidth]{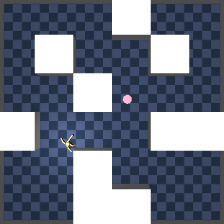}
        \includegraphics[width=0.16\linewidth]{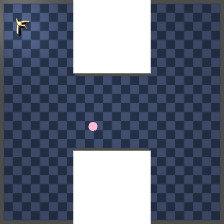}
        \includegraphics[width=0.16\linewidth]{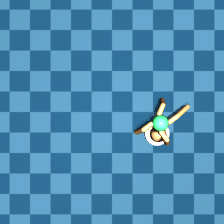}
        \includegraphics[width=0.16\linewidth]{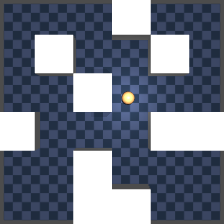}
        \includegraphics[width=0.16\linewidth]{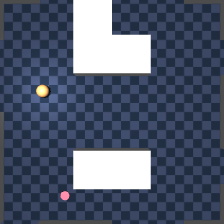}
        \includegraphics[width=0.16\linewidth]{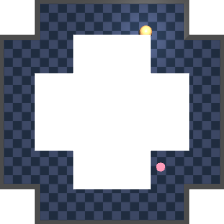}
    \end{minipage}\\[1em]
    \begin{minipage}{\textwidth}%
        \centering
        MiniArcade\\[0.2em]
        \includegraphics[width=0.16\linewidth]{visualizations/tasks/pygame-bird-attack-2.png}
        \includegraphics[width=0.16\linewidth]{visualizations/tasks/pygame-foraging-2.png}
        \includegraphics[width=0.16\linewidth]{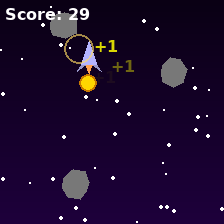}
        \includegraphics[width=0.16\linewidth]{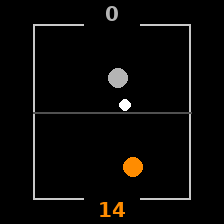}
        \includegraphics[width=0.16\linewidth]{visualizations/tasks/pygame-dungeon-explorer1-0.png}
        \includegraphics[width=0.16\linewidth]{visualizations/tasks/pygame-whirlpool-1.png}
    \end{minipage}\\[1em]
    \begin{minipage}{\textwidth}%
        \centering
        Atari\\[0.2em]
        \includegraphics[width=0.16\linewidth]{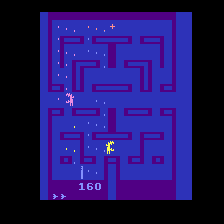}
        \includegraphics[width=0.16\linewidth]{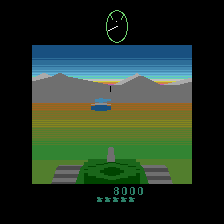}
        \includegraphics[width=0.16\linewidth]{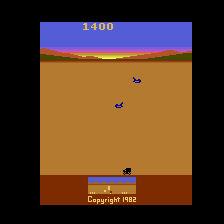}
        \includegraphics[width=0.16\linewidth]{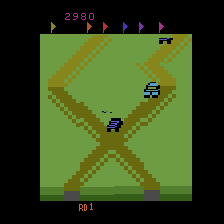}
        \includegraphics[width=0.16\linewidth]{visualizations/tasks/atari-road-runner-3.png}
        \includegraphics[width=0.16\linewidth]{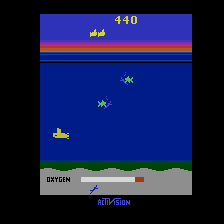}
    \end{minipage}\\[1em]
    \caption{\textbf{Visualization of task domains (2 of 2).} We show sample tasks from each of the 10 task domains that we consider.}
    \label{fig:appendix-task-grid}
\end{figure}

\begin{table}[t]
\centering
\caption{\textbf{Task list for finetuning experiments.} We finetune on a set of 10 seen (also used in pretraining) + 10 unseen (held out) tasks. The two task sets are listed below. Note that unseen tasks are developed specifically for MMBench2.}
\label{tab:e8-task-list}
\vspace{0.05in}
\begin{tabular}{ll}
\toprule
Task & Domain \\
\midrule
\multicolumn{2}{l}{\underline{\textcolor{nhdpurple}{\textbf{Seen task set}}}} \\
cup-catch                & DMControl  \\
finger-turn-easy         & DMControl  \\
mw-push                  & Meta-World \\
ms-push-cube             & ManiSkill  \\
lunarlander-hover        & Box2D      \\
og-point-maze            & OGBench    \\
og-point-bottleneck      & OGBench    \\
pygame-point-maze-var1   & MiniArcade     \\
pygame-pong              & MiniArcade     \\
pygame-bird-attack       & MiniArcade     \\
\midrule
\multicolumn{2}{l}{\underline{\textcolor{nhdpurple}{\textbf{Unseen task set}}}} \\
cup-catch-var1           & DMControl  \\
finger-turn-easy-var1    & DMControl  \\
ms-push-banana           & ManiSkill  \\
og-point-var1            & OGBench    \\
og-point-var2            & OGBench    \\
pygame-point-maze-var4   & MiniArcade     \\
pygame-reacher-easy      & MiniArcade     \\
pygame-dungeon-explorer1 & MiniArcade     \\
pygame-foraging          & MiniArcade     \\
pygame-whirlpool         & MiniArcade     \\
\bottomrule
\end{tabular}
\end{table}

\clearpage
\newpage

\section{Data Collection}
\label{sec:data-collection}

We base our data collection on that of MMBench, a 200-task benchmark designed primarily for online RL; it provides a total of 4k expert demonstrations collected via single-task expert policies, as well as live environments with ground-truth action and reward labels for all tasks. While MMBench is well suited for the multi-task online RL setting it was originally developed for, a dataset that consists solely of expert demonstrations lacks diversity in terms of behavior which, as our experiments show, is a significant source of hallucination in world models.

\textbf{The overarching goal of MMBench2 is to produce a large, diverse dataset for visual world modeling} and research in common failure modes such as hallucination. To do so, we use the single-task expert policies $\pi^{\star}$ of MMBench as a basis for our data collection, but crucially augment policies to generate diverse behaviors, and also collect data via non-expert policies, including humans. Our methods of data collection can be summarized as follows:
\begin{itemize}[itemsep=2pt, topsep=2pt, parsep=0pt, leftmargin=1.75em]
    \vspace{-0.05in}
    \item \textbf{Random policy.} Sample actions uniformly in $[-1, 1]$. Diverse actions; poor task performance.
    \item \textbf{No-op actions.} Set actions $\mathbf{a} = \mathbf{0}$. Models dynamics without agent interference.
    \item \textbf{Expert actions.} Sample from the expert policy $\pi^{\star}$ without added noise. High task performance; low data diversity despite it being a stochastic policy.
    \item \textbf{Transformed expert actions.} Sample from $\pi^{\star}$, then apply any of five transforms with some probability: scale by $\varepsilon \in [0, 1]$, action dropout, flip sign, all-zero action, repeat previous action. The same transform may be applied for multiple steps. Provides counterfactual transitions.
    \item \textbf{Structured noise.} Sample from $\pi^{\star}$, then apply any of three noise types with parameters sampled on a per-episode basis: Gaussian noise, Ornstein-Uhlenbeck noise (temporally correlated), convex mixture with random policy. Mixed performance; improves data diversity.
    \item \textbf{Curiosity-driven.} Sample trajectories that maximize our developed hallucination predictor $u_r^{\mathrm{norm}}$, selected by CEM-based planning with the pretrained world model.
    \item \textbf{Human play data.} Actions are selected by a human interacting with the environment via a keyboard interface. High data diversity; not task-driven.
\end{itemize}

\begin{figure}[h]
    \centering
    \includegraphics[width=\linewidth]{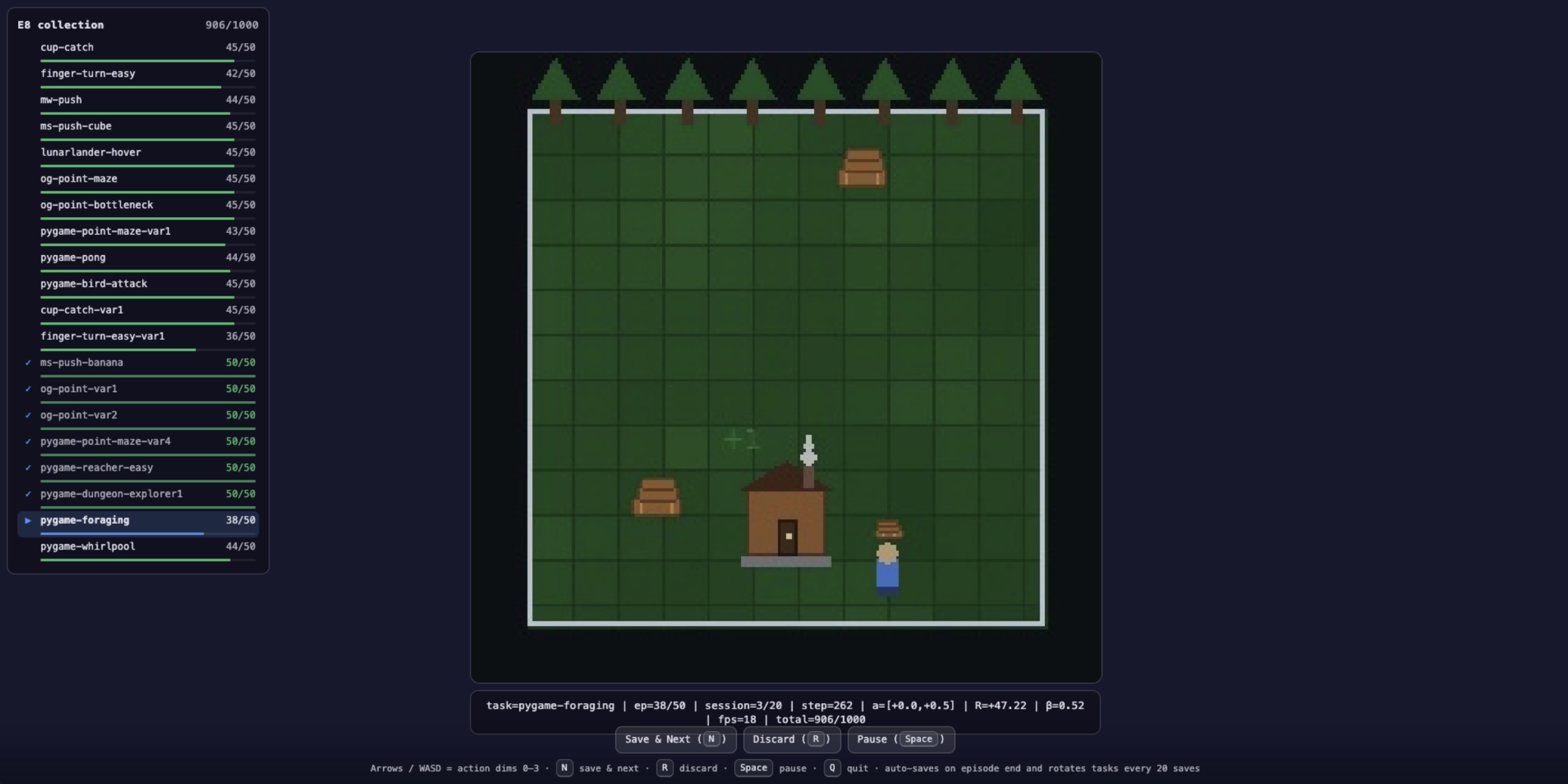}
    \caption{\textbf{Web interface for human play data collection.} We develop a simple web interface for data collection. A human user interacts with the environment using the keyboard, and interaction data is saved and later used for world model training. We collect a total of $1{,}400$ trajectories using this interface.}
    \label{fig:data-collection-interface}
\end{figure}

With the exception of human play data, we frequently switch between any of the above behaviors within a single episode to maximize diversity of our training data. For example, we may apply structured noise at random for a number of steps to visit a less frequently visited part of the state-action space and then switch back to the expert policy to generate recovery behavior. Data is collected across all 210 live environments, and ground-truth actions (the action selected by the behavior policy) and reward labels (task reward obtained from the environment) are logged. We collect pretraining data for 200 tasks, and additional data for targeted finetuning across 10 seen (included in pretraining) as well as 10 unseen tasks, with data partitions separated by behavior policy: \emph{random}, \emph{noop}, \emph{expert}, \emph{mixed}, \emph{curiosity}, and \emph{human}. Figure~\ref{fig:data-collection-interface} shows a screenshot of the interface used to collect human play data.

\clearpage
\newpage

\section{Additional Results}
\label{sec:appendix-additional-results}

\begin{table}[h]
\centering
\caption{\textbf{Detecting hallucination events.} Per-task AUROC against two hallucination labels (action-ignored, scene-diverging) on held-out test data from all 200 pretraining tasks. Our three proposed predictors $u_r^{\text{norm}}, u_f^{\text{norm}}, u_s^{\text{norm}}$ reliably predict hallucinations. Higher is better ($\uparrow$).}
\label{tab:hallucination-metrics}
\vspace{0.05in}
\begin{tabular}{lcc}
\toprule
Predictor & \shortstack{Action\\ignored} & \shortstack{Scene\\divergent} \\
\midrule
Tokenizer residual $u_r^{\text{norm}}$           & $\mathbf{0.887}$ & $0.919$          \\
Flow instability $u_f^{\text{norm}}$           & $0.868$          & $\mathbf{0.939}$ \\
Inter-seed variance $u_s^{\text{norm}}$           & $0.873$          & $0.934$          \\
Scene motion (latent) $m$     & $0.803$          & $0.927$          \\
kNN distance (global)         & $0.814$          & $0.731$          \\
Flow instability $u_f$ (raw)  & $0.752$          & $0.854$          \\
$n_\text{frames}$ (baseline)  & $0.596$          & $0.534$          \\
\bottomrule
\end{tabular}
\end{table}

\begin{figure}[h]
    \centering
    \begin{minipage}{\textwidth}%
        \centering
        \begin{minipage}[b]{0.16\textwidth}%
            \centering
            Task\\[0.1em]
            \includegraphics[width=\linewidth]{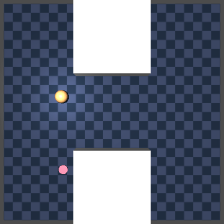}
        \end{minipage}
        \begin{minipage}[b]{0.16\textwidth}%
            \centering
            No-op\\[0.1em]
            \includegraphics[width=\linewidth]{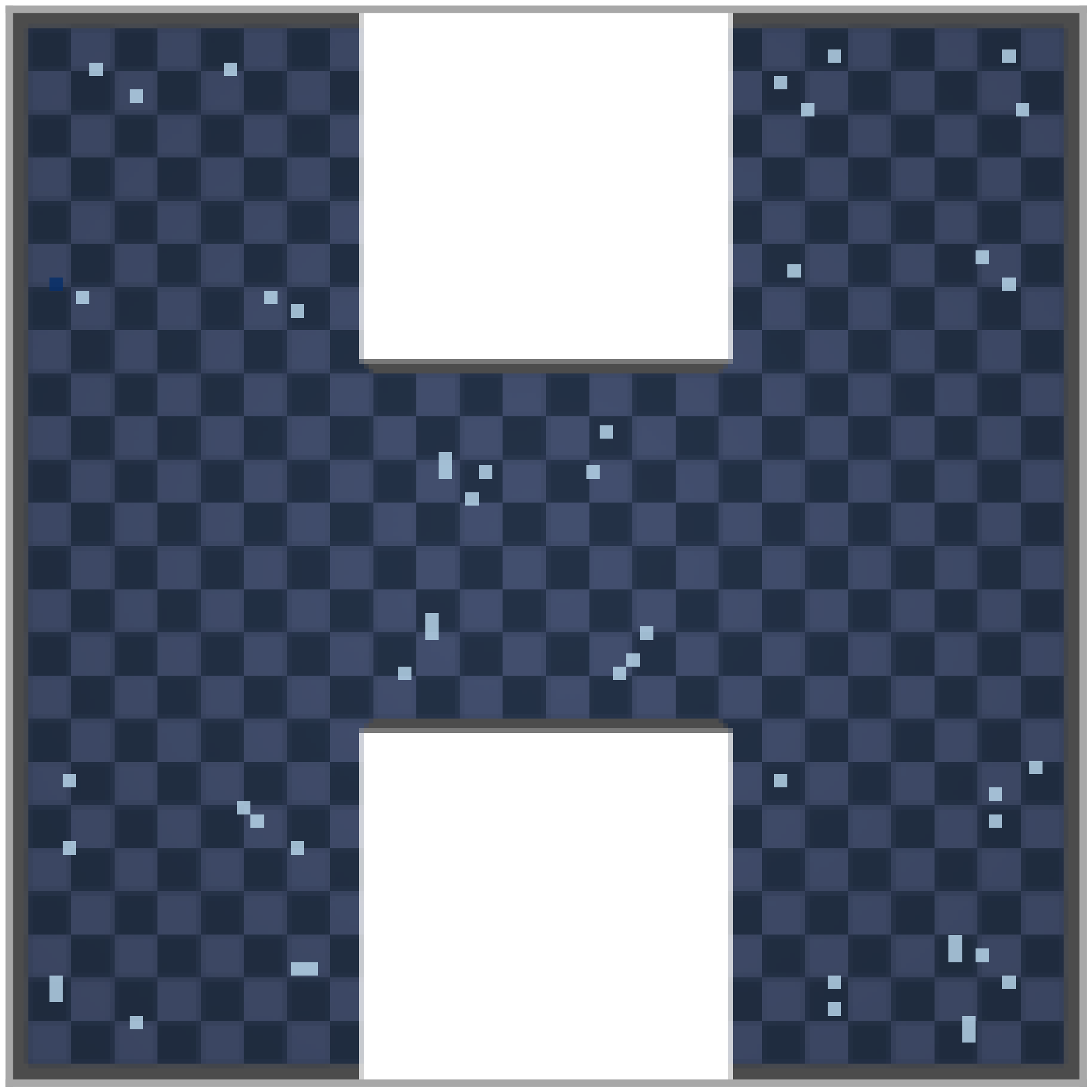}
        \end{minipage}
        \begin{minipage}[b]{0.16\textwidth}%
            \centering
            Random\\[0.1em]
            \includegraphics[width=\linewidth]{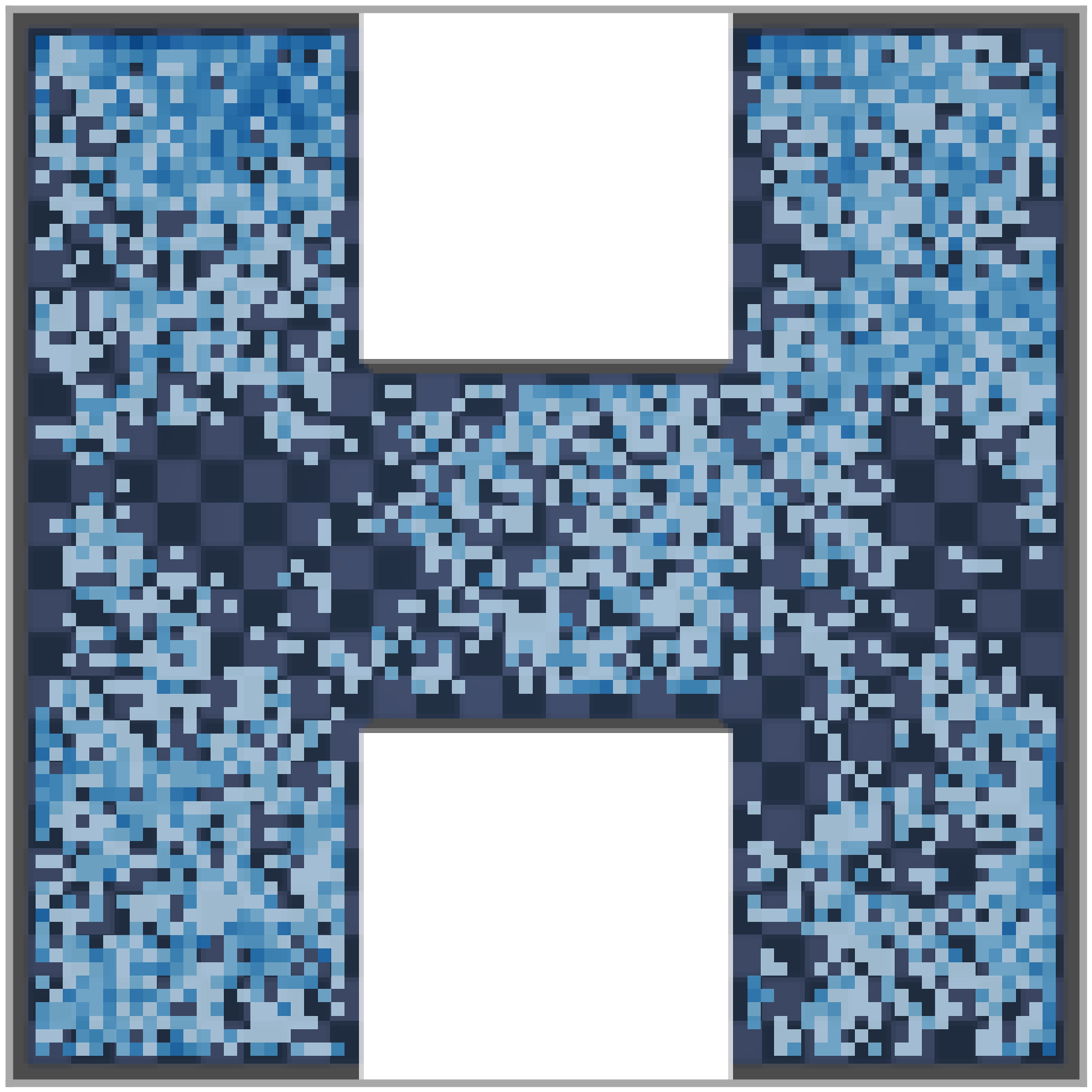}
        \end{minipage}
        \begin{minipage}[b]{0.16\textwidth}%
            \centering
            Expert $\pi$\\[0.1em]
            \includegraphics[width=\linewidth]{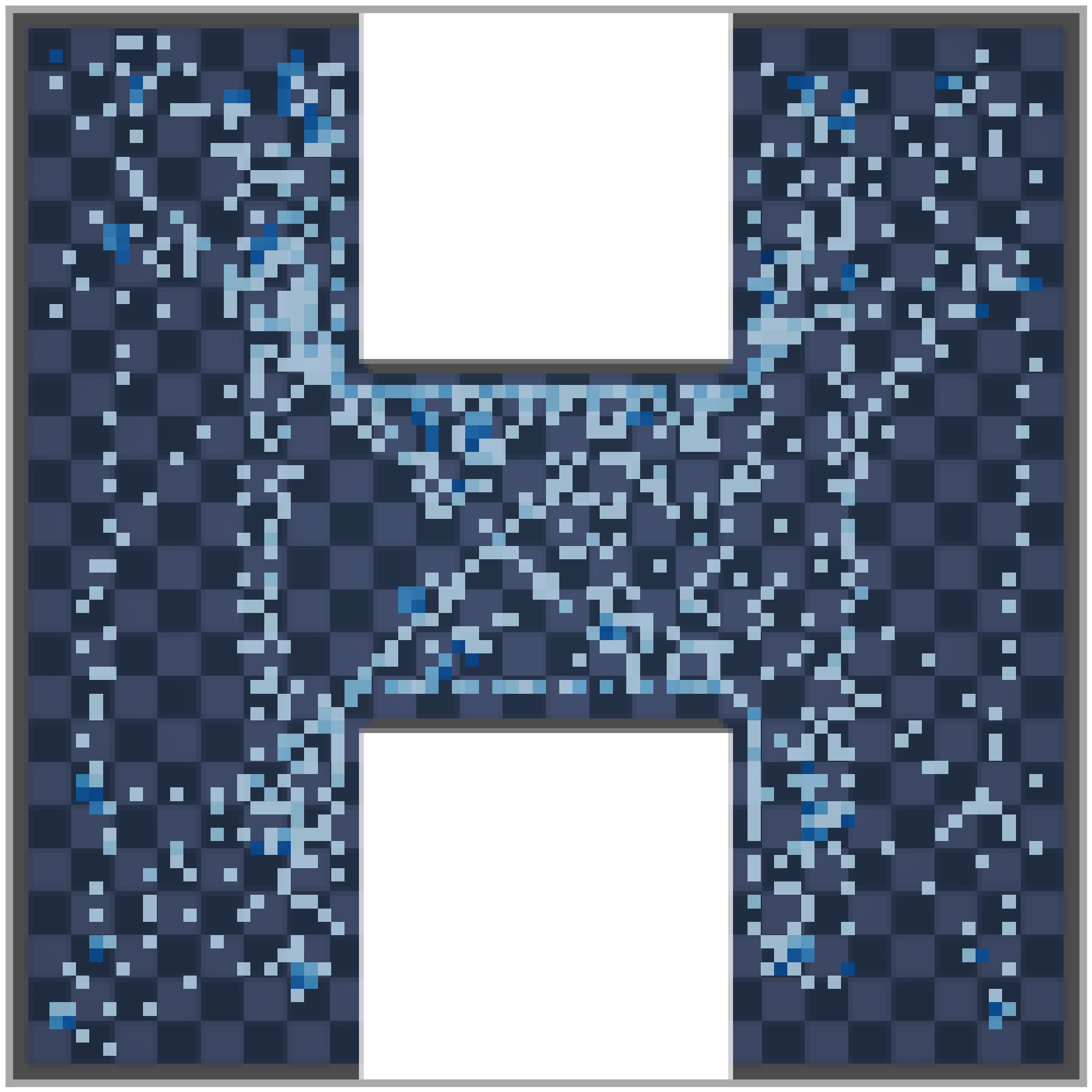}
        \end{minipage}
        \begin{minipage}[b]{0.16\textwidth}%
            \centering
            Curiosity\\[0.1em]
            \includegraphics[width=\linewidth]{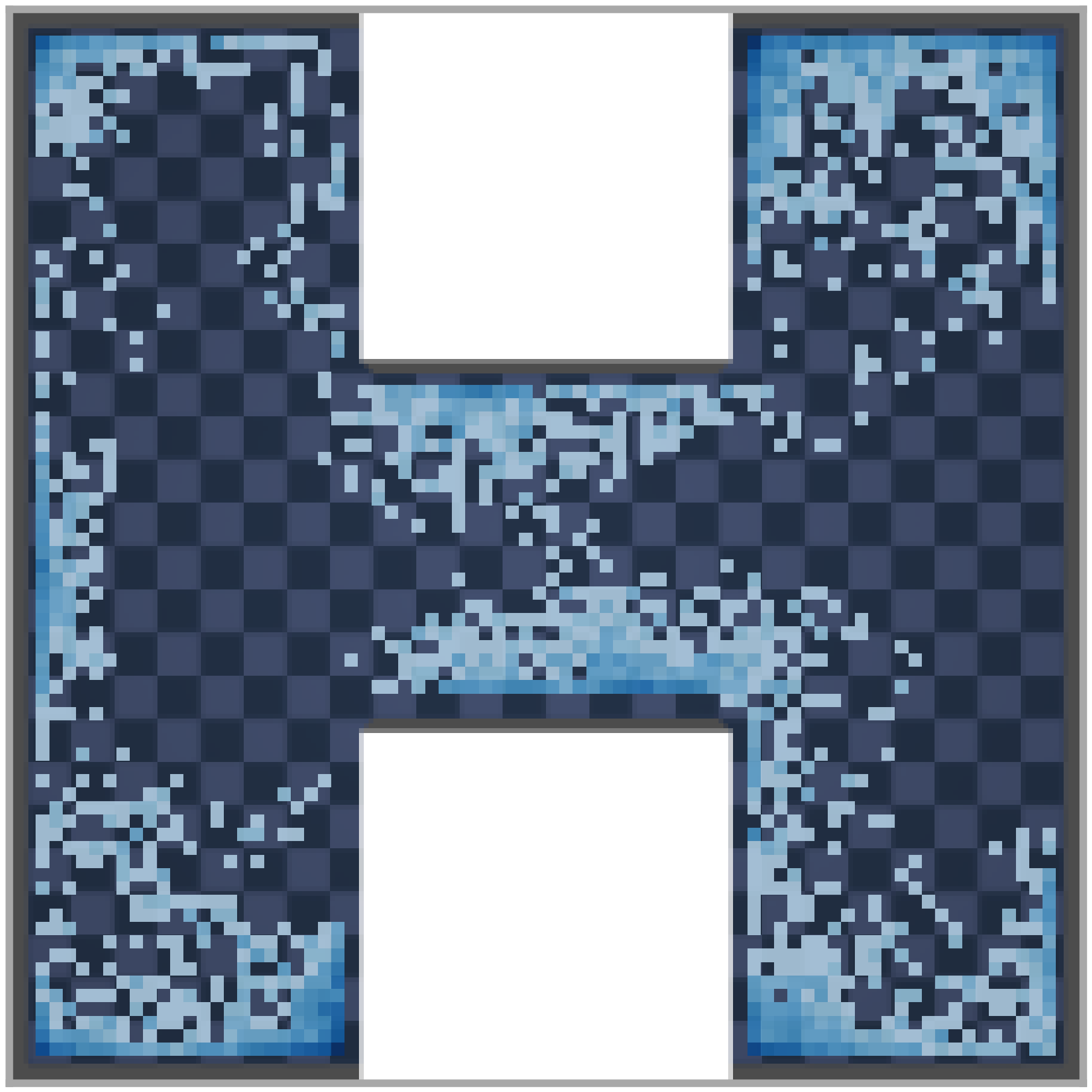}
        \end{minipage}
        \begin{minipage}[b]{0.16\textwidth}%
            \centering
            Human\\[0.1em]
            \includegraphics[width=\linewidth]{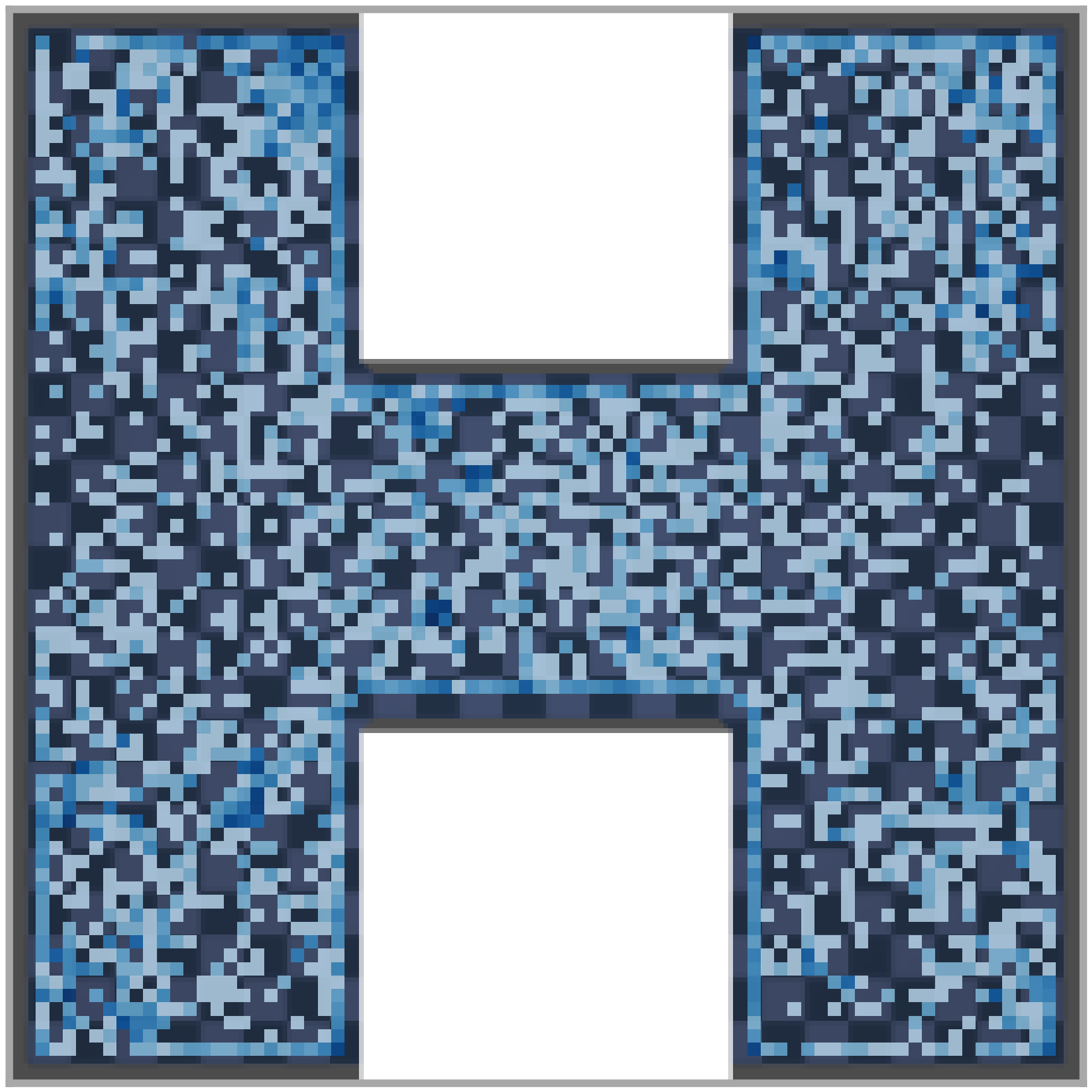}
        \end{minipage}\\[0.2em]
        Point Maze Bottleneck (OGBench)
    \end{minipage}\\[1em]
    \begin{minipage}{\textwidth}%
        \centering
        \begin{minipage}[b]{0.16\textwidth}%
            \centering
            Task\\[0.1em]
            \includegraphics[width=\linewidth]{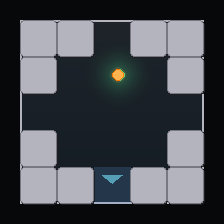}
        \end{minipage}
        \begin{minipage}[b]{0.16\textwidth}%
            \centering
            No-op\\[0.1em]
            \includegraphics[width=\linewidth]{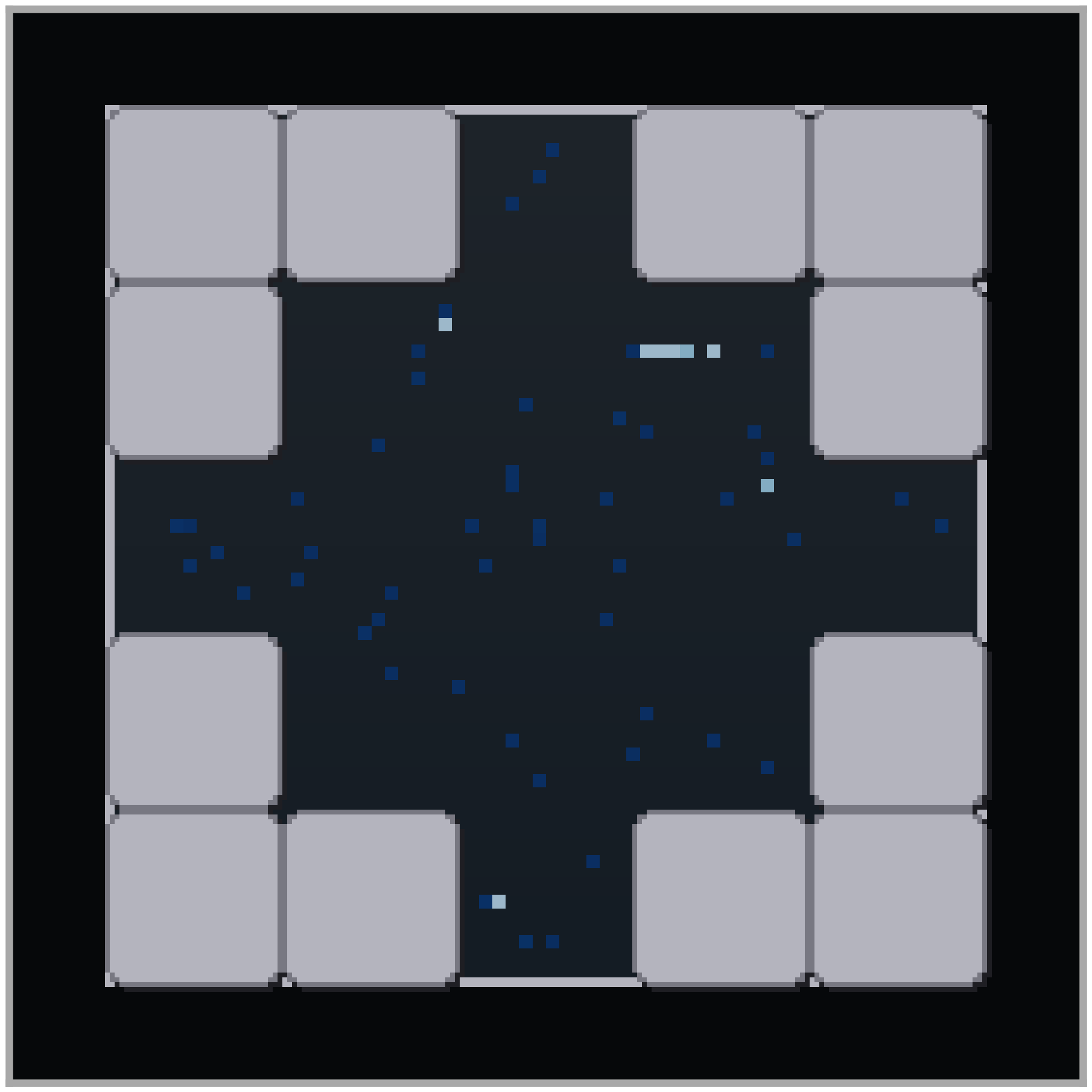}
        \end{minipage}
        \begin{minipage}[b]{0.16\textwidth}%
            \centering
            Random\\[0.1em]
            \includegraphics[width=\linewidth]{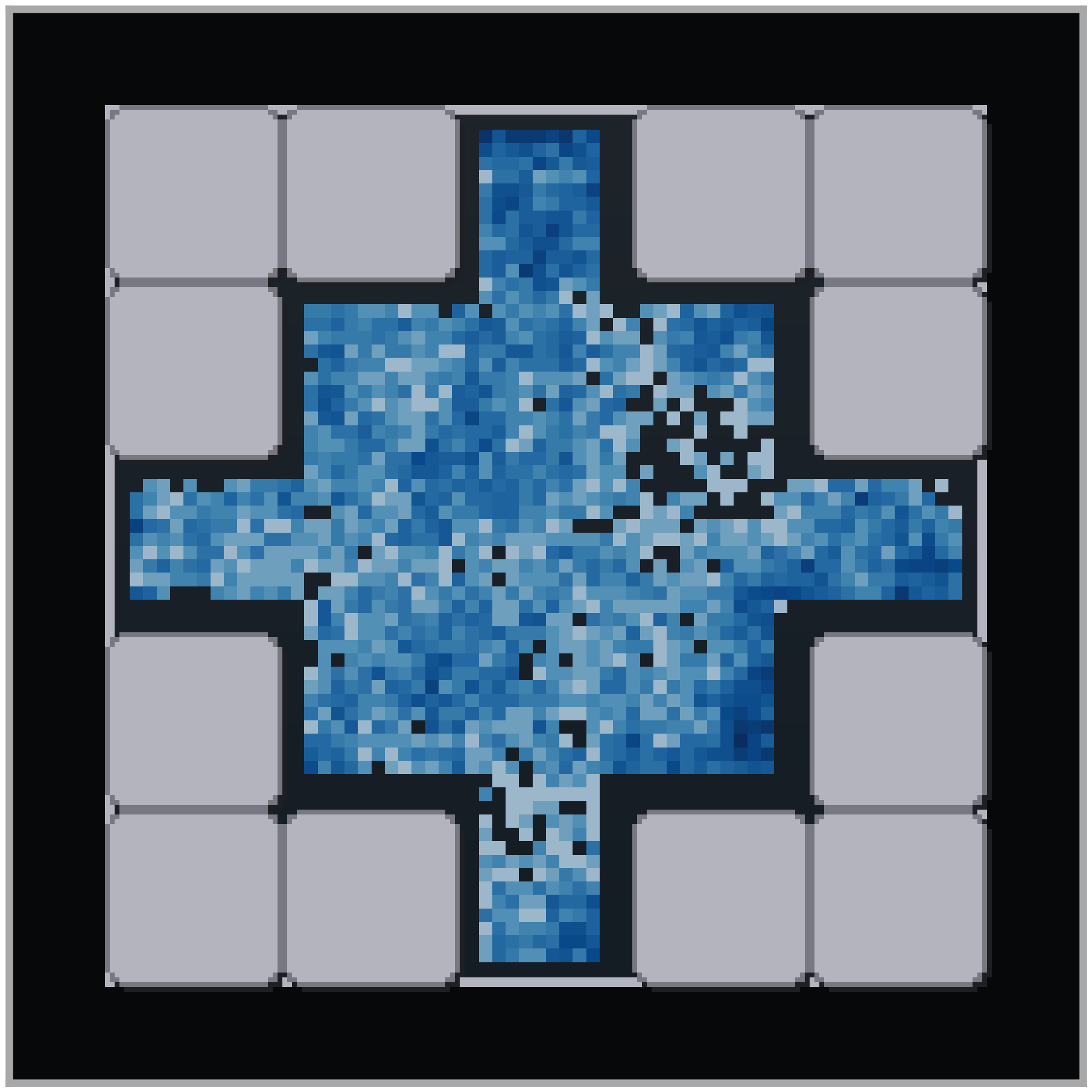}
        \end{minipage}
        \begin{minipage}[b]{0.16\textwidth}%
            \centering
            Expert $\pi$\\[0.1em]
            \includegraphics[width=\linewidth]{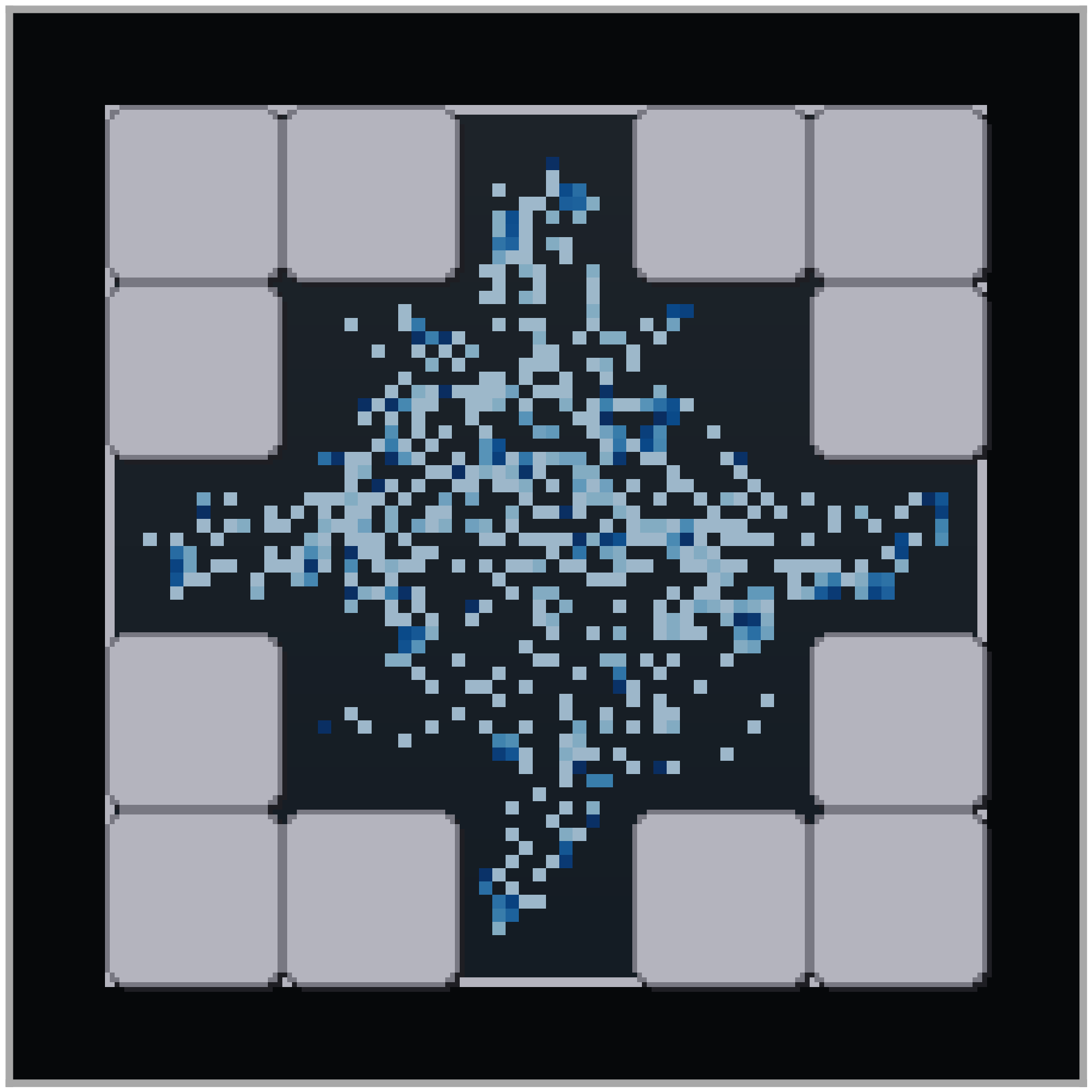}
        \end{minipage}
        \begin{minipage}[b]{0.16\textwidth}%
            \centering
            Curiosity\\[0.1em]
            \includegraphics[width=\linewidth]{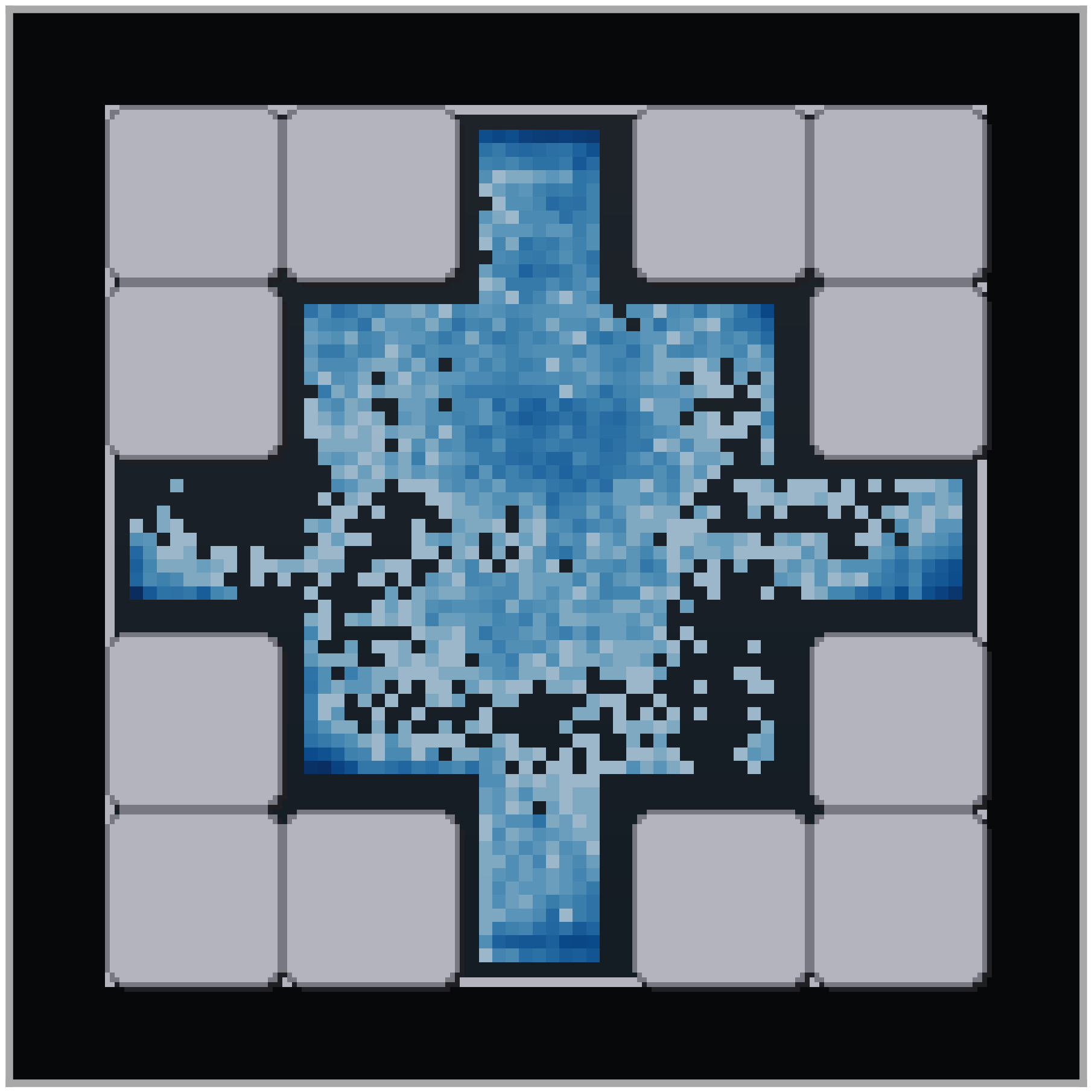}
        \end{minipage}
        \begin{minipage}[b]{0.16\textwidth}%
            \centering
            Human\\[0.1em]
            \includegraphics[width=\linewidth]{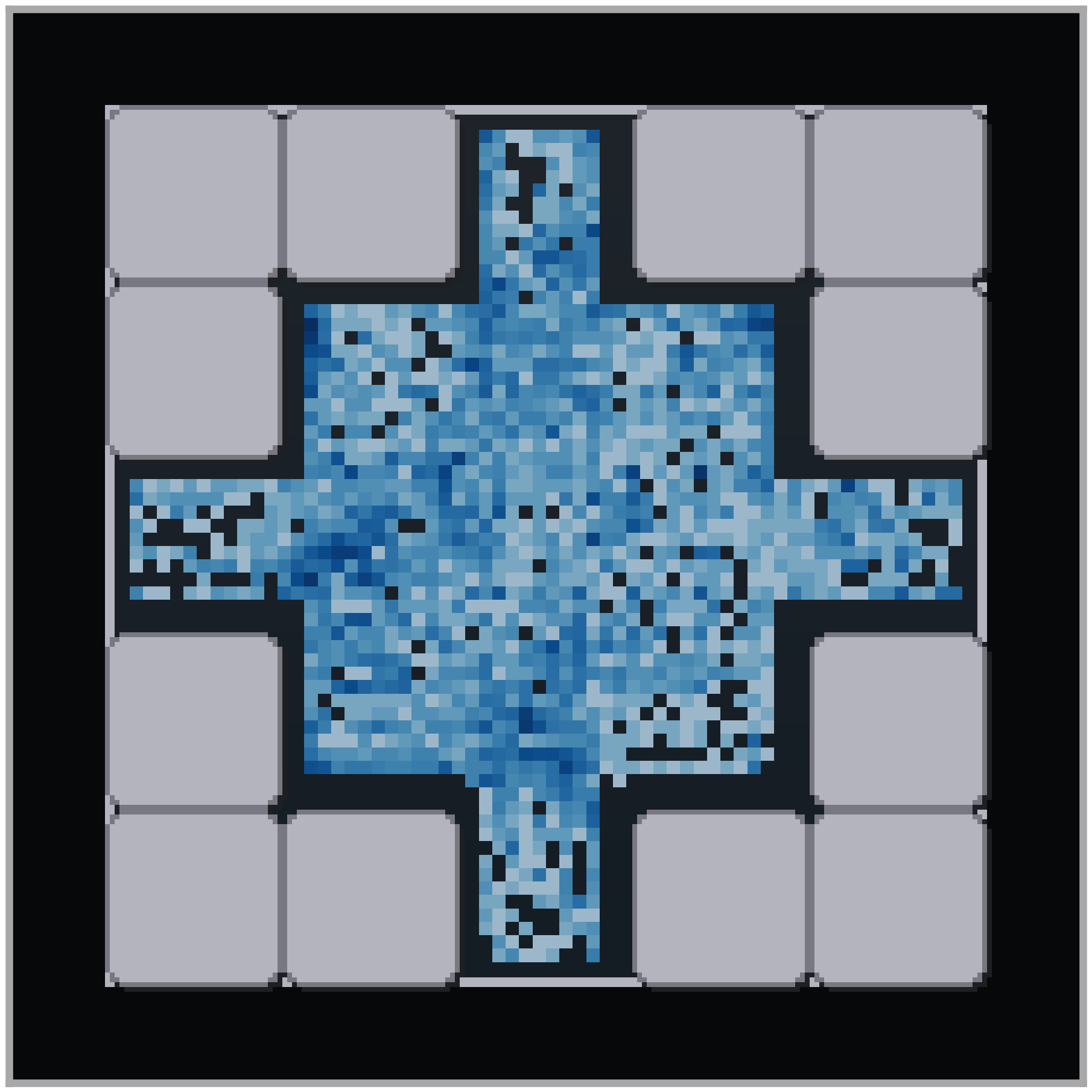}
        \end{minipage}\\[0.2em]
        Point Maze Variation 4 (MiniArcade)
    \end{minipage}\\[1em]
    \begin{minipage}{\textwidth}%
        \centering
        \begin{minipage}[b]{0.16\textwidth}%
            \centering
            Task\\[0.1em]
            \includegraphics[width=\linewidth]{visualizations/tasks/cup-catch-1.png}
        \end{minipage}
        \begin{minipage}[b]{0.16\textwidth}%
            \centering
            No-op\\[0.1em]
            \includegraphics[width=\linewidth]{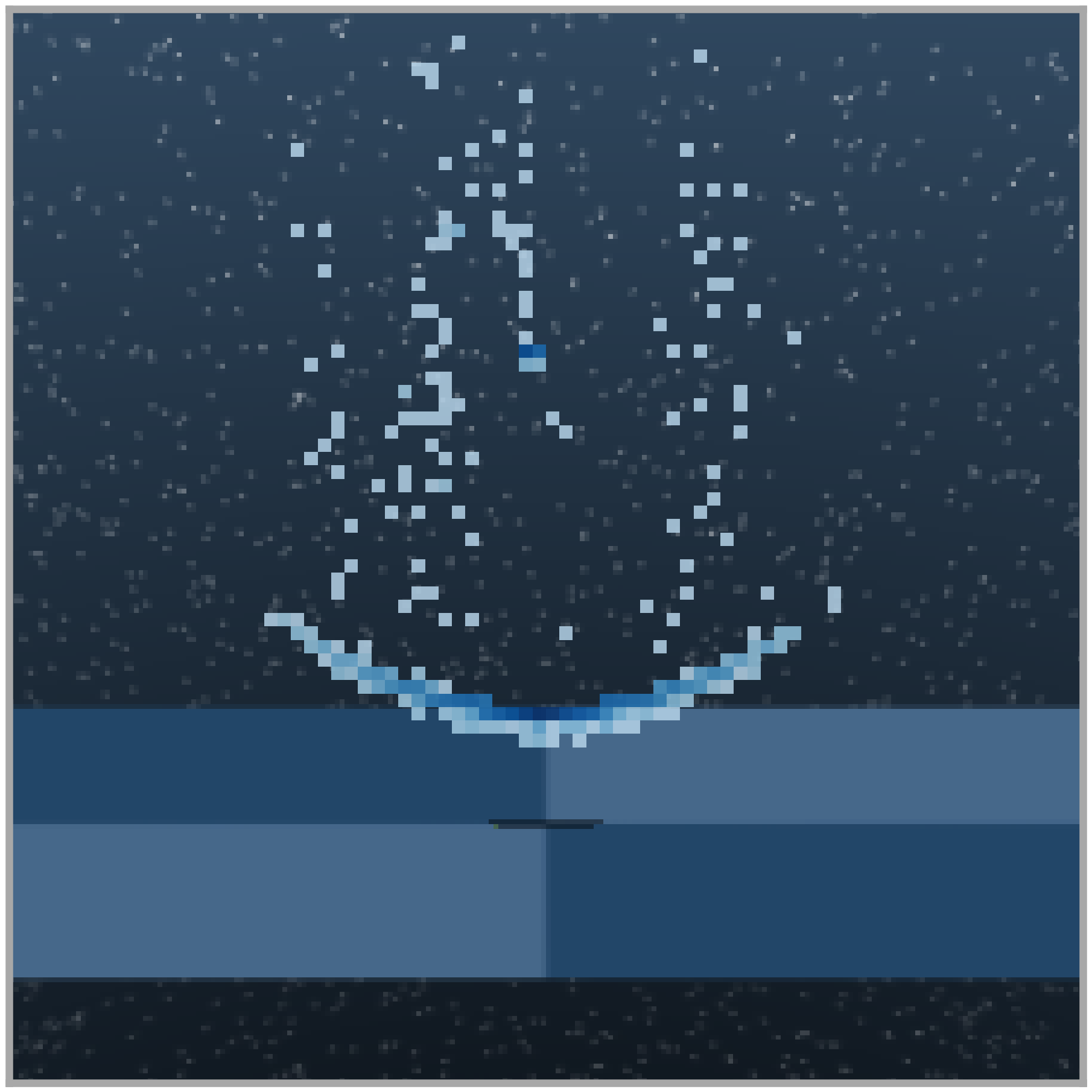}
        \end{minipage}
        \begin{minipage}[b]{0.16\textwidth}%
            \centering
            Random\\[0.1em]
            \includegraphics[width=\linewidth]{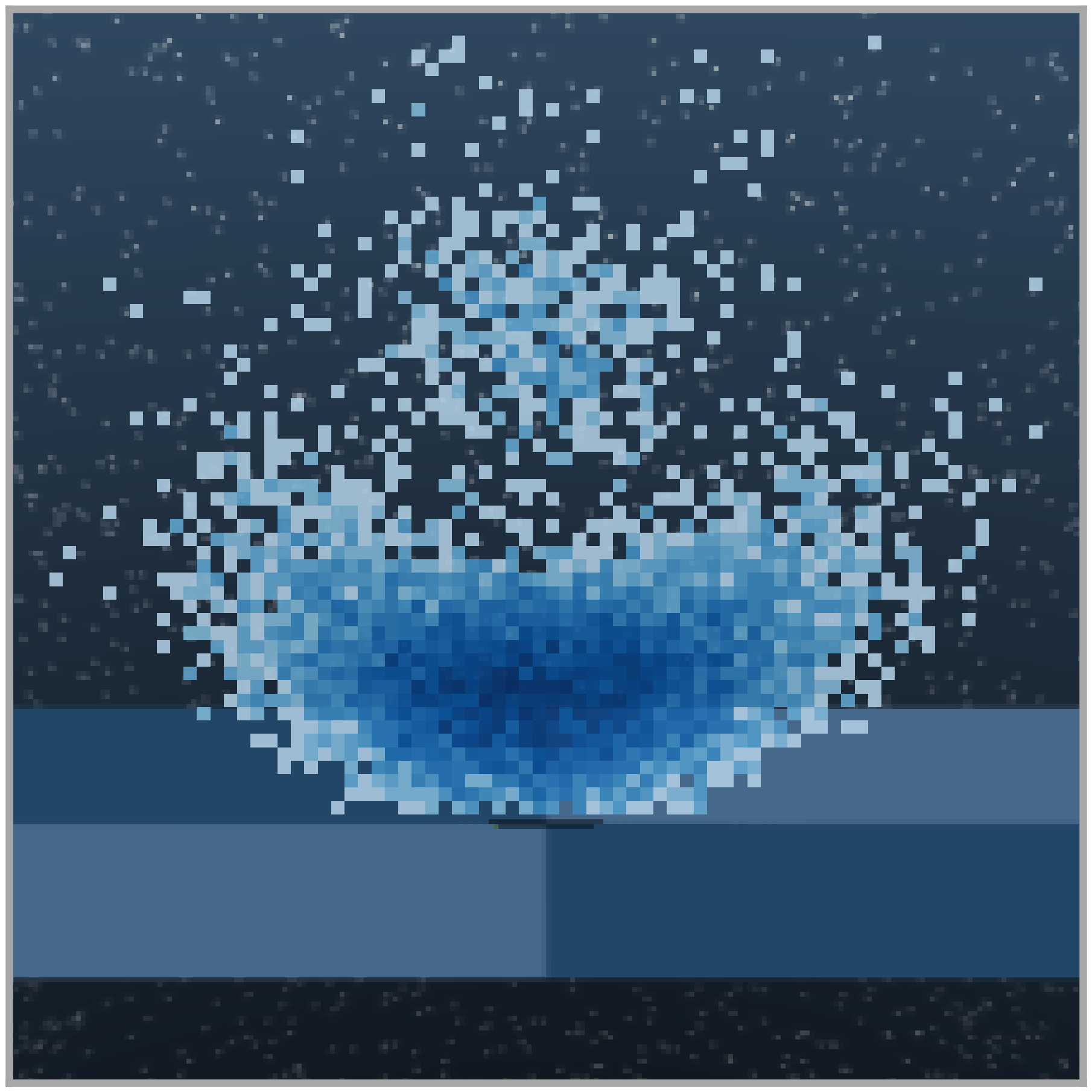}
        \end{minipage}
        \begin{minipage}[b]{0.16\textwidth}%
            \centering
            Expert $\pi$\\[0.1em]
            \includegraphics[width=\linewidth]{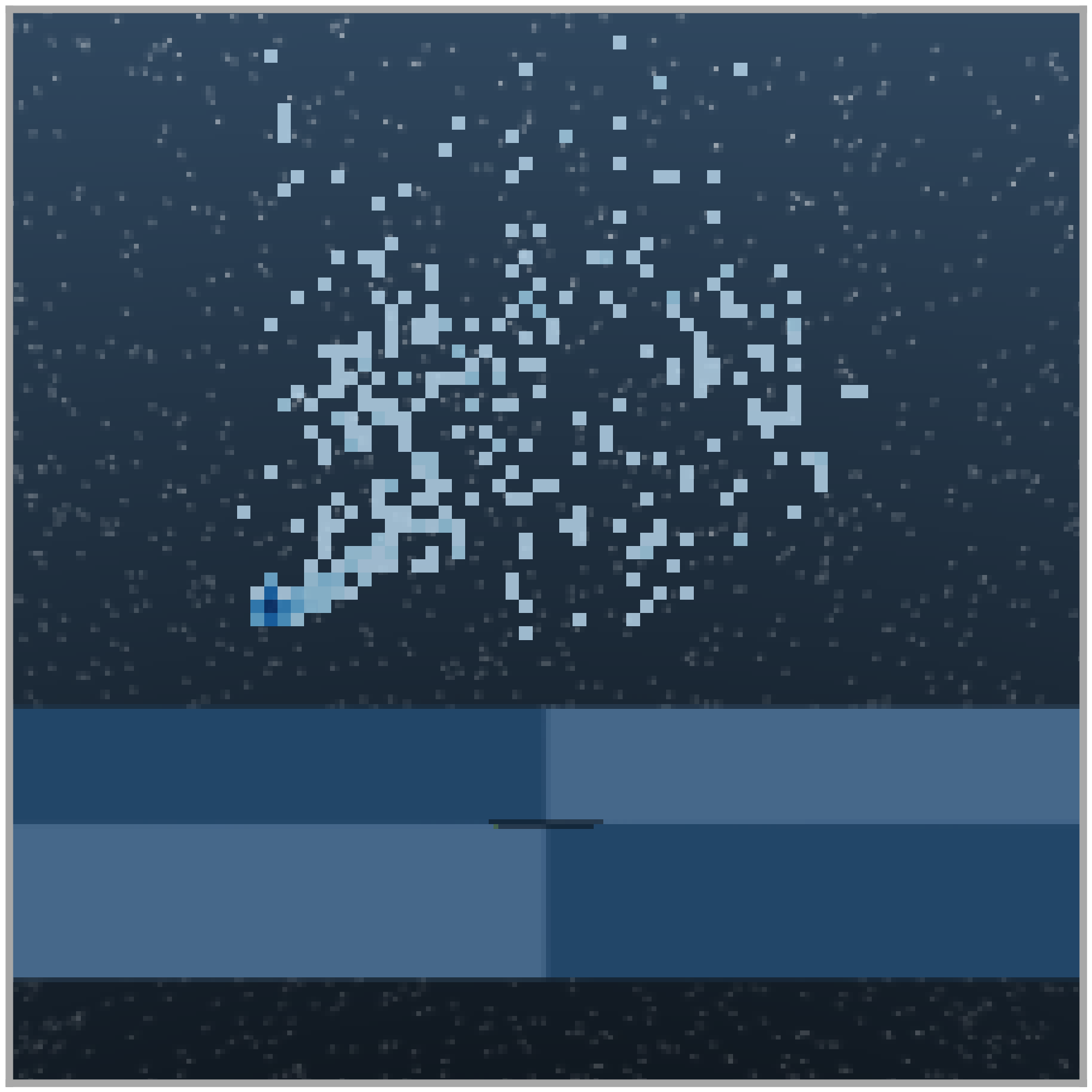}
        \end{minipage}
        \begin{minipage}[b]{0.16\textwidth}%
            \centering
            Curiosity\\[0.1em]
            \includegraphics[width=\linewidth]{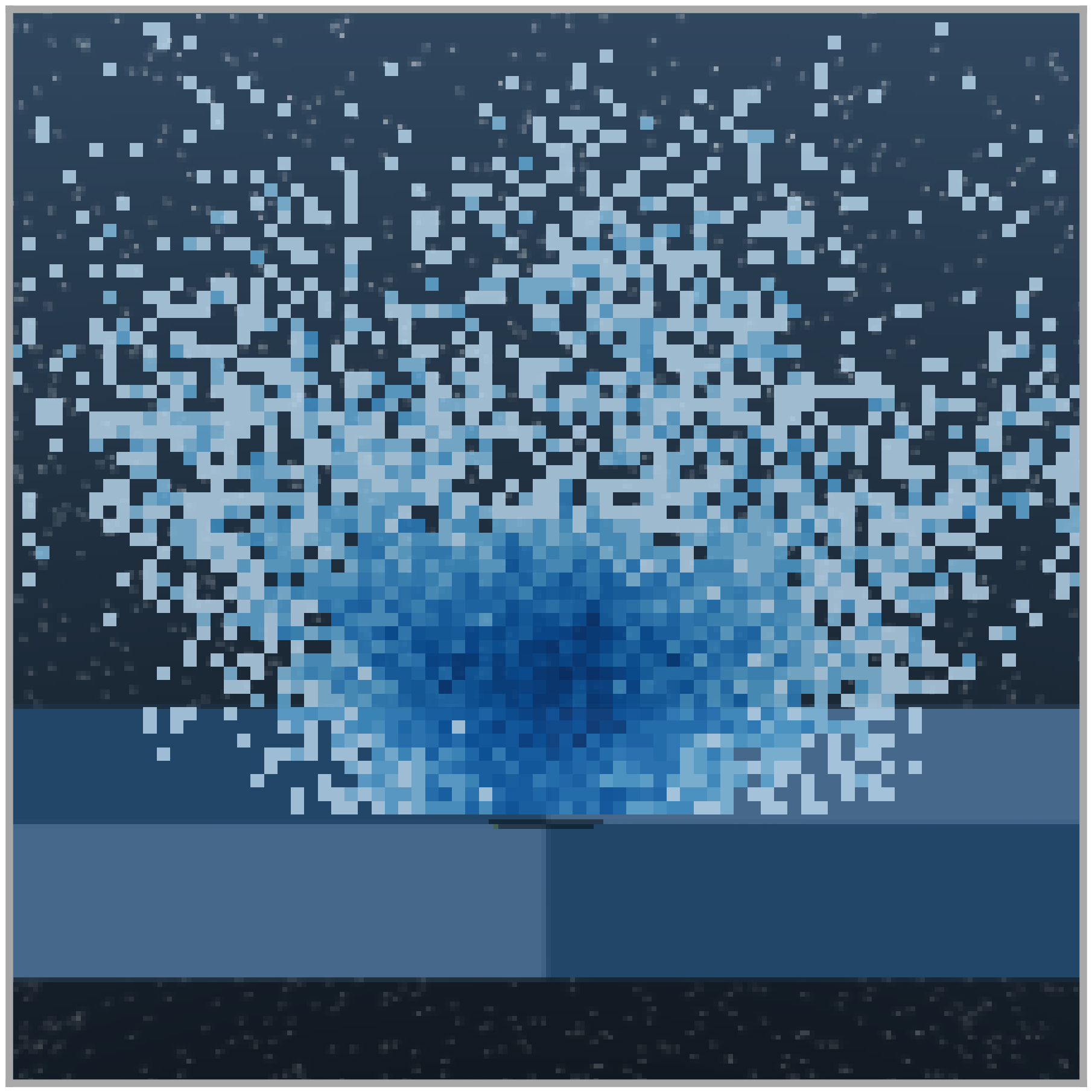}
        \end{minipage}
        \begin{minipage}[b]{0.16\textwidth}%
            \centering
            Human\\[0.1em]
            \includegraphics[width=\linewidth]{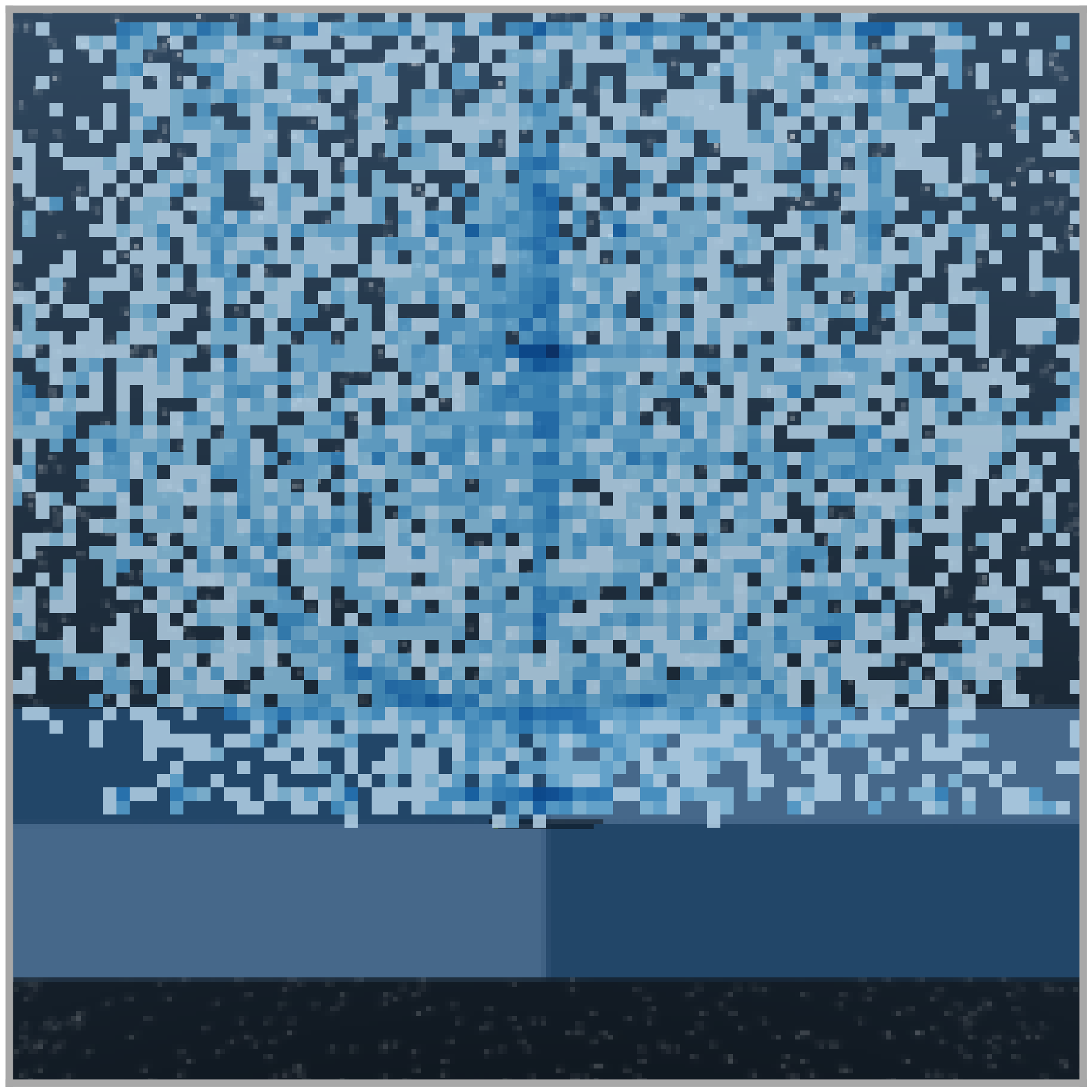}
        \end{minipage}\\[0.2em]
        Cup Catch (DMControl)
    \end{minipage}
    \vspace{-0.025in}
    \caption{\textbf{Data coverage by collection method.} We show state densities for five online data collection policies (no-op, random, expert, curiosity, and human) across three additional tasks besides those in Figure~\ref{fig:data-collection-policies}.}
    \label{fig:appendix-data-collection-policies}
\end{figure}

\begin{table}[t]
\centering
\small
\caption{\textbf{Targeted data collection for finetuning on 10 unseen tasks (\emph{expert} test set).} We finetune our world model on a set of 10 seen + 10 unseen tasks, varying data source and finetuning strategy. Each data source contains 50 trajectories per task. The offline results in Table~\ref{tab:e8-unseen} are computed over a test set that consists of trajectories from both expert policies and human play; this table complements those results by evaluating only on the \emph{expert} trajectories. We find that \emph{the policy that aligns best with the test set performs the best in evaluations}. Task performance is measured via closed-loop MPC using CEM; we report mean performance normalized to $[0, 1]$ across 3 episodes per task.}
\label{tab:e8-unseen-expert}
\vspace{0.075in}
\begin{tabular}{lcc@{\hskip 8pt}ccccc}
\toprule
Method & Tok FT & Dyn FT & \makecell{Recon\\PSNR $\uparrow$} & \makecell{Rollout\\$\Delta$PSNR $\uparrow$} & \makecell{Action\\shuf. $\uparrow$} & \makecell{$u_r^{\mathrm{norm}} \downarrow$} & \makecell{Task perf.\\(MPC) $\uparrow$} \\
\midrule
Random policy                                    &   &   & ---     & ---      & ---    & ---     & $0.118$ \\
Base                                             & \xmark & \xmark & $17.29$ & $-14.19$ & $1.08$ & $4.884$ & --- \\
Ours, pre-FT                                     & \xmark & \xmark & $17.07$ & $-14.22$ & $1.20$ & $4.591$ & $0.276$ \\
\midrule
No-op actions                                    & \xmark & \cmark & $17.09$ & $-13.40$ & $1.27$ & $5.052$ & --- \\
No-op actions                                    & \cmark & \cmark & $34.73$ & $-0.09$ & $1.38$ & $1.843$ & $0.163$ \\
Random policy                                    & \xmark & \cmark & $17.07$ & $-13.03$ & $1.50$ & $5.058$ & --- \\
Random policy                                    & \cmark & \cmark & $35.59$ & $+1.74$ & $1.72$ & $1.423$ & $0.228$ \\
Expert policy                                    & \cmark & \cmark & $37.05$ & $+2.64$ & $1.95$ & $1.281$ & $0.362$ \\
Human play                                       & \cmark & \cmark & $36.37$ & $+2.20$ & $1.90$ & $1.185$ & $0.362$ \\
\rowcolor{nhpurple!20} Curiosity ($u_r^{\mathrm{norm}}$)       & \cmark & \cmark & $36.63$ & $+2.66$ & $1.75$ & $1.342$ & $0.325$ \\
\midrule
\textbf{All} (combined)                          & \cmark & \cmark & $\mathbf{38.20}$ & $\mathbf{+3.41}$ & $\mathbf{2.11}$ & $\mathbf{1.090}$ & $\mathbf{0.390}$ \\
\bottomrule
\end{tabular}
\end{table}

\begin{table}[h]
\centering
\small
\caption{\textbf{Targeted data collection for finetuning on 10 unseen tasks (\emph{human play} test set).} We finetune our world model on a set of 10 seen + 10 unseen tasks, varying data source and finetuning strategy. Each data source contains 50 trajectories per task. The offline results in Table~\ref{tab:e8-unseen} are computed over a test set that consists of trajectories from both expert policies and human play; this table complements those results by evaluating only on the \emph{human play} trajectories. We find that \emph{the policy that aligns best with the test set performs the best in evaluations}. Task performance is measured via closed-loop MPC using CEM; we report mean performance normalized to $[0, 1]$ across 3 episodes per task.}
\label{tab:e8-unseen-human}
\vspace{0.075in}
\begin{tabular}{lcc@{\hskip 8pt}ccccc}
\toprule
Method & Tok FT & Dyn FT & \makecell{Recon\\PSNR $\uparrow$} & \makecell{Rollout\\$\Delta$PSNR $\uparrow$} & \makecell{Action\\shuf. $\uparrow$} & \makecell{$u_r^{\mathrm{norm}} \downarrow$} & \makecell{Task perf.\\(MPC) $\uparrow$} \\
\midrule
Random policy                                    &   &   & ---     & ---      & ---    & ---     & $0.118$ \\
Base                                             & \xmark & \xmark & $17.44$ & $-10.70$ & $1.16$ & $2.835$ & --- \\
Ours, pre-FT                                     & \xmark & \xmark & $17.35$ & $-10.82$ & $1.39$ & $2.947$ & $0.276$ \\
\midrule
No-op actions                                    & \xmark & \cmark & $17.32$ & $-9.92$  & $1.56$ & $3.298$ & --- \\
No-op actions                                    & \cmark & \cmark & $34.75$ & $+1.41$  & $1.71$ & $1.128$ & $0.163$ \\
Random policy                                    & \xmark & \cmark & $17.35$ & $-9.55$  & $1.96$ & $3.276$ & --- \\
Random policy                                    & \cmark & \cmark & $36.04$ & $+3.59$  & $2.29$ & $0.980$ & $0.228$ \\
Expert policy                                    & \cmark & \cmark & $34.68$ & $+3.04$  & $2.14$ & $0.982$ & $0.362$ \\
Human play                                       & \cmark & \cmark & $\mathbf{37.84}$ & $\mathbf{+5.58}$ & $\mathbf{2.93}$ & $\mathbf{0.820}$ & $0.362$ \\
\rowcolor{nhpurple!20} Curiosity ($u_r^{\mathrm{norm}}$)       & \cmark & \cmark & $35.47$ & $+3.35$  & $2.24$ & $0.946$ & $0.325$ \\
\midrule
\textbf{All} (combined)                          & \cmark & \cmark & $37.61$ & $+4.63$  & $2.57$ & $0.861$ & $\mathbf{0.390}$ \\
\bottomrule
\end{tabular}
\end{table}

\begin{table}[t]
\centering
\small
\caption{\textbf{Targeted data collection for finetuning on 10 \emph{seen} tasks.} We finetune our world model on a set of 10 seen + 10 unseen tasks, varying data source and finetuning strategy. Each data source contains 50 trajectories per task. Offline metrics were computed using a test set of expert trajectories and human play data in equal amount. Task performance is measured via closed-loop MPC using CEM; we report mean performance normalized to $[0, 1]$ across 3 episodes per task.}
\label{tab:e8-seen}
\vspace{0.075in}
\begin{tabular}{lcc@{\hskip 8pt}ccccc}
\toprule
Method & Tok FT & Dyn FT & \makecell{Recon\\PSNR $\uparrow$} & \makecell{Rollout\\$\Delta$PSNR $\uparrow$} & \makecell{Action\\shuf. $\uparrow$} & \makecell{$u_r^{\mathrm{norm}} \downarrow$} & \makecell{Task perf.\\(MPC) $\uparrow$} \\
\midrule
Random policy                                    &   &   & ---     & ---     & ---    & ---     & $0.083$ \\
Base                                             & \xmark & \xmark & $37.70$ & $+3.00$ & $1.59$ & $1.325$ & --- \\
Ours, pre-FT                                     & \xmark & \xmark & $38.18$ & $+4.24$ & $1.84$ & $1.133$ & $0.256$ \\
\midrule
No-op actions                                    & \xmark & \cmark & $38.18$ & $+4.87$ & $2.11$ & $1.113$ & --- \\
No-op actions                                    & \cmark & \cmark & $39.04$ & $\mathbf{+5.30}$ & $2.04$ & $1.039$ & $0.316$ \\
Random policy                                    & \xmark & \cmark & $38.14$ & $+4.46$ & $2.07$ & $1.063$ & --- \\
Random policy                                    & \cmark & \cmark & $39.09$ & $+5.17$ & $2.07$ & $1.057$ & $0.286$ \\
Expert policy                                    & \cmark & \cmark & $38.71$ & $+5.21$ & $\mathbf{2.16}$ & $\mathbf{0.982}$ & $0.355$ \\
Human play                                       & \cmark & \cmark & $39.00$ & $+5.05$ & $\mathbf{2.16}$ & $1.036$ & $\mathbf{0.384}$ \\
\rowcolor{nhpurple!20} Curiosity ($u_r^{\mathrm{norm}}$)       & \cmark & \cmark & $38.96$ & $+5.24$ & $2.13$ & $1.015$ & $0.332$ \\
\midrule
\textbf{All} (combined)                          & \cmark & \cmark & $\mathbf{39.21}$ & $+5.09$ & $2.08$ & $1.019$ & $0.308$ \\
\bottomrule
\end{tabular}
\end{table}

\begin{table}[h]
\centering
\caption{\textbf{Effect of reward finetuning.} Results for two variants that both extend the pretrained base model with $30$k additional dynamics steps; one jointly trains the dynamics model and a reward head (backpropagating gradients from the reward back into dynamics), and the other is a reward-free control. Mean over all $200$ pretraining tasks on a held-out test set. We do not observe a significant difference in results as a result of finetuning with rewards.}
\label{tab:reward-ft-effect}
\vspace{0.05in}
\begin{tabular}{lr>{\columncolor{nhpurple!20}}rr}
\toprule
Metric                                  & w/o reward & \textbf{w/ reward} & $\Delta_{\text{rew}}$ \\
\midrule
Recon PSNR (dB) $\uparrow$              & $35.67$    & $35.68$   & $+0.01$  \\
Action-shuffle ratio $\uparrow$         & $1.67$     & $1.62$    & $-0.05$  \\
Rollout $\Delta$PSNR (dB) $\uparrow$    & $3.01$     & $3.14$    & $+0.13$  \\
$u_r^{\text{norm}}$ $\downarrow$        & $1.306$    & $1.318$   & $+0.012$ \\
$u_f^{\text{norm}}$ $\downarrow$        & $0.304$    & $0.286$   & $-0.018$ \\
$u_s^{\text{norm}}$ $\downarrow$        & $0.515$    & $0.510$   & $-0.005$ \\
\bottomrule
\end{tabular}
\end{table}

\clearpage
\newpage

\section{Implementation Details}
\label{sec:appendix-implementation-details}

Our Dreamer 4 world model is a reproduction of the original method as described in \citet{hafner2025training}. This section provides an overview of our implementation and design choices.

\textbf{Language embeddings.} For task conditioning we use frozen text embeddings from \texttt{openai/clip-vit-base-patch32} (CLIP; \citet{radford2021learning}), which produces $512$-dimensional continuous embeddings of per-task language instructions. Only the reward prediction and BC policy heads are conditioned on language embeddings.

\textbf{Block-causal Transformer backbone.} Each Transformer is a stack of block-causal layers consisting of \emph{(i)} space self-attention over the per-frame token sequence, \emph{(ii)} causal time self-attention along the temporal axis, and \emph{(iii)} a SiLU-gated MLP with ratio $4$. Attention uses RoPE \citep{su2024roformer} on Q/K with a KV-cache-aware offset, QK-normalization \citep{henry2020query}, and RMSNorm \citep{zhang2019root} pre-norm with no biases on the normalization layers.

\textbf{Modality-aware mask.} Within space self-attention, we apply a modality-aware mask that depends on the role of each token. In the tokenizer encoder, latent queries attend to all tokens while patch queries only attend within the image modality, preventing patch tokens from mixing across modalities before being bottlenecked through the latents; in the decoder, the directions are reversed so patch queries can read from the latent bottleneck but not from each other directly. In the dynamics model, action, shortcut, spatial, and register tokens are mutually visible while agent tokens (reward head and BC policy) are asymmetrically isolated: agent queries attend to everything, but non-agent queries do not see agent keys.

\textbf{Spatial packing.} The tokenizer produces per-frame latents of shape $(n_L, d_b) = (64, 64)$. Before being fed to the dynamics, these are spatially packed at factor $k{=}2$ to shape $(n_\mathrm{spatial}, d_\mathrm{spatial}) = (32, 128)$, halving attention cost along the spatial axis at the cost of doubling channel dimension; the inverse \texttt{unpack} is applied at decode time. Concretely, the dynamics model sees the following token layout per timestep: {\footnotesize\texttt{[ACTION x 1, SHORTCUT x 1, SPATIAL x 32, REGISTER x 4, AGENT x 4]}}
where the action token is produced by a $2$-layer MLP, the shortcut-conditioning token concatenates two embeddings of the discretized noise level $\sigma$ and step size $d{=}1/2^\mathrm{step}$, register tokens \citep{darcet2024vision} are $4$ learnable ``sink'' tokens, and agent tokens are initialized from the per-task CLIP embedding broadcast over time.

\textbf{Loss normalization.} Pixel MSE and LPIPS terms in the tokenizer, the empirical and self-consistency branches of the dynamics objective, and reward two-hot cross-entropy are each separately normalized by their own running RMS before weighting. This decouples loss weights from absolute scale and removes the need to tune them when the dataset, resolution, or backbone changes.

\textbf{Shortcut flow matching.} The dynamics model is trained with the shortcut flow-matching objective of \citet{frans2025one}. The discretized noise level $\sigma$ is indexed by an integer in $\{0, \dots, k_\mathrm{max}\}$ (with $k_\mathrm{max}{=}64$ in our experiments; $0$ is pure noise, $k_\mathrm{max}$ is clean) and the step is indexed by an integer in $\{0, \dots, \log_2(k_\mathrm{max})\}$ corresponding to step size $d{=}1/2^\mathrm{step}$. For a fraction $\rho_\mathrm{self}{=}0.25$ of each batch we apply a self-consistency bootstrap: at $(\sigma, \mathrm{step})$, two coarser half-steps at $\mathrm{step}{+}1$ are run under \texttt{no\_grad} and their averaged velocity is used as a stop-gradient target for the current step's predicted velocity. The remaining $0.75$ of the batch uses the empirical one-step regression term at the finest step.

\textbf{Reward and BC heads.} Both heads read from the agent tokens via attention pooling against a learnable query. The reward head predicts $L{=}8$ multi-step symlog two-hot distributions over $255$ bins on the range $[-10, 10]$. The BC head is a deterministic Gaussian policy with diagonal covariance trained via an MSE loss on the ground-truth $16$-dimensional padded action.

\textbf{Sampling.} At inference we run a shortcut Euler integrator with step size $d{=}0.125$ (\emph{i.e.}, $K{=}8$ substeps) for each new frame: starting from $z\sim\mathcal{N}(0, I)$, we solve $b = (\hat{x}_1 - z)/(1-\sigma)$ followed by $z \leftarrow z + b\cdot d$. We apply context corruption to past tokens.

\textbf{Hyperparameters.} We summarize our hyperparameters in Table~\ref{tab:appendix-hparams}.

\begin{table}[h]
\centering
\parbox{\textwidth}{
\caption{\textbf{Hyperparameters.} Key hyperparameters used to train our world model.}
\label{tab:appendix-hparams}
\vspace{0.05in}
\centering
\begin{tabular}{@{}ll@{}}
\toprule
\textbf{Hyperparameter}                 & \textbf{Value} \\ \midrule
\textbf{\textcolor{nhdpurple}{\underline{\smash{Data}}}} \\
Image resolution                        & $224 \times 224 \times 3$ \\
Action dim. (zero-padded)               & $16$ \\
\\
\textbf{\textcolor{nhdpurple}{\underline{\smash{Tokenizer (\textasciitilde 100M)}}}} \\
Patch size                              & $14$ \\
Embedding dim. ($d_\mathrm{model}$)     & $512$ \\
Heads                                   & $8$ \\
Depth                                   & $12$ \\
MLP ratio                               & $4$ \\
Number of latents ($n_L$)               & $64$ \\
Bottleneck dim. ($d_b$)                 & $64$ \\
MAE keep range                          & $[0.1,\, 1.0]$ \\
LPIPS weight                            & $0.2$ \\
\\
\textbf{\textcolor{nhdpurple}{\underline{\smash{Dynamics (\textasciitilde 230M)}}}} \\
Embedding dim. ($d_\mathrm{model}$)     & $1024$ \\
Heads                                   & $8$ \\
Depth                                   & $16$ \\
MLP ratio                               & $4$ \\
Spatial packing factor                  & $2$ \\
Register tokens                         & $4$ \\
Agent tokens                            & $4$ \\
Self-consistency fraction               & $0.25$ \\
Context corruption ($\tau_\mathrm{ctx}$)& $0.1$ \\
\\
\textbf{\textcolor{nhdpurple}{\underline{\smash{Reward + BC heads (\textasciitilde 20M)}}}} \\
Multi-step horizon ($L$)                & $8$ \\
Reward bins                             & $255$ \\
Reward symlog range                     & $[-10,\, 10]$ \\
\\
\textbf{\textcolor{nhdpurple}{\underline{\smash{Optimization}}}} \\
Optimizer                               & AdamW \\
Learning rate                           & $1\times10^{-4}$ \\
Weight decay                            & $1\times10^{-2}$ \\
Sequence length                         & $24$ \\
Effective batch size                    & Tok: $96$ / Dyn: $512$ \\
\\
\textbf{\textcolor{nhdpurple}{\underline{\smash{Sampling (inference)}}}} \\
Integrator schedule                     & Shortcut \\
Step size ($d$)                         & $0.125$ \\
\\
\textbf{\textcolor{nhdpurple}{\underline{\smash{Planning (CEM)}}}} \\
Plan horizon ($H$)                      & $32$ \\
Replan every ($K$)                      & $16$ \\
CEM Iterations                          & $3$ \\
Population size                         & $32$ \\
Rollouts per candidate                  & $2$ \\
Warm start (mean)                       & BC prior \\
\bottomrule
\end{tabular}%
}
\end{table}

\clearpage
\newpage

\end{document}